\documentclass[10pt,twocolumn,letterpaper]{article}

\usepackage{wacv}
\usepackage{times}
\usepackage{epsfig}
\usepackage{graphicx}
\usepackage{amsmath}
\usepackage{amssymb}
\usepackage{booktabs}
\usepackage{multirow,subcaption}
\usepackage[accsupp]{axessibility}  
\usepackage[table,xcdraw]{xcolor}

\def\wacvPaperID{632} 

\wacvalgorithmstrack   

\wacvfinalcopy 

\ifwacvfinal
\usepackage[breaklinks=true,bookmarks=false]{hyperref}
\else
\usepackage[pagebackref=true,breaklinks=true,colorlinks,bookmarks=false]{hyperref}
\fi

\begin{document}

\title{Modeling the Lighting in Scenes as Style for Auto White-Balance Correction}

\author{Furkan Kınlı\textsuperscript{1} \qquad Doğa Yılmaz\textsuperscript{2} \qquad Barış Özcan\textsuperscript{3} \qquad Furkan Kıraç\textsuperscript{4} \\
Vision and Graphics Lab, Özyeğin University, Türkiye\\
{\tt\small \{furkan.kinli\textsuperscript{1}, furkan.kirac\textsuperscript{4}\}@ozyegin.edu.tr,} \\ {\tt\small \{doga.yilmaz.11481\textsuperscript{2}, baris.ozcan.10097\textsuperscript{3}\}@ozu.edu.tr}
}

\maketitle
\thispagestyle{empty}

\begin{abstract}
   Style may refer to different concepts (\textit{e.g.} painting style, hairstyle, texture, color, filter, \textit{etc}.) depending on how the feature space is formed. In this work, we propose a novel idea of interpreting the lighting in the single- and multi-illuminant scenes as the concept of style. To verify this idea, we introduce an enhanced auto white-balance (AWB) method that models the lighting in single- and mixed-illuminant scenes as the style factor. Our AWB method does not require any illumination estimation step, yet contains a network learning to generate the weighting maps of the images with different WB settings. Proposed network utilizes the style information, extracted from the scene by a multi-head style extraction module. AWB correction is completed after blending these weighting maps and the scene. Experiments on single- and mixed-illuminant datasets demonstrate that our proposed method achieves promising correction results when compared to the recent works. This shows that the lighting in the scenes with multiple illuminations can be modeled by the concept of style. Source code and trained models are available on \url{https://github.com/birdortyedi/lighting-as-style-awb-correction}.
\end{abstract}

\section{Introduction}
\label{sec:intro}

Perceptual systems are generally intended to separate the content and style factors of the observations \cite{6790155}. Words spoken in an unfamiliar accent, letters written in a novel hand-writing style, or the objects displayed under different lighting conditions can be considered as some examples of the style factors integrated into the content of audio, text or image, respectively. Earlier studies \cite{10.5555/2987189.2987190,10.1162/neco.1995.7.5.889,10.2307/56656,6790155} attack this problem with different computational factor models to provide expressive representations of these factors. The term of \textit{factor} is used to represent the well-characterized representation of the observations \cite{NIPS1994_20aee3a5}. Particularly, separating the content from the style in natural images is a challenging problem. Convolutional Neural Networks (CNNs) have the ability to produce generic feature representations, which can be used for independently processing the content and the style of natural images. Previous studies attempt to process the content and the style separately on texture recognition \cite{Cimpoi_2015_CVPR} and synthesis \cite{NIPS2015_a5e00132,10.1145/218380.218446,park2020swapping}, classifying the artistic style \cite{BMVC_28_122} and filter style \cite{Wu_Wu_Singh_Davis_2020,Kinli_2021_CVPR}, style transfer \cite{gatys_artistic_style,Ghiasi2017ExploringTS,huang_adain}, style removal \cite{Kinli_2021_CVPR} and generative image synthesis \cite{Karras_2019_CVPR,Karras2020ada,Karras2021}. These studies demonstrate that the style representation can be distilled by forming a particular feature space of images via learning objectives.

The concept of style can be interpreted in different ways. For example, it can represent the age of a person, the haircut type and wearing glasses or not as the compact style of the face images \cite{Karras_2019_CVPR}. The affine parameters can be extracted by the mapping network where they form the feature maps as the way that packs the different attributes together in the feature space. To achieve this, the mapping network exploits a random vector or the feature vector extracted by pre-trained networks (\textit{e.g.} VGG \cite{Simonyan15}). On the other hand, the concept of style may refer to the painting style of an artist \cite{gatys_artistic_style} or the filters applied to the natural images \cite{Kinli_2021_CVPR}. This time, the affine parameters stand for the correlation between the features, and they can be directly used for manipulating the painting style of an image or removing the filters applied to an image. Based upon these findings, one may argue that any disruptive or modifying factors for the whole image can be modeled as the style factor. 

The image signal processor (ISP) applies consecutive processing operations to the raw-RGB sensor image to obtain the standard RGB (sRGB) output image. Some examples of these operations are noise reduction, white-balance (WB), gamma correction, auto-exposure and tone-mapping. WB is one of the earliest ISP operations applied to the raw-RGB sensor image, which normalizes the effect of different lighting conditions in the captured scene \cite{color_constancy_marc_ebner}. Auto white-balance (AWB) corrects the captured image by estimated illuminant color of the scene, assuming that the illuminant in the scene is global. This operation makes it possible to perceive a particular color in the scene content as the same when viewed under different illuminations, as similar to the human visual system \cite{seeing_black_white_alan_gilchrist}.

Prior works on AWB correction \cite{10.1007/978-3-319-46493-0_23,Hu_2017_CVPR,Barron2015ConvolutionalCC,46440} thoroughly focuses on global illuminant estimation. The recent studies \cite{Xu_2020_CVPR,Hernandez-Juarez_2020_CVPR,Lo_2021_CVPR} achieve significant improvements on this task specialized on single-illuminant scenes. As a common practice, a diagonal-based correction matrix \cite{5719167} is applied to the images to perform WB. More recently, it is shown that the diagonal correction matrix can be replaced by different static non-linear \cite{color_temp_tuning,afifi2019color} or learnable \cite{Afifi_2020_CVPR} functions. Beyond the single-illuminant scenarios, performing single-illuminant AWB algorithms on the scenes illuminated by multiple light sources leads to produce color tint in the sRGB output image. Instead of directly estimating the illumination in the scene, blending the weighting maps of the scenes illuminated by different WB settings \cite{Afifi_2022_WACV} is a recently proposed solution introducing a method for increasing the robustness of AWB on mixed-illuminant scenes.

\paragraph{Contribution:} In this work, we propose a novel AWB method that models the lighting in single- and mixed-illuminant scenes as style. Our proposed method contains a network that learns the effect of different lighting conditions on the scene with the help of the style information of the scene. Assuming that multiple illuminations in the scene basically stands for the additional style information injected to the scene, our model normalizes the feature maps of the encoder by style information in adaptive manner. Then, it simply employs these normalized feature maps for the learning process of the pixel-wise weighting maps of the same scene with different WB settings. Our AWB strategy does not require to apply any illuminant estimation algorithm, but learns to blend the scene and the pixel-wise weighting maps, as practiced in \cite{Afifi_2022_WACV}. We evaluate our method on well-known single illuminant datasets \cite{banic2019unsupervised,global_tonal_adj}, synthetic mixed-illuminant dataset \cite{Afifi_2022_WACV} and night photography rendering set. Moreover, we evaluate the performance of our method when changed the patch size for training and the set of WB settings used to generate the weighting maps.

\section{Related Works}
This section briefly reviews the previous studies on illuminant estimation, WB correction methods and the prior work on learning the style factor.

\subsection{Illuminant Estimation}
The main aim of the illuminant estimation in the previous studies is to predict the global scene illumination color. In the literature, this problem has been attacked with different strategies. The prior work can be divided into two main categories as statistical methods and learning-based approaches. Statistical illuminant estimation methods mostly use some statistical hypothesis in order to estimate the scene illuminant color. Although these methods are computationally-efficient, they struggle to predict the correct illumination color for real-world scenarios. These methods can be listed as gray-world hypothesis \cite{Buchsbaum1980ASP}, white-patch hypothesis \cite{Brainard1986AnalysisOT}, the shades-of-gray \cite{Finlayson2004ShadesOG}, the gray-edges \cite{4287009,6042872}, the bright-and-dark colors PCA \cite{Cheng_14}, the bright pixels \cite{bright_pixels}, the gray pixel \cite{Qian2019RevisitingGP} and the grayness index \cite{qian2019cvpr}. Recently proposed learning-based methods can produce more accurate results, due to the usage of the information from real-world examples, which better represent the real-world illumination. Learning-based illuminant estimation methods include gamut-based methods \cite{Finlayson_Gamut_constrained_illuminant,Forsyth2004ANA,Finlayson_869188,Arjan_generalized_gamut_mapping}, Bayesian methods \cite{brainard1997bayesian,David_Bayesian_surface_illuminant,gehler2008bayesian,Hernandez-Juarez_2020_CVPR}, and neural network-based methods \cite{0a3800c558b24222b2cd1846816c5776,Cardei02,BMVC2015_76}. Advanced neural network-based methods further involve different learning strategies such as patch-wise learning \cite{10.1007/978-3-319-46493-0_23,Hu_2017_CVPR}, achromatic pixel detection \cite{Bianco_2019_CVPR}, metric learning \cite{Xu_2020_CVPR}, contrastive learning \cite{Lo_2021_CVPR}, cross-camera illumination estimation \cite{C5} and weighting map blending \cite{Afifi_2022_WACV}.

\subsection{WB Correction}
Given the estimated illumination color of the scene, a simple diagonal-based correction matrix \cite{5719167} is employed for white-balancing the raw images. In real-world scenarios,  multiple illuminants may occur in the same scene, hence the AWB modules are prone to misinterpret the intensity and the color of the illuminant in particular parts of the scene. This makes WB correction a challenging problem in post-capture. To overcome this problem in multiple illuminant cases, a few attempts are proposed that replaces the diagonal-based correction matrix with a non-linear correction function \cite{afifi2020interactive,afifi2019color,afifi2019color_temp}. Moreover, the recent deep learning-based strategies \cite{Afifi_2020_CVPR,Afifi_2022_WACV} take in place to perform WB in multiple illuminant scenarios.

\begin{figure*}[ht!]
    \centering
    \includegraphics[width=\textwidth]{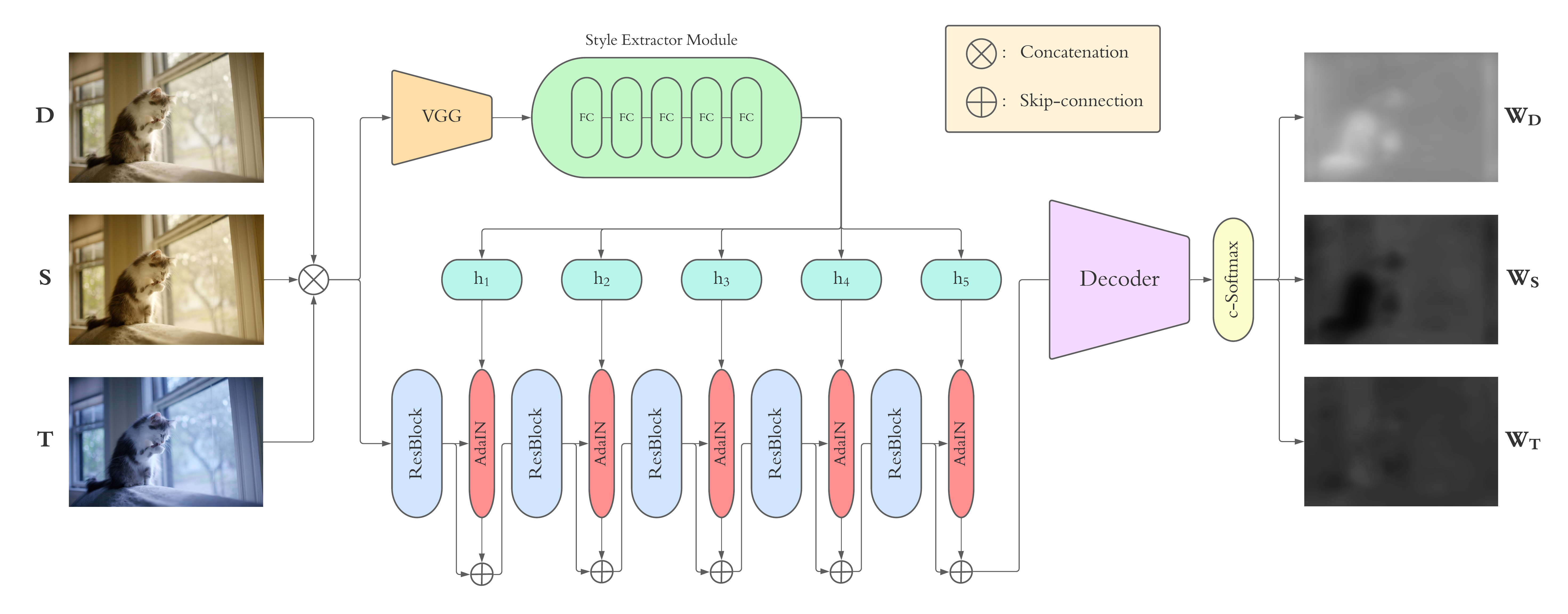}
    \caption{Overall design of proposed learning mechanism for the weighting maps. The latent representations of the images with different WB settings (\textit{i.e.} daylight (\textbf{D}), shade (\textbf{S}), tungsten (\textbf{T})) are fed into the style extractor module, then the affine parameters are computed, and sent to the corresponding AdaIN layer to discard the external style from the feature maps. The decoder part generates the weighting maps for all WB settings (\textit{i.e.} $\textbf{W}_D$, $\textbf{W}_S$, $\textbf{W}_T$).}\label{fig:arch} 
\end{figure*}

\subsection{Learning Style Factor}

Separating the content and style factors is a well-known topic that aims to process the content and the style information independently. With the help of this idea, the content of an observation can be expressed in a more compact way \cite{6790155,10.5555/2987189.2987190,10.1162/neco.1995.7.5.889,10.2307/56656,6790155}, the style of an observation can be recognized \cite{Cimpoi_2015_CVPR,BMVC_28_122,Wu_Wu_Singh_Davis_2020} and manipulated to the desired style \cite{gatys_artistic_style,Ghiasi2017ExploringTS,huang_adain,Kinli_2021_CVPR} or even novel contents can be generated with a particular style \cite{NIPS2015_a5e00132,10.1145/218380.218446,Karras_2019_CVPR,Karras2020ada,Karras2021}. For image domain, the style representation can be captured by designing a dedicated feature space of the images \cite{gatys_artistic_style}. This feature space is mainly built on top of the correlations between different filter responses extracted by a particular layer of a pre-trained network, generally VGG-16 \cite{Simonyan15}. Intuitively, the style may refer to an abstract concept. Depending on how the learning objective shapes the space, the feature space can model the artistic painting style of an artwork, the hairstyle of a person, the texture of a clothing item, or the color of a cat as style. The prior work \cite{Kinli_2021_CVPR} demonstrates that image filters that corrupt the original version of the image can be modeled as style. From this point of view, we propose a method that models the lighting in the scenes with single or multiple illuminants as style. Note that this method does not transfer the particular style of an image to the another one, but aims to normalize the additional injected style information, in which the illumination is considered as the style factor.


\section{Methodology}

Our proposed AWB strategy models the lighting in the scenes with multiple illuminants as style injected to the scene by different light sources. The weighting maps of different WB settings are extracted by using the affine parameters learned by a style extractor module, which is adapted from \cite{Kinli_2021_CVPR}. Proposed network adaptively normalizes different filter responses in any layer of the encoder by a particular style latent code. To finalize the capture, we follow the methods used in \cite{Afifi_2022_WACV} for the inference-time post-processing.

\subsection{Modified Camera Image Signal Processor}

Following the prior work on modified camera ISP \cite{Afifi_2020_CVPR,Afifi_2022_WACV}, we employ a method for producing the high-resolution image with a fixed WB settings (\textit{i.e.} daylight) and additional small images rendered with a set of pre-defined WB settings. The formula for rendering the small images can be described as follows
\begin{equation}
    \hat{I}_{c_i} = M_{c_i}\phi(I_{init})
\end{equation}
where $I_{init}$ is the initial high-resolution image rendered with a fixed WB setting (\textit{i.e.} daylight), $\hat{I}_{c_i}$ represents the output image mapped to the target WB setting, $M_{c_i}$ is the matrix that maps the colors of the initial image represented in a higher-dimensional space, and $\phi(\cdot)$ is a polynomial kernel function projecting the colors of the initial image into the higher-dimensional space. $\phi(\cdot)$ is optimized by minimizing the sum-squared error between the colors of target and source images, as in \cite{Afifi_2022_WACV}. As distinct from \cite{Afifi_2022_WACV}, we consider this part as a pre-processing for training, and save the target images before training, instead of computing them on-the-fly.

After extracting the small images, following the method in \cite{Afifi_2022_WACV}, we employ a learning mechanism for the weighting maps of different scenes with a pre-defined set of WB settings. The details of the learning mechanism are explained in Section \ref{sec:learning}. We use this learned weighting maps for generating the final sRGB output image by linearly combining them with the small images, as shown in the following equation:
\begin{equation}
    \Tilde{I}_{corr}=\sum_{i}W_i\odot \Tilde{I}_{c_{i}}
\end{equation}
where $\Tilde{I}_{corr}$ is the corrected small sRGB image, $\odot$ is Hadamard product, $W_i$ represents the weighting map for $i^{th}$ WB setting (\textit{i.e.} $c_i$), and $\Tilde{I}_{c_{i}}$ denotes the small image rendered with $c_i$.

\begin{figure*}[ht!]
        \centering
        \begin{subfigure}{\textwidth}
                \includegraphics[width=0.24\textwidth]{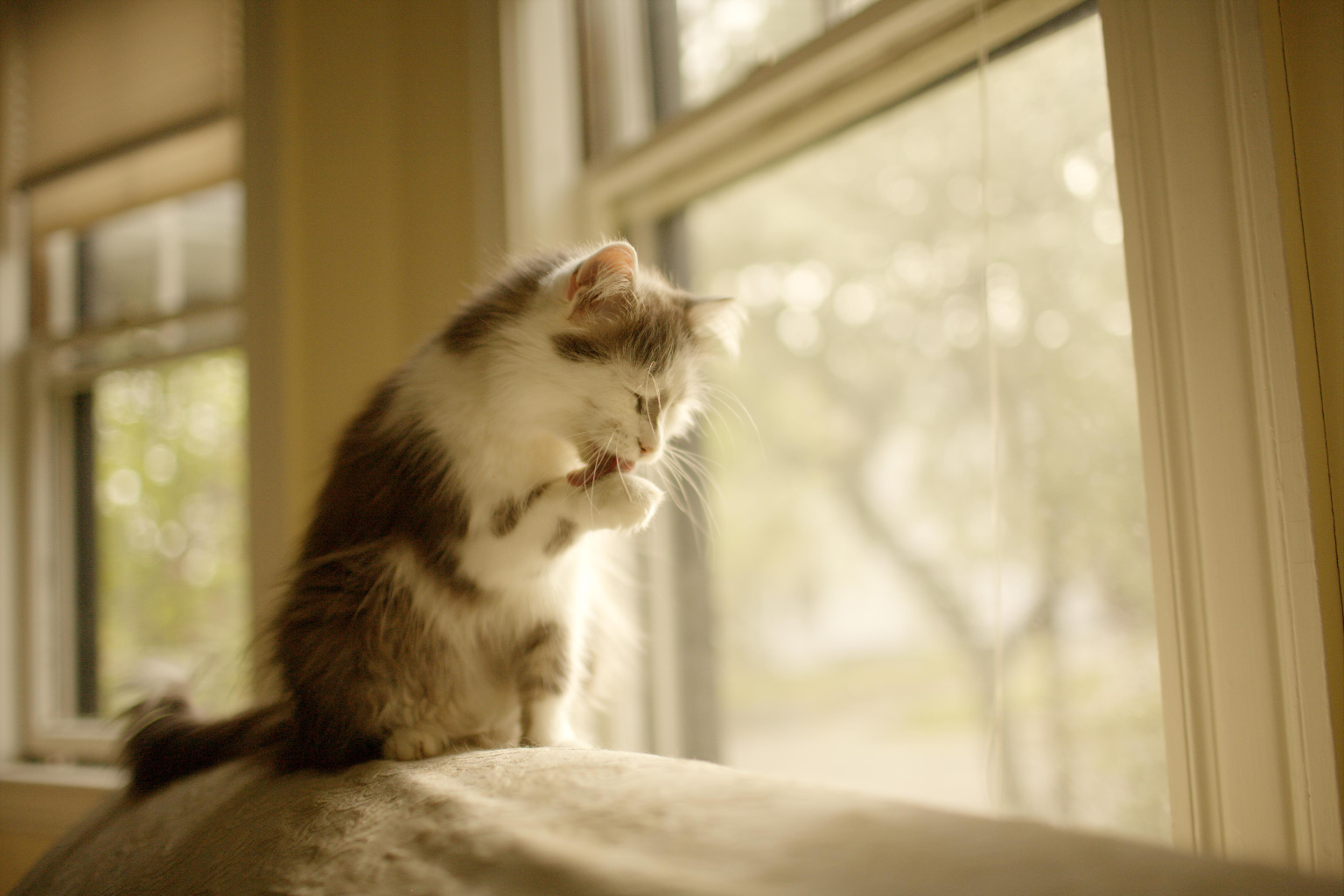}
                \includegraphics[width=0.24\textwidth]{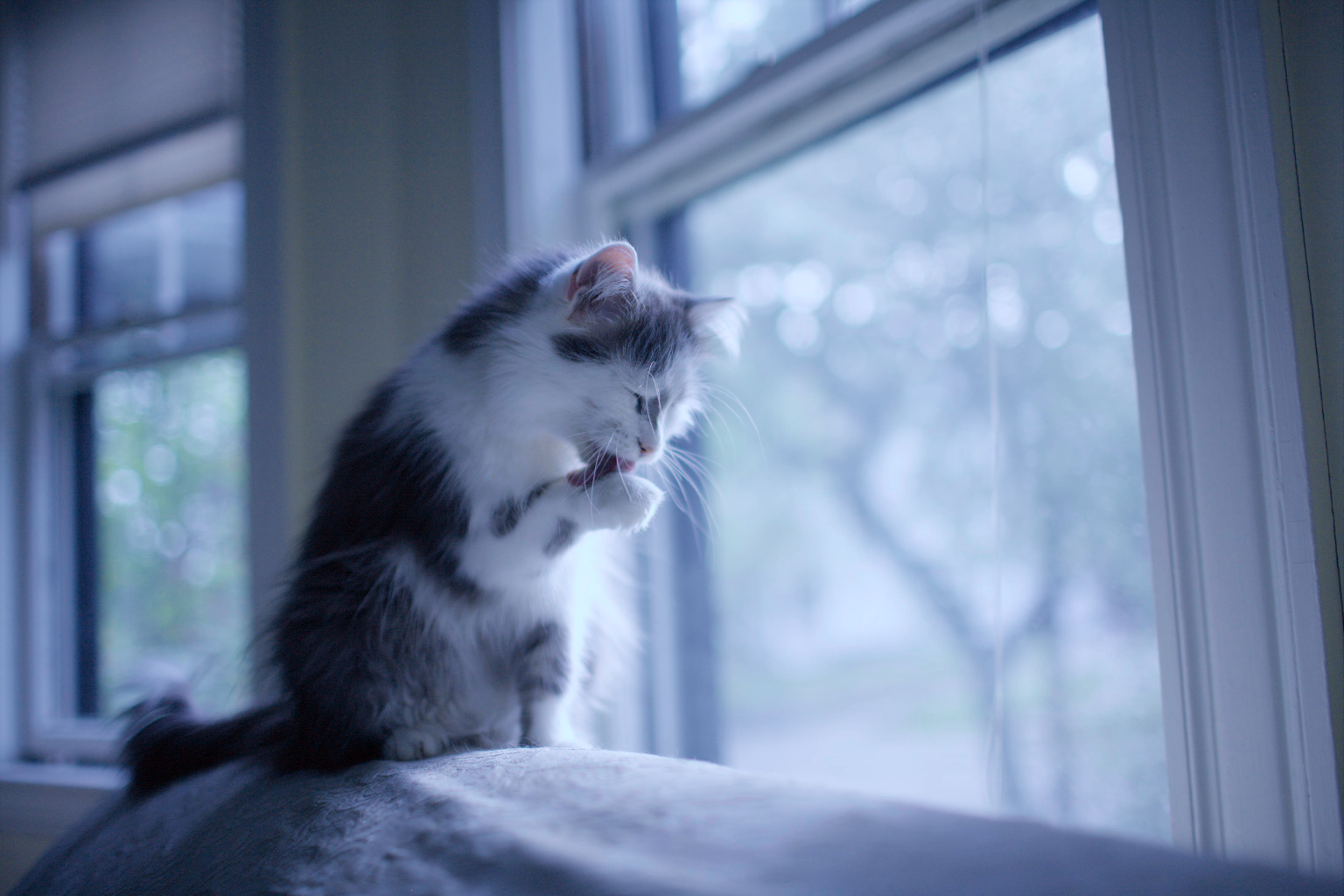}
                \includegraphics[width=0.24\textwidth]{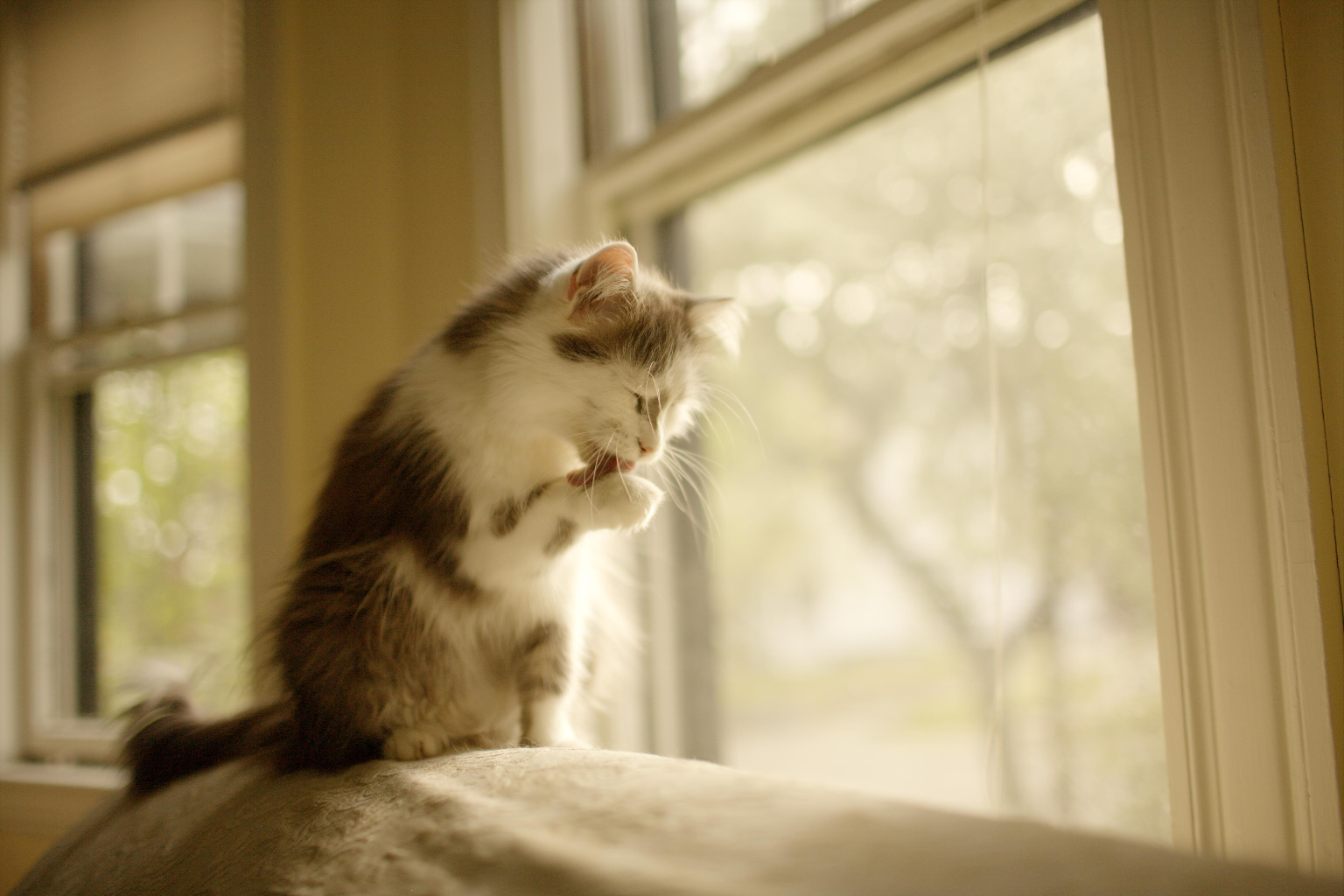}
                 \includegraphics[width=0.24\textwidth]{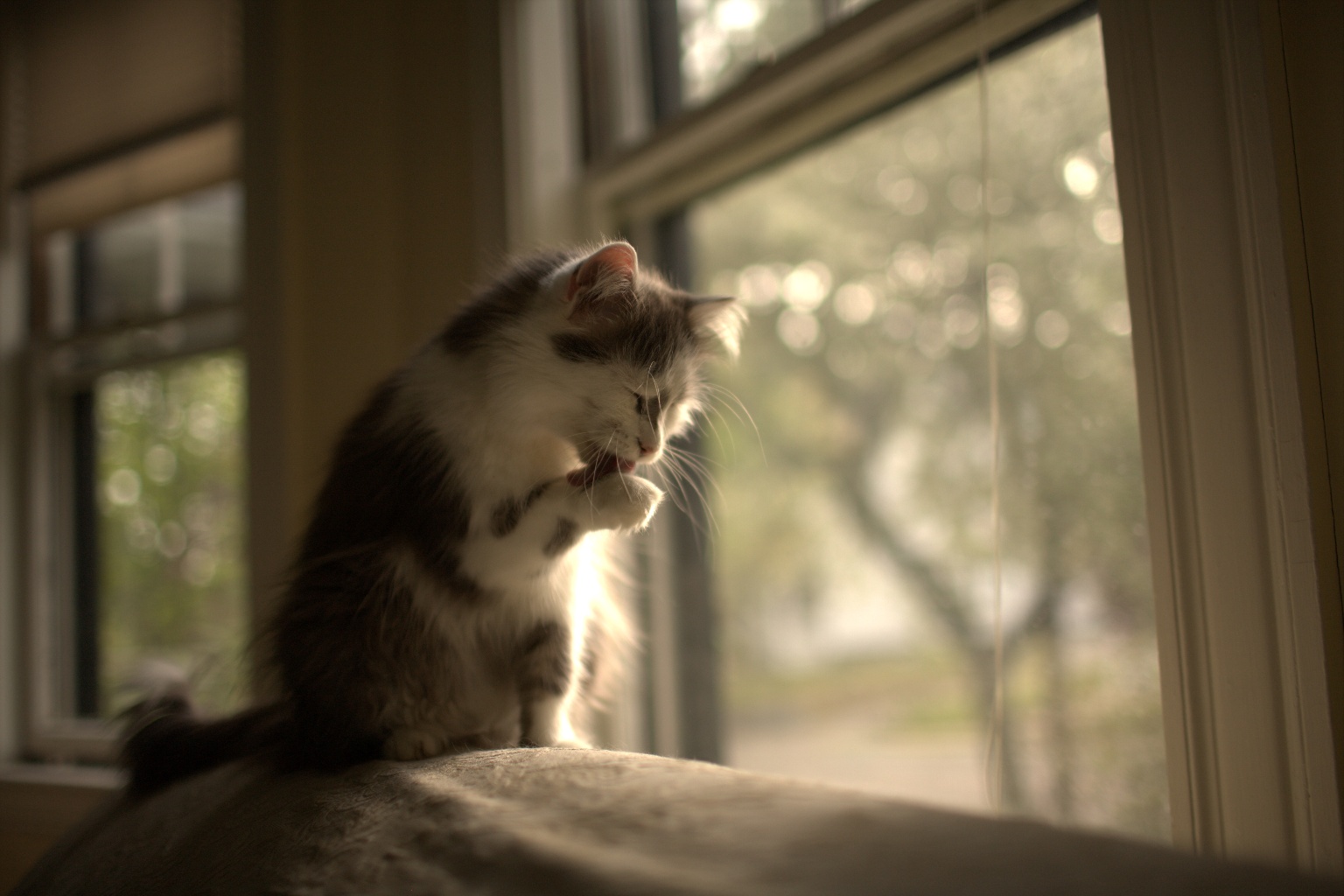}
        \end{subfigure}
        \begin{subfigure}{\textwidth}
                \includegraphics[width=0.24\textwidth]{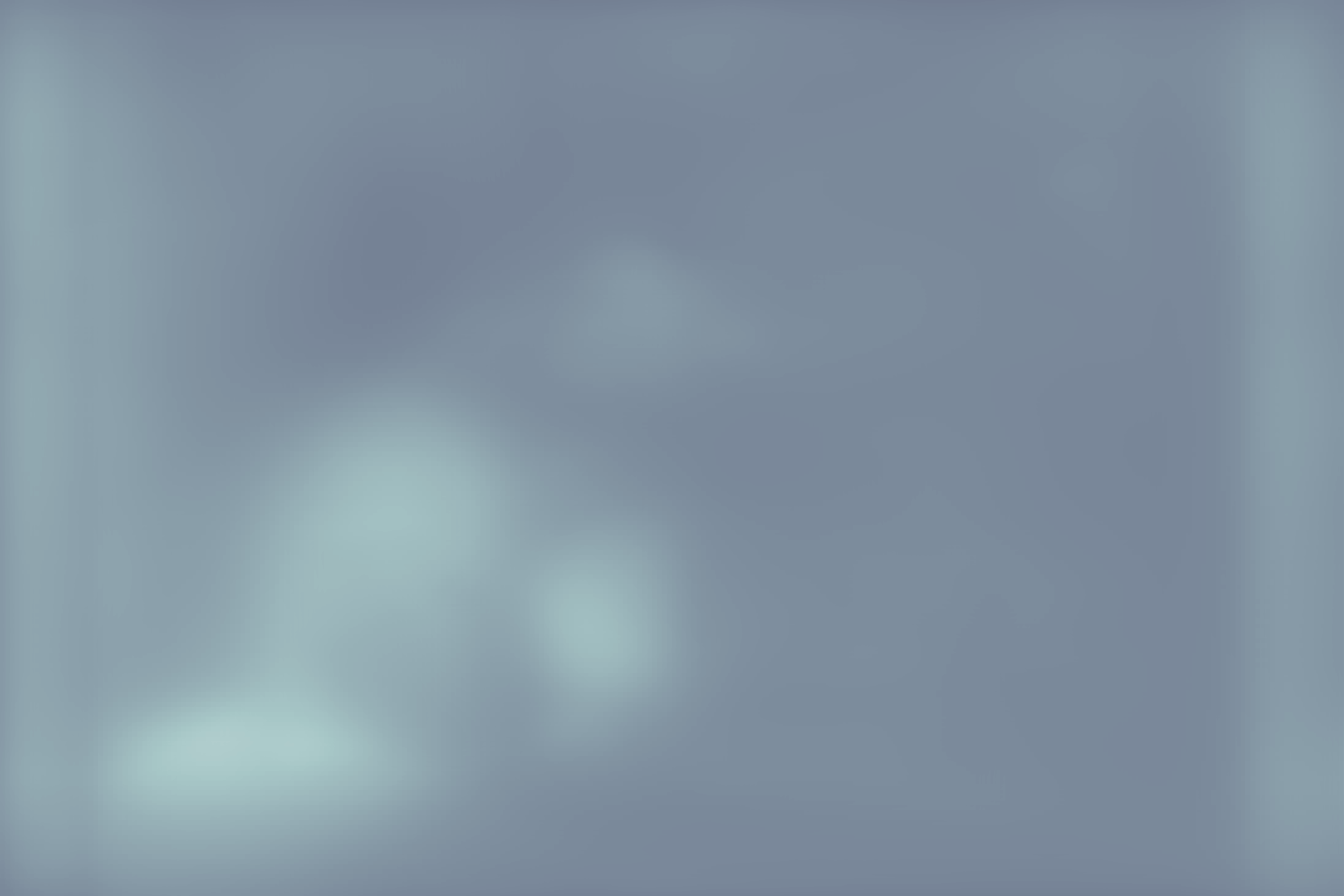}
                \includegraphics[width=0.24\textwidth]{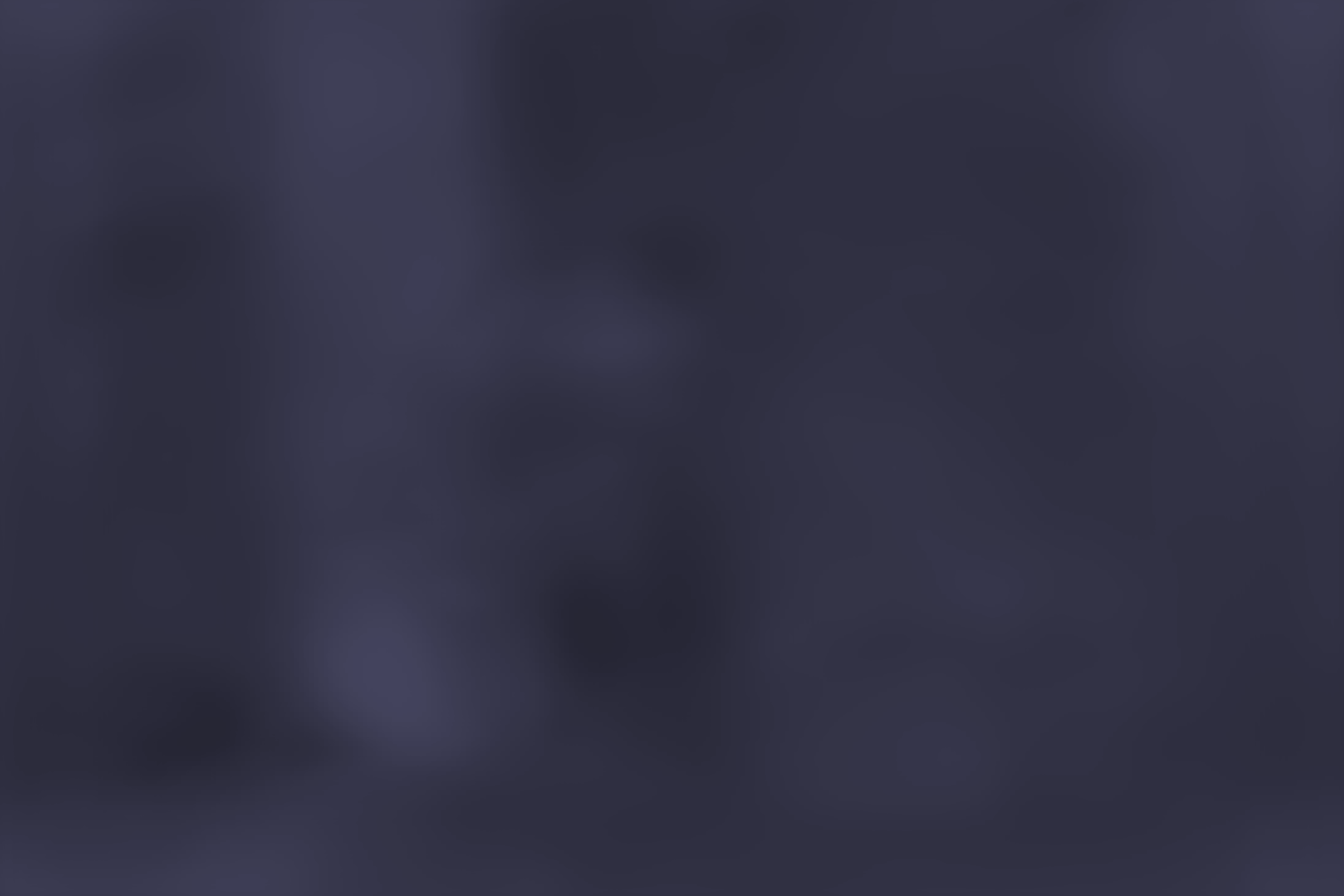}
                \includegraphics[width=0.24\textwidth]{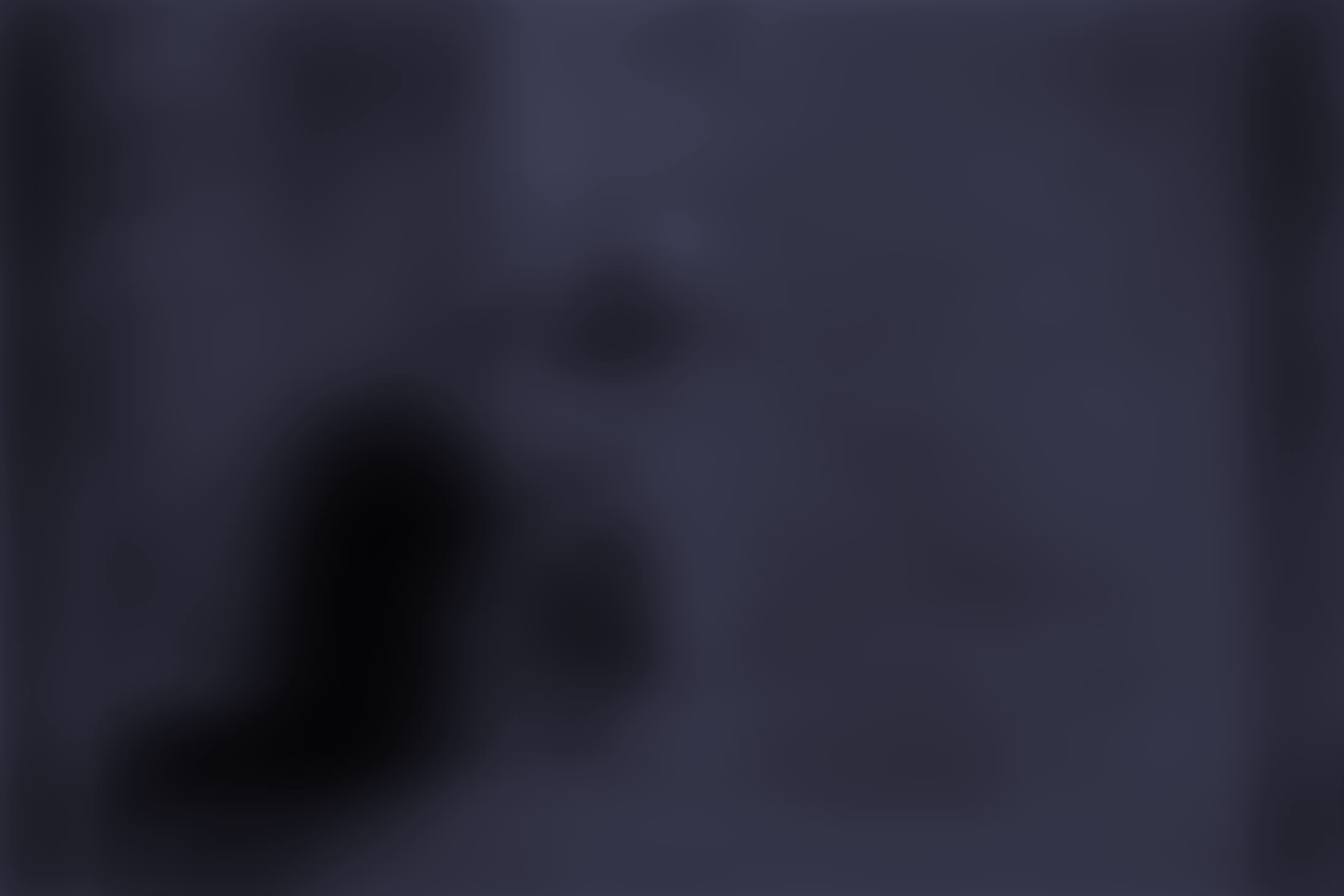}
                \includegraphics[width=0.24\textwidth]{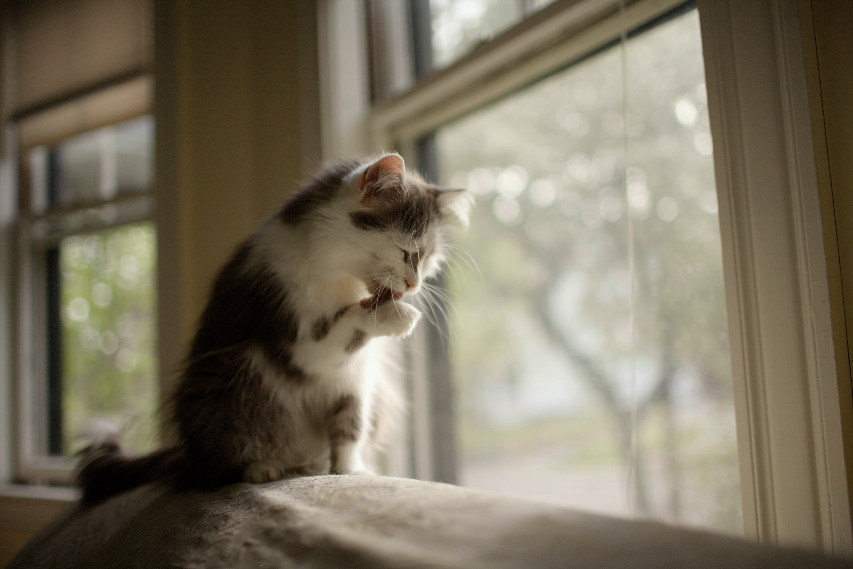} 
        \end{subfigure}
        
        \caption{Example of predictions for the weighting maps and AWB results by blending these maps. We render the linear raw DNG files for the images in MIT-Adobe FiveK dataset \cite{global_tonal_adj}  (id: 2808) in different WB settings. The rendered images with given WB settings and the AWB results of traditional camera pipeline are presented in the first row. The predicted weighting maps and the AWB correction results of our method are demonstrated in the second row.} \label{fig:qual-wm} 
\end{figure*}

\subsection{Learning Weighting Maps by Style} \label{sec:learning}

Given a set of small images $\Tilde{I}_{c_{i}}$, the learning mechanism learns to estimate $\{W_i\}$. In this work, we adapt a style removal network proposed in \cite{Kinli_2021_CVPR} as the learning mechanism of the weighting maps. This network consists of an encoder-decoder structure that employs adaptive feature normalization strategy to all layers of the encoder part. With the help of this strategy, any internal factor in the images can be modeled as an external style, which needs to be discarded or adjusted into another style. The main component to achieve this is Adaptive Instance Normalization (AdaIN) \cite{huang_adain} for each encoder layer, which transfers the feature statistics computed across spatial locations. AdaIN simply aligns the channel-wise mean $\mu$ and variance $\sigma$ of the feature maps of the content image $x$ to the statistics of the style input $y$, as formulated in Equation \ref{eq:adain}.
\begin{equation}
    \label{eq:adain}
    \text{AdaIN}(x, y) = \sigma(y) \left(\frac{x - \mu(x)}{\sigma(x)}\right) + \mu(y)
\end{equation}

To extract the style input for the images, we use a multi-head mapping module that maps the feature representations encoded by a pre-trained VGG network to the style latent space. The style latent code $\textbf{w}$ is fed into different heads for different encoder levels, and each head $h_i$ is attached to a projection layer $p_i$ (\textit{i.e.} fully-connected), which adapts the affine parameters $y_i$ of each normalization layer in the encoder.
\begin{equation}
     \textbf{w} = M(\textbf{z}), y_i = p_i(h_i(\textbf{w}))
\end{equation}
where $\textbf{z}$ is the feature representation of the input image $x$ extracted by VGG, and $M$ denotes the style extractor module mapping the input latent space to the style latent space.

In our design, the style extractor module is composed of 5-layer MLP. The encoder contains 5 residual blocks, each of which has specific AdaIN layer to normalize the feature maps with the affine parameters projected by the corresponding head. The network takes the concatenated feature representations of the small images rendered with different WB settings as input, and learns to produce the weighting maps for these WB settings. As suggested in \cite{Kinli_2021_CVPR}, we use skip-connections between encoder layers to preserve the related information while distilling the style. Overall design of proposed learning mechanism is shown in Figure \ref{fig:arch}.

\captionsetup[subfigure]{labelfont=bf, labelformat=empty}
\begin{figure*}[ht!]
        \centering
        \begin{subfigure}{0.24\textwidth}
                \includegraphics[width=\textwidth]{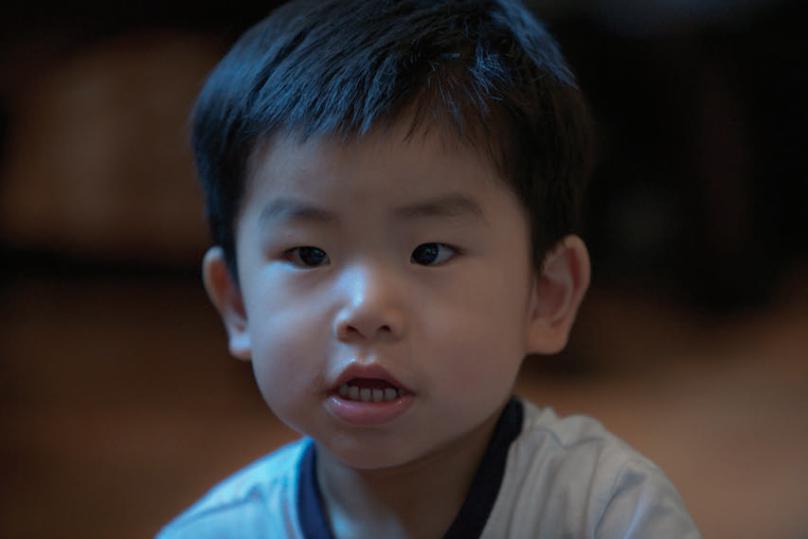}
                \includegraphics[width=\textwidth]{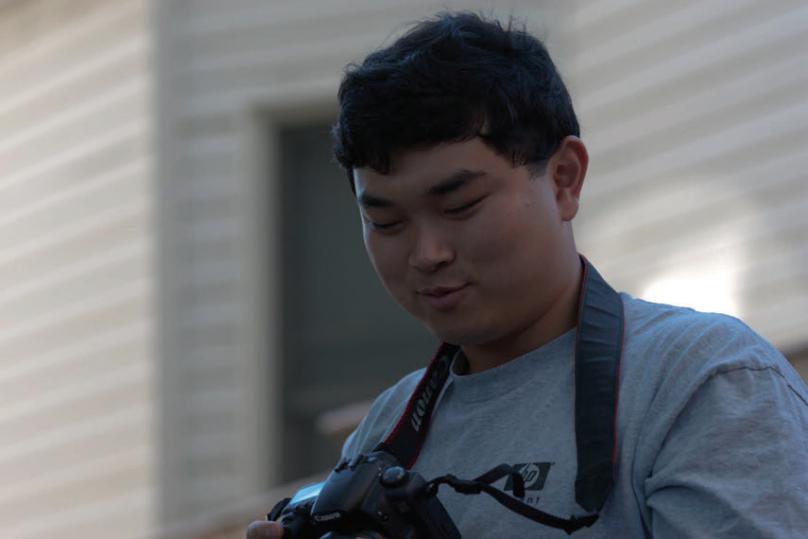}
                \includegraphics[width=\textwidth]{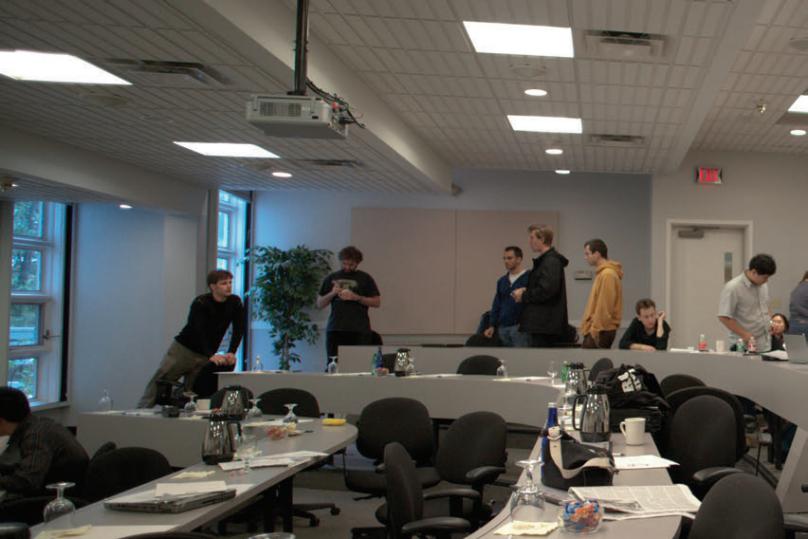}
                \includegraphics[width=\textwidth]{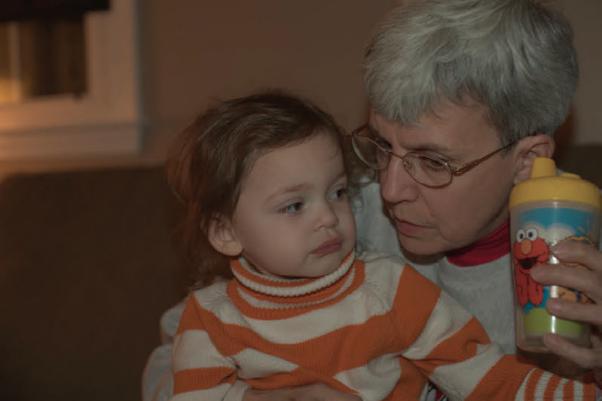}
                \includegraphics[width=\textwidth]{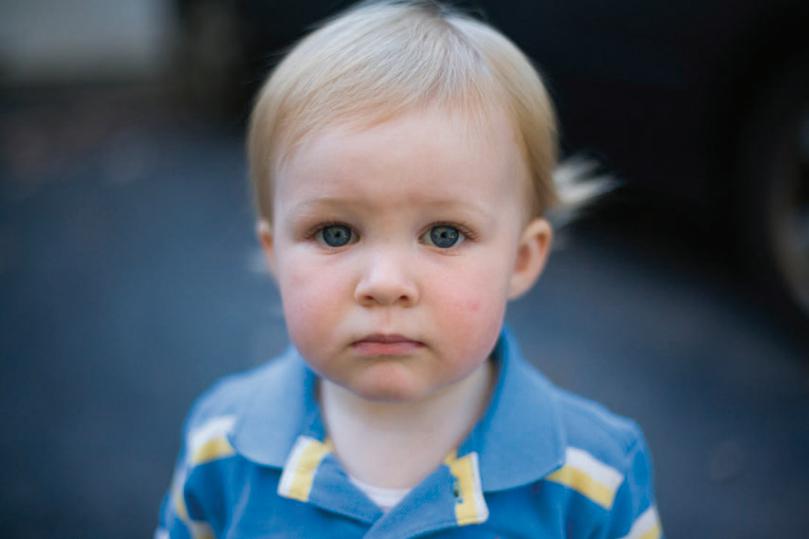}
                \caption{\textbf{(A)} Camera AWB}
        \end{subfigure}
        \begin{subfigure}{0.24\textwidth}
                \includegraphics[width=\textwidth]{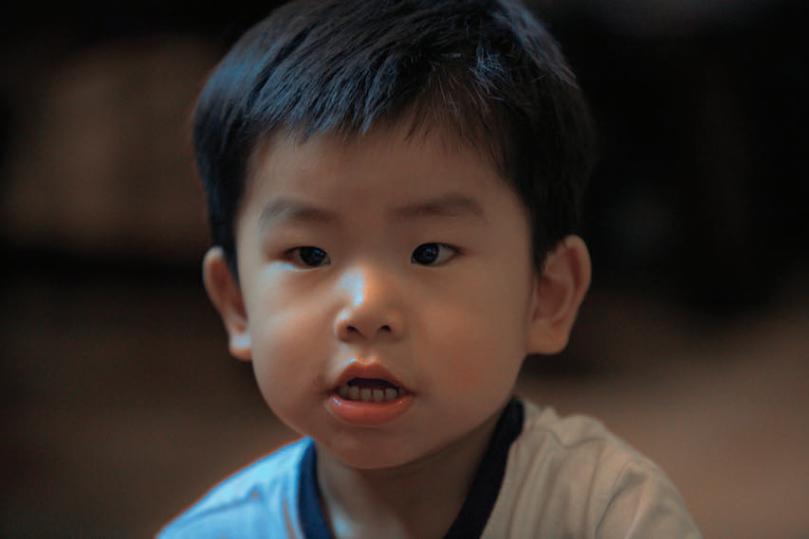}
                \includegraphics[width=\textwidth]{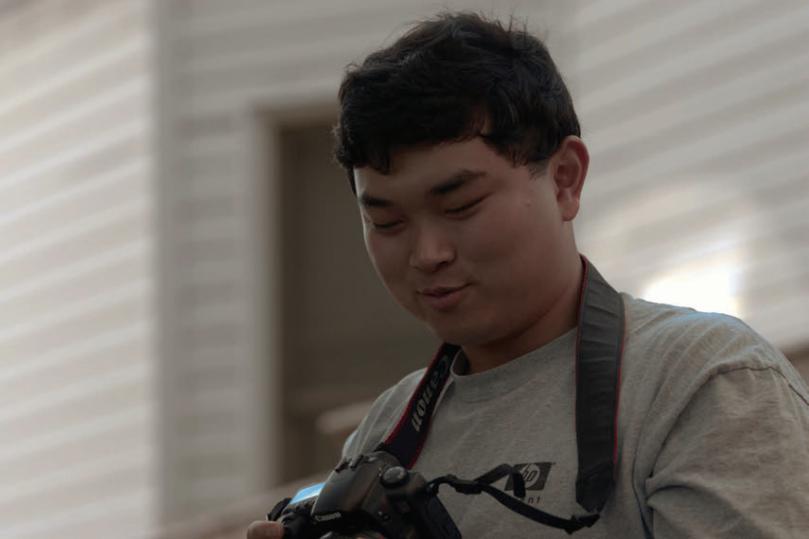}
                \includegraphics[width=\textwidth]{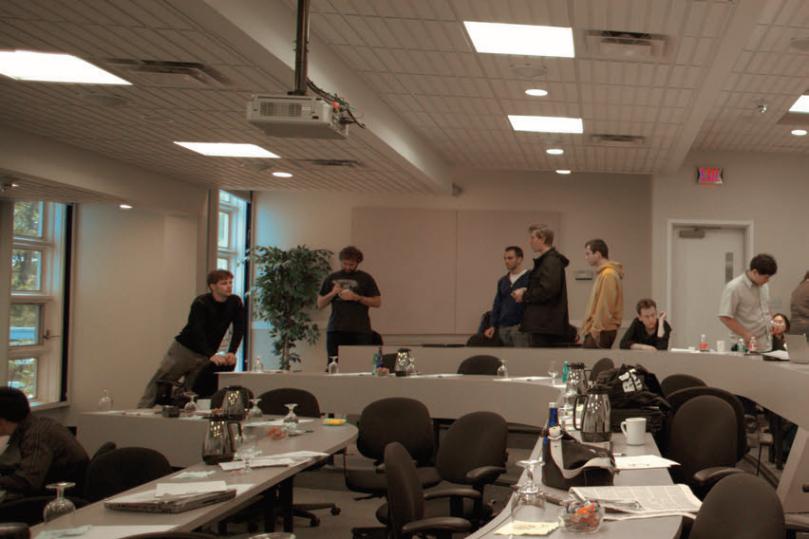}
                \includegraphics[width=\textwidth]{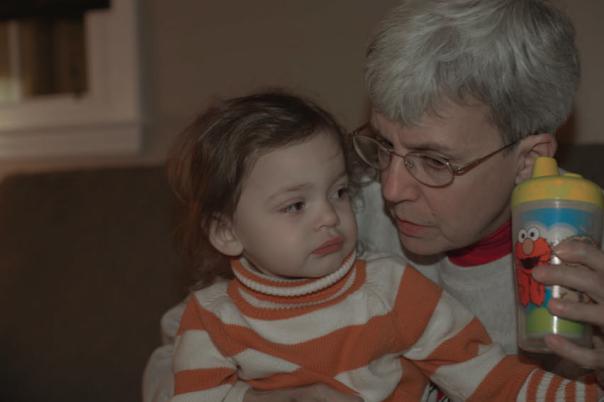}
                \includegraphics[width=\textwidth]{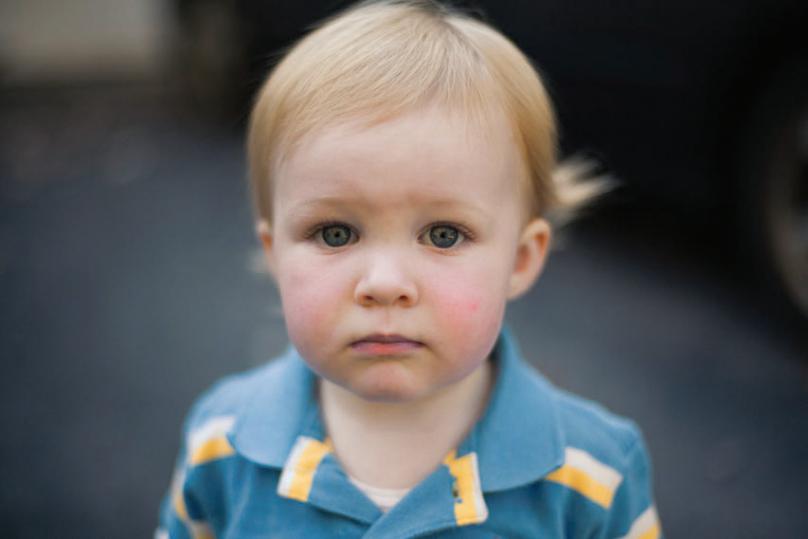}
                \caption{\textbf{(B)} Deep WB}
        \end{subfigure}
        \begin{subfigure}{0.24\textwidth}
                \includegraphics[width=\textwidth]{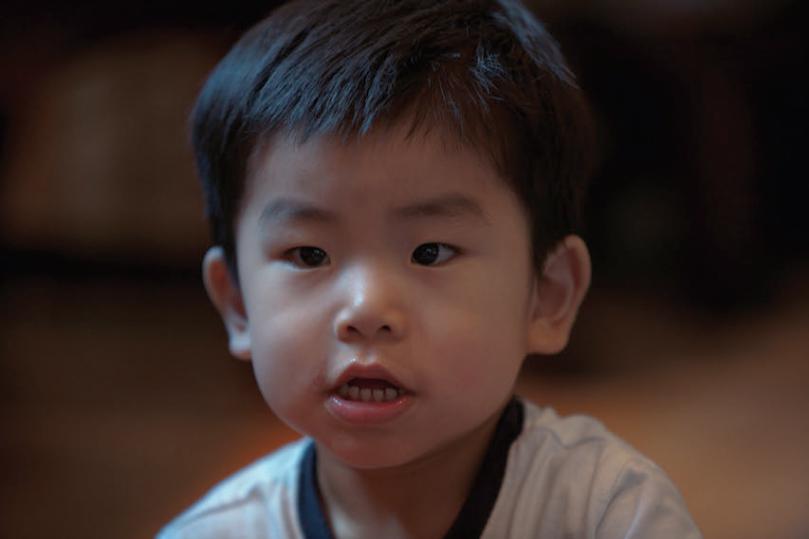}
                \includegraphics[width=\textwidth]{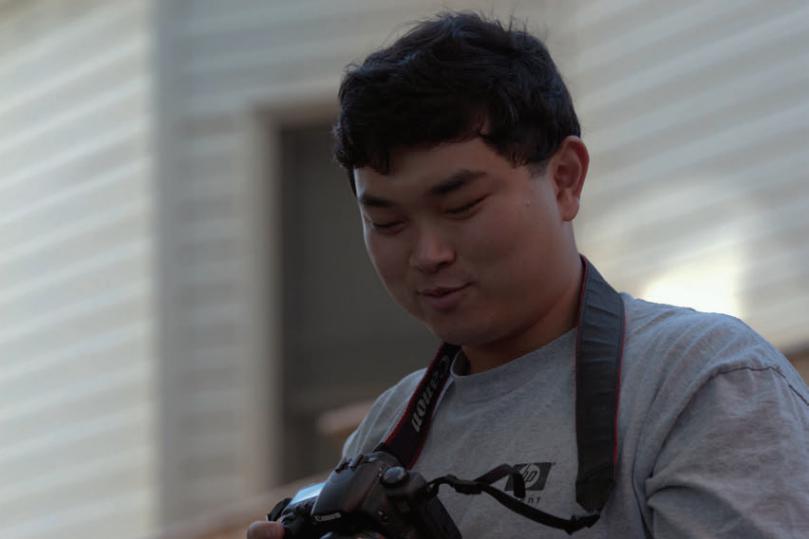}
                \includegraphics[width=\textwidth]{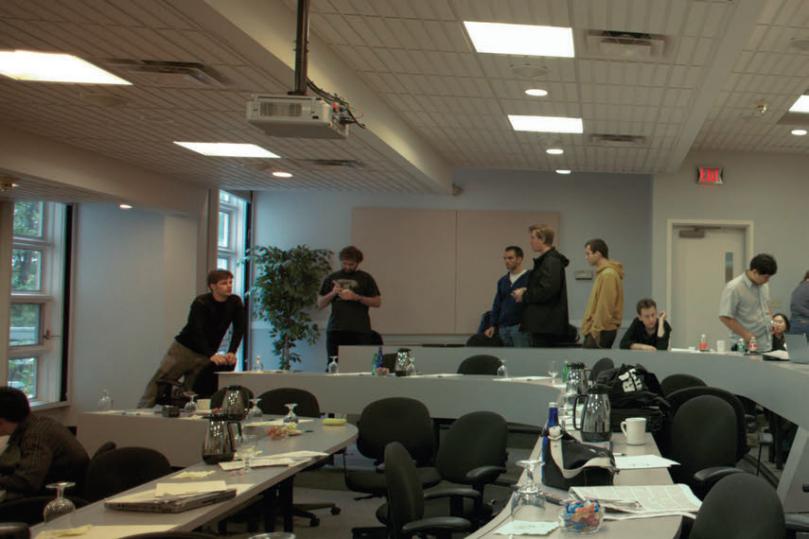}
                \includegraphics[width=\textwidth]{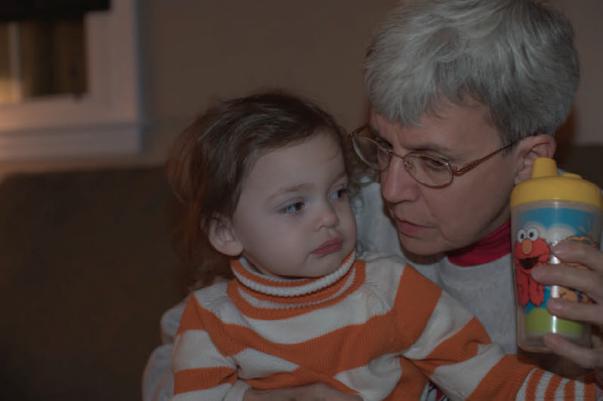} 
                \includegraphics[width=\textwidth]{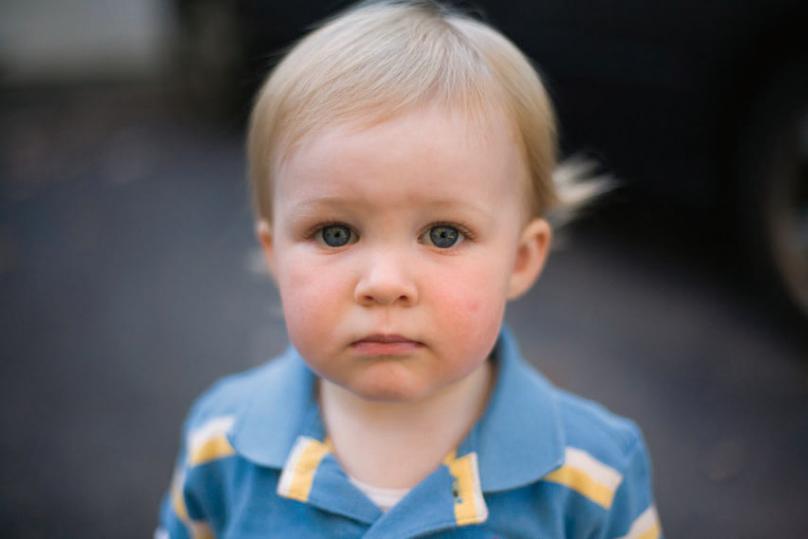}
                \caption{\textbf{(C)} Mixed WB}
        \end{subfigure}
        \begin{subfigure}{0.24\textwidth}
                \includegraphics[width=\textwidth]{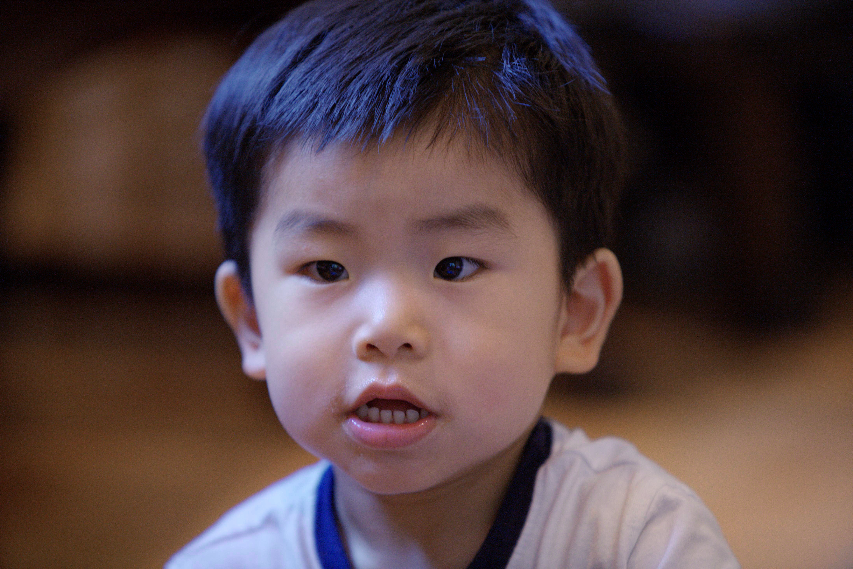}
                \includegraphics[width=\textwidth]{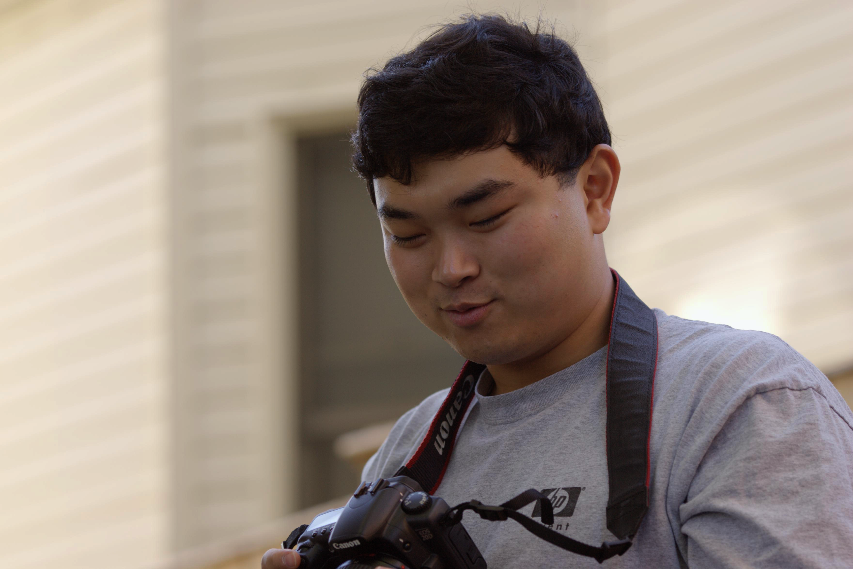}
                \includegraphics[width=\textwidth]{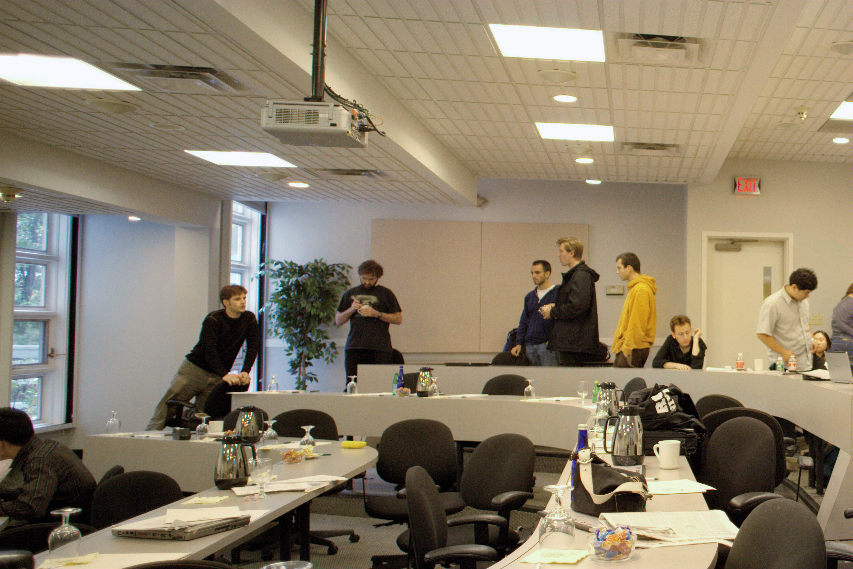}
                \includegraphics[width=\textwidth]{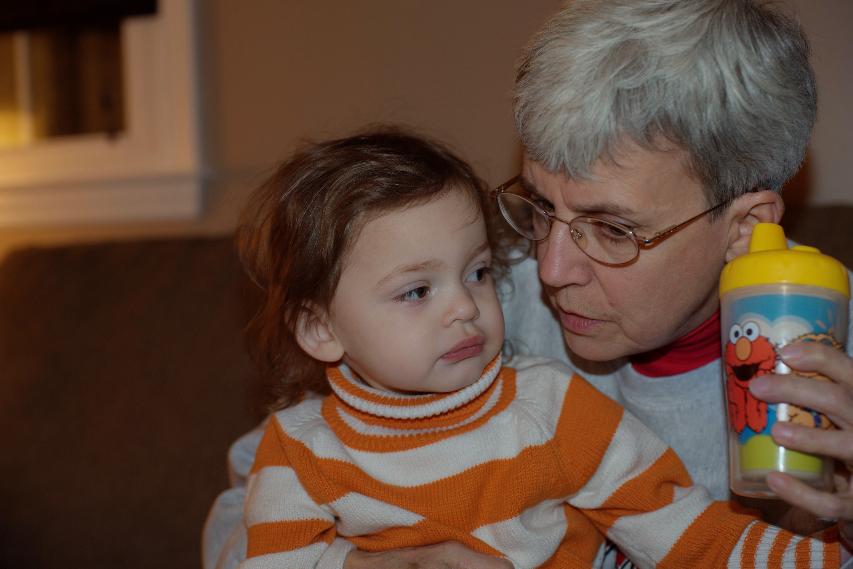}
                \includegraphics[width=\textwidth]{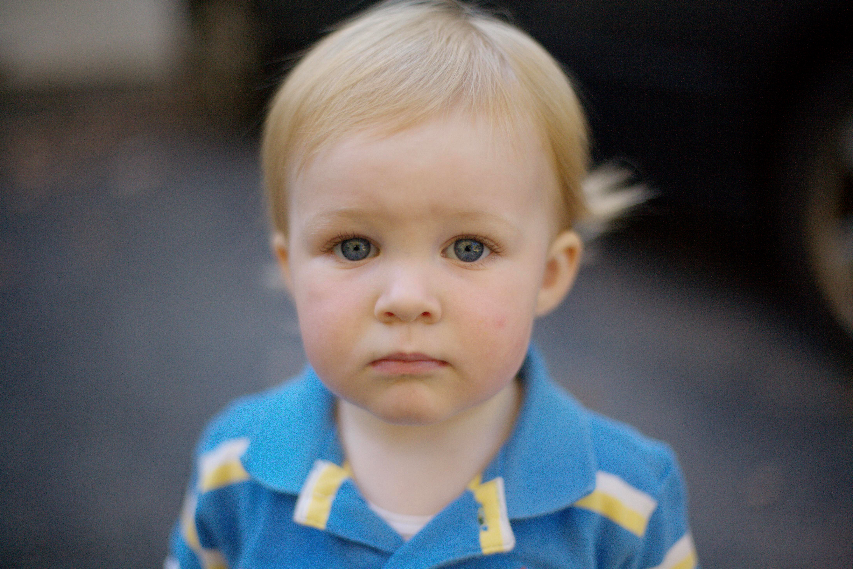}
                \caption{\textbf{(D)} Ours}
        \end{subfigure}
    
        \caption{Comparison of the qualitative results of our AWB method and the other methods on the selected samples from MIT-Adobe FiveK dataset \cite{global_tonal_adj}. We compare our results with the traditional camera AWB, Deep WB \cite{Afifi_2020_CVPR} and Mixed WB \cite{Afifi_2022_WACV}. Image indices from top to bottom: 2882, 606, 659, 2431, 2550.}\label{fig:qual-mit}
\end{figure*}

We optimize this network by minimizing the reconstruction error between the ground truth and corrected patches, as shown in Equation \ref{eq:recon}.
\begin{equation} 
    \label{eq:recon}
    \mathcal{L}_r = || P_{gt} - \sum_{i}{\hat{W}_i \odot P_{c_i}} ||^2_F     
\end{equation}
where $P_{gt}$ and $P_{c_i}$ denote the ground truth patch and input patch rendered with WB setting of $c_i$, $\hat{W}_i$ is the weighting map of $c_i$, as the output of the network. We include the smoothing loss \cite{Afifi_2022_WACV} to our final objective function. 
\begin{equation}
    \label{eq:smooth}
    \mathcal{L}_s = \sum_{i}{|| \hat{W}_i * \nabla_x ||^2_F  + || \hat{W}_i * \nabla_y ||^2_F }
\end{equation}
where $\nabla_x$ and $\nabla_y$ are the horizontal and vertical Sobel filters with $3 \times 3$ kernel size. We did not include the perceptual loss since it dramatically increases the computational complexity of training. Our final objective function can be represented as follows:
\begin{equation}
    \mathcal{L} = \mathcal{L}_r + \lambda\mathcal{L}_s
\end{equation}
where $\lambda$ denotes the regularization coefficient, which is set to 100 in our experiments.

\begin{table*}[ht]
\centering
\caption{Benchmark on single-illuminant Cube+ dataset \cite{banic2019unsupervised}. Following the prior work \cite{afifi2019color}, we reported the mean, first (\textbf{Q1}), second (\textbf{Q2}) and third (\textbf{Q3}) quantile of mean-squared error (\textbf{MSE}), mean angular error (\textbf{MAE}) and color difference ($\Delta$\textbf{E 2000}) metrics. Different WB settings are denoted as \texttt{\{t,f,d,c,s\}}, which refers to tungsten, fluorescent, daylight, cloudy, and shade, respectively. $p$ refers to the patch size. The top results are indicated with colored cells as, the best: {\color{green}{\textbf{green}}}, the second: {\color{yellow}{yellow}}, the third: {\color{red}{red}}.}
\resizebox{\textwidth}{!}{
\begin{tabular}{|lccccccccccccc|}
\hline
\multicolumn{1}{|c|}{\multirow{2}{*}{\textbf{Method}}}                                                       & \multicolumn{4}{c|}{\textbf{MSE}}                                                                                                           & \multicolumn{4}{c|}{\textbf{MAE}}                                                                                                           & \multicolumn{4}{c|}{$\Delta$\textbf{E 2000}}   & \multicolumn{1}{|c|}{\multirow{2}{*}{\textbf{Size}}}                                                                                   \\ \cline{2-13} 
\multicolumn{1}{|c|}{}                                                                                       & \multicolumn{1}{c|}{\textbf{Mean}} & \multicolumn{1}{c|}{\textbf{Q1}} & \multicolumn{1}{c|}{\textbf{Q2}} & \multicolumn{1}{c|}{\textbf{Q3}} & \multicolumn{1}{c|}{\textbf{Mean}} & \multicolumn{1}{c|}{\textbf{Q1}} & \multicolumn{1}{c|}{\textbf{Q2}} & \multicolumn{1}{c|}{\textbf{Q3}} & \multicolumn{1}{c|}{\textbf{Mean}} & \multicolumn{1}{c|}{\textbf{Q1}} & \multicolumn{1}{c|}{\textbf{Q2}} & \multicolumn{1}{c|}{\textbf{Q3}} & \\ \hline
\multicolumn{1}{|l|}{FC4 \cite{Hu_2017_CVPR}}                                                                      & \multicolumn{1}{c|}{371.90}          & \multicolumn{1}{c|}{79.15}        & \multicolumn{1}{c|}{213.41}        & \multicolumn{1}{c|}{467.33}        & \multicolumn{1}{c|}{6.49$^{\circ}$}          & \multicolumn{1}{c|}{3.34$^{\circ}$}        & \multicolumn{1}{c|}{5.59$^{\circ}$}        & \multicolumn{1}{c|}{8.59$^{\circ}$}        & \multicolumn{1}{c|}{10.38}          & \multicolumn{1}{c|}{6.60}        & \multicolumn{1}{c|}{9.76}        & \multicolumn{1}{c|}{13.26}     
   & 5.89 MB     \\ \hline
\multicolumn{1}{|l|}{Quasi-U CC \cite{Bianco_2019_CVPR}}                                                                      & \multicolumn{1}{c|}{292.18}          & \multicolumn{1}{c|}{15.57}        & \multicolumn{1}{c|}{55.41}        & \multicolumn{1}{c|}{261.58}        & \multicolumn{1}{c|}{6.12$^{\circ}$}          & \multicolumn{1}{c|}{1.95$^{\circ}$}        & \multicolumn{1}{c|}{3.88$^{\circ}$}        & \multicolumn{1}{c|}{8.83$^{\circ}$}        & \multicolumn{1}{c|}{7.25}          & \multicolumn{1}{c|}{2.89}        & \multicolumn{1}{c|}{5.21}        & \multicolumn{1}{c|}{10.37}    &  622 MB  \\ \hline
\multicolumn{1}{|l|}{KNN WB \cite{afifi2019color}}                                                                      & \multicolumn{1}{c|}{194.98}          & \multicolumn{1}{c|}{27.43}        & \multicolumn{1}{c|}{57.08}        & \multicolumn{1}{c|}{118.21}        & \multicolumn{1}{c|}{4.12$^{\circ}$}          & \multicolumn{1}{c|}{1.96$^{\circ}$}        & \multicolumn{1}{c|}{3.17$^{\circ}$}        & \multicolumn{1}{c|}{5.04$^{\circ}$}        & \multicolumn{1}{c|}{5.68}          & \multicolumn{1}{c|}{3.22}        & \multicolumn{1}{c|}{4.61}        & \multicolumn{1}{c|}{6.70}       &   21.8 MB  \\ \hline
\multicolumn{1}{|l|}{Interactive WB \cite{afifi2020interactive}}                                                                & \multicolumn{1}{c|}{159.88}          & \multicolumn{1}{c|}{21.94}        & \multicolumn{1}{c|}{54.76}        & \multicolumn{1}{c|}{125.02}        & \multicolumn{1}{c|}{4.64$^{\circ}$}          & \multicolumn{1}{c|}{2.12$^{\circ}$}        & \multicolumn{1}{c|}{3.64$^{\circ}$}        & \multicolumn{1}{c|}{5.98$^{\circ}$}        & \multicolumn{1}{c|}{6.20}          & \multicolumn{1}{c|}{3.28}        & \multicolumn{1}{c|}{5.17}        & \multicolumn{1}{c|}{7.45}    &  \cellcolor[HTML]{C0FDBF}\textbf{38 KB}  \\ \hline
\multicolumn{1}{|l|}{Deep WB \cite{Afifi_2020_CVPR}}                                                                      & \multicolumn{1}{c|}{\cellcolor[HTML]{C0FDBF}\textbf{80.46}}          & \multicolumn{1}{c|}{15.43}        & \multicolumn{1}{c|}{33.88}        & \multicolumn{1}{c|}{74.42}        & \multicolumn{1}{c|}{3.45$^{\circ}$}          & \multicolumn{1}{c|}{1.87$^{\circ}$}        & \multicolumn{1}{c|}{2.82$^{\circ}$}        & \multicolumn{1}{c|}{4.26$^{\circ}$}        & \multicolumn{1}{c|}{4.59}          & \multicolumn{1}{c|}{2.68}        & \multicolumn{1}{c|}{3.81}        & \multicolumn{1}{c|}{5.53}     &   16.7 MB     \\ \hline
\multicolumn{14}{|c|}{\cellcolor[HTML]{B4EBF1}\textbf{Mixed WB \cite{Afifi_2022_WACV} results}}   \\ \hline
\multicolumn{1}{|l|}{$p = 64$, WB=\texttt{\{t,d,s\}}}                                                                      & \multicolumn{1}{c|}{168.38}          & \multicolumn{1}{c|}{\cellcolor[HTML]{FFCCC9}8.97}        & \multicolumn{1}{c|}{19.87}        & \multicolumn{1}{c|}{105.22}        & \multicolumn{1}{c|}{4.20$^{\circ}$}          & \multicolumn{1}{c|}{1.39$^{\circ}$}        & \multicolumn{1}{c|}{2.18$^{\circ}$}        & \multicolumn{1}{c|}{5.54$^{\circ}$}        & \multicolumn{1}{c|}{5.03}          & \multicolumn{1}{c|}{2.07}        & \multicolumn{1}{c|}{3.12}        & \multicolumn{1}{c|}{7.19}     &    \cellcolor[HTML]{C0FDBF}5.09 MB    \\ \hline
\multicolumn{1}{|l|}{$p = 64$, WB=\texttt{\{t,f,d,c,s\}}}                                                                      & \multicolumn{1}{c|}{161.80}          & \multicolumn{1}{c|}{9.01}        & \multicolumn{1}{c|}{\cellcolor[HTML]{FFCCC9}19.33}        & \multicolumn{1}{c|}{90.81}        & \multicolumn{1}{c|}{4.05$^{\circ}$}          & \multicolumn{1}{c|}{1.40$^{\circ}$}        & \multicolumn{1}{c|}{\cellcolor[HTML]{FFCCC9}2.12$^{\circ}$}        & \multicolumn{1}{c|}{4.88$^{\circ}$}        & \multicolumn{1}{c|}{4.89}          & \multicolumn{1}{c|}{2.16}        & \multicolumn{1}{c|}{\cellcolor[HTML]{FFCCC9}3.10}        & \multicolumn{1}{c|}{6.78}     &    \cellcolor[HTML]{FFCCC9}5.10 MB    \\ \hline
\multicolumn{1}{|l|}{$p = 128$, WB=\texttt{\{t,f,d,c,s\}}}                                                                      & \multicolumn{1}{c|}{176.38}          & \multicolumn{1}{c|}{16.96}        & \multicolumn{1}{c|}{35.91}        & \multicolumn{1}{c|}{115.50}        & \multicolumn{1}{c|}{4.71$^{\circ}$}          & \multicolumn{1}{c|}{2.10$^{\circ}$}        & \multicolumn{1}{c|}{3.09$^{\circ}$}        & \multicolumn{1}{c|}{5.92$^{\circ}$}        & \multicolumn{1}{c|}{5.77}          & \multicolumn{1}{c|}{3.01}        & \multicolumn{1}{c|}{4.27}        & \multicolumn{1}{c|}{7.71}     &   \cellcolor[HTML]{FFCCC9}5.10 MB    \\ \hline

\multicolumn{14}{|c|}{\cellcolor[HTML]{B4EBF1}\textbf{Our results}}   \\ \hline
\multicolumn{1}{|l|}{$p = 64$, WB=\texttt{\{t,d,s\}}}         & \multicolumn{1}{c|}{\cellcolor[HTML]{FFCCC9}92.65}         & \multicolumn{1}{c|}{\cellcolor[HTML]{C0FDBF}\textbf{6.52}}        & \multicolumn{1}{c|}{\cellcolor[HTML]{C0FDBF}\textbf{14.23}}       & \multicolumn{1}{c|}{\cellcolor[HTML]{C0FDBF}\textbf{35.01}}       & \multicolumn{1}{c|}{\cellcolor[HTML]{C0FDBF}\textbf{2.47$^{\circ}$}}          & \multicolumn{1}{c|}{\cellcolor[HTML]{C0FDBF}\textbf{0.82$^{\circ}$}}        & \multicolumn{1}{c|}{\cellcolor[HTML]{C0FDBF}\textbf{1.44$^{\circ}$}}        & \multicolumn{1}{c|}{\cellcolor[HTML]{C0FDBF}\textbf{2.49$^{\circ}$}}        & \multicolumn{1}{c|}{\cellcolor[HTML]{C0FDBF}\textbf{2.99}}          & \multicolumn{1}{c|}{\cellcolor[HTML]{C0FDBF}\textbf{1.36}}        & \multicolumn{1}{c|}{\cellcolor[HTML]{C0FDBF}\textbf{2.04}}        & \multicolumn{1}{c|}{\cellcolor[HTML]{C0FDBF}\textbf{3.32}}   &    61.0 MB    \\ \hline

\multicolumn{1}{|l|}{$p = 64$, WB=\texttt{\{t,f,d,c,s\}}}   & \multicolumn{1}{c|}{151.38}        & \multicolumn{1}{c|}{29.49}       & \multicolumn{1}{c|}{56.35}       & \multicolumn{1}{c|}{125.33}      & \multicolumn{1}{c|}{4.18$^{\circ}$}          & \multicolumn{1}{c|}{2.13$^{\circ}$}        & \multicolumn{1}{c|}{3.03$^{\circ}$}        & \multicolumn{1}{c|}{4.81$^{\circ}$}        & \multicolumn{1}{c|}{5.42}          & \multicolumn{1}{c|}{3.11}        & \multicolumn{1}{c|}{4.42}        & \multicolumn{1}{c|}{6.76}      &  61.1 MB \\ \hline

\multicolumn{1}{|l|}{$p = 128$, WB=\texttt{\{t,d,s\}}}  & \multicolumn{1}{c|}{\cellcolor[HTML]{FFFFC7}88.03}        & \multicolumn{1}{c|}{\cellcolor[HTML]{FFFFC7}7.92}       & \multicolumn{1}{c|}{\cellcolor[HTML]{FFFFC7}17.73}       & \multicolumn{1}{c|}{\cellcolor[HTML]{FFFFC7}45.01}       & \multicolumn{1}{c|}{\cellcolor[HTML]{FFFFC7}2.61$^{\circ}$}          & \multicolumn{1}{c|}{\cellcolor[HTML]{FFFFC7}0.93$^{\circ}$}        & \multicolumn{1}{c|}{\cellcolor[HTML]{FFFFC7}1.58$^{\circ}$}        & \multicolumn{1}{c|}{\cellcolor[HTML]{FFFFC7}2.85$^{\circ}$}        & \multicolumn{1}{c|}{\cellcolor[HTML]{FFFFC7}3.24}          & \multicolumn{1}{c|}{\cellcolor[HTML]{FFFFC7}1.50}        & \multicolumn{1}{c|}{\cellcolor[HTML]{FFFFC7}2.30}        & \multicolumn{1}{c|}{\cellcolor[HTML]{FFFFC7}3.95}     &  61.2 MB \\ \hline

\multicolumn{1}{|l|}{$p = 128$, WB=\texttt{\{t,f,d,c,s\}}}  & \multicolumn{1}{c|}{100.24}        & \multicolumn{1}{c|}{10.77}       & \multicolumn{1}{c|}{37.74}       & \multicolumn{1}{c|}{\cellcolor[HTML]{FFCCC9}70.18}       & \multicolumn{1}{c|}{\cellcolor[HTML]{FFCCC9}3.09$^{\circ}$}          & \multicolumn{1}{c|}{\cellcolor[HTML]{FFCCC9}1.15$^{\circ}$}        & \multicolumn{1}{c|}{2.61$^{\circ}$}        & \multicolumn{1}{c|}{\cellcolor[HTML]{FFCCC9}3.87$^{\circ}$}        & \multicolumn{1}{c|}{\cellcolor[HTML]{FFCCC9}3.96}          & \multicolumn{1}{c|}{\cellcolor[HTML]{FFCCC9}1.59}        & \multicolumn{1}{c|}{3.55}        & \multicolumn{1}{c|}{\cellcolor[HTML]{FFCCC9}5.51}    &   61.3 MB \\ \hline
\end{tabular}}
\label{table:results_1}
\end{table*}

\begin{table*}[ht!]
\centering
\caption{Benchmark on mixed-illuminant evaluation set \cite{Afifi_2022_WACV}. Highlights and symbols are the same as in Table \ref{table:results_1}.}
\resizebox{\textwidth}{!}{
\begin{tabular}{|lcccccccccccc|}
\hline
\multicolumn{1}{|c|}{}                                                                                       & \multicolumn{4}{c|}{\textbf{MSE}}                                                                                                           & \multicolumn{4}{c|}{\textbf{MAE}}                                                                                                           & \multicolumn{4}{c|}{ $\Delta$ \textbf{E 2000}}                                                                                                     \\ \cline{2-13} 
\multicolumn{1}{|c|}{\multirow{-2}{*}{\textbf{Method}}}                                                      & \multicolumn{1}{c|}{\textbf{Mean}} & \multicolumn{1}{c|}{\textbf{Q1}} & \multicolumn{1}{c|}{\textbf{Q2}} & \multicolumn{1}{c|}{\textbf{Q3}} & \multicolumn{1}{c|}{\textbf{Mean}} & \multicolumn{1}{c|}{\textbf{Q1}} & \multicolumn{1}{c|}{\textbf{Q2}} & \multicolumn{1}{c|}{\textbf{Q3}} & \multicolumn{1}{c|}{\textbf{Mean}} & \multicolumn{1}{c|}{\textbf{Q1}} & \multicolumn{1}{c|}{\textbf{Q2}} & \textbf{Q3}                   \\ \hline
\multicolumn{1}{|l|}{Gray Pixel \cite{Qian2019RevisitingGP}}                                                                      & \multicolumn{1}{c|}{4959.20}          & \multicolumn{1}{c|}{3252.14}        & \multicolumn{1}{c|}{4209.12}        & \multicolumn{1}{c|}{5858.69}        & \multicolumn{1}{c|}{19.67$^{\circ}$}          & \multicolumn{1}{c|}{11.92$^{\circ}$}        & \multicolumn{1}{c|}{17.21$^{\circ}$}        & \multicolumn{1}{c|}{27.05$^{\circ}$}        & \multicolumn{1}{c|}{25.13}          & \multicolumn{1}{c|}{19.07}        & \multicolumn{1}{c|}{22.62}        & 27.46                          \\ \hline
\multicolumn{1}{|l|}{Grayness index \cite{qian2019cvpr}}                                                                      & \multicolumn{1}{c|}{1345.47}          & \multicolumn{1}{c|}{727.90}        & \multicolumn{1}{c|}{1055.83}        & \multicolumn{1}{c|}{1494.81}        & \multicolumn{1}{c|}{6.39$^{\circ}$}          & \multicolumn{1}{c|}{4.72$^{\circ}$}        & \multicolumn{1}{c|}{5.65$^{\circ}$}        & \multicolumn{1}{c|}{7.06$^{\circ}$}        & \multicolumn{1}{c|}{12.84}          & \multicolumn{1}{c|}{9.57}        & \multicolumn{1}{c|}{12.49}        & 14.60                         \\ \hline
\multicolumn{1}{|l|}{KNN WB \cite{afifi2019color}}                                                                      & \multicolumn{1}{c|}{1226.57}          & \multicolumn{1}{c|}{680.65}        & \multicolumn{1}{c|}{1062.64}        & \multicolumn{1}{c|}{1573.89}        & \multicolumn{1}{c|}{5.81$^{\circ}$}          & \multicolumn{1}{c|}{4.29$^{\circ}$}        & \multicolumn{1}{c|}{5.76$^{\circ}$}        & \multicolumn{1}{c|}{6.85$^{\circ}$}        & \multicolumn{1}{c|}{12.00}          & \multicolumn{1}{c|}{\cellcolor[HTML]{FFCCC9}9.37}        & \multicolumn{1}{c|}{11.56}        & 13.61                          \\ \hline
\multicolumn{1}{|l|}{Interactive WB \cite{afifi2020interactive}}                                                                      & \multicolumn{1}{c|}{1059.88}          & \multicolumn{1}{c|}{\cellcolor[HTML]{FFCCC9}616.24}        & \multicolumn{1}{c|}{896.90}        & \multicolumn{1}{c|}{1265.62}        & \multicolumn{1}{c|}{5.86$^{\circ}$}          & \multicolumn{1}{c|}{4.56$^{\circ}$}        & \multicolumn{1}{c|}{5.62$^{\circ}$}        & \multicolumn{1}{c|}{6.62$^{\circ}$}        & \multicolumn{1}{c|}{\cellcolor[HTML]{FFCCC9}11.41}          & \multicolumn{1}{c|}{\cellcolor[HTML]{FFFFC7}8.92}        & \multicolumn{1}{c|}{\cellcolor[HTML]{FFCCC9}10.99}        & 12.84                          \\ \hline
\multicolumn{1}{|l|}{Deep WB \cite{Afifi_2020_CVPR}}                                                                               & \multicolumn{1}{c|}{1130.60}        & \multicolumn{1}{c|}{621.00}        & \multicolumn{1}{c|}{886.32}        & \multicolumn{1}{c|}{1274.72}          & \multicolumn{1}{c|}{\cellcolor[HTML]{C0FDBF}\textbf{4.53$^{\circ}$}}        & \multicolumn{1}{c|}{\cellcolor[HTML]{C0FDBF}\textbf{3.55$^{\circ}$}}        & \multicolumn{1}{c|}{\cellcolor[HTML]{FFFFC7}4.19$^{\circ}$}        & \multicolumn{1}{c|}{\cellcolor[HTML]{C0FDBF}\textbf{5.21$^{\circ}$}}          & \multicolumn{1}{c|}{\cellcolor[HTML]{FFFFC7}10.93}        & \multicolumn{1}{c|}{\cellcolor[HTML]{C0FDBF}\textbf{8.59}}    
& \multicolumn{1}{c|}{\cellcolor[HTML]{C0FDBF}\textbf{9.82}}  & \cellcolor[HTML]{FFFFC7}11.96                         \\ \hline
\multicolumn{13}{|c|}{\cellcolor[HTML]{B4EBF1}\textbf{Mixed WB \cite{Afifi_2022_WACV} results}}   \\ \hline
\multicolumn{1}{|l|}{$p = 64$, WB=\texttt{\{t,d,s\}}}                                                                      & \multicolumn{1}{c|}{\cellcolor[HTML]{C0FDBF}\textbf{819.47}}          & \multicolumn{1}{c|}{655.88}        & \multicolumn{1}{c|}{\cellcolor[HTML]{FFCCC9}845.79}        & \multicolumn{1}{c|}{\cellcolor[HTML]{FFFFC7}1000.82}        & \multicolumn{1}{c|}{5.43$^{\circ}$}          & \multicolumn{1}{c|}{4.27$^{\circ}$}        & \multicolumn{1}{c|}{4.89$^{\circ}$}        & \multicolumn{1}{c|}{6.23$^{\circ}$}        & \multicolumn{1}{c|}{\cellcolor[HTML]{C0FDBF}\textbf{10.61}}          & \multicolumn{1}{c|}{9.42}        & \multicolumn{1}{c|}{\cellcolor[HTML]{FFFFC7}10.72}        & \cellcolor[HTML]{C0FDBF}\textbf{11.81}                         \\ \hline
\multicolumn{1}{|l|}{$p = 64$, WB=\texttt{\{t,f,d,c,s\}}}                                                                      & \multicolumn{1}{c|}{938.02}          & \multicolumn{1}{c|}{757.49}        & \multicolumn{1}{c|}{961.55}        & \multicolumn{1}{c|}{1161.52}        & \multicolumn{1}{c|}{\cellcolor[HTML]{FFFFC7}4.67$^{\circ}$}          & \multicolumn{1}{c|}{\cellcolor[HTML]{FFFFC7}3.71$^{\circ}$}        & \multicolumn{1}{c|}{\cellcolor[HTML]{C0FDBF}\textbf{4.14$^{\circ}$}}        & \multicolumn{1}{c|}{\cellcolor[HTML]{FFFFC7}5.35$^{\circ}$}        & \multicolumn{1}{c|}{12.26}          & \multicolumn{1}{c|}{10.80}        & \multicolumn{1}{c|}{11.58}        & 12.76                          \\ \hline
\multicolumn{1}{|l|}{$p = 128$, WB=\texttt{\{t,d,s\}}}                                                                      & \multicolumn{1}{c|}{\cellcolor[HTML]{FFCCC9}830.20}          & \multicolumn{1}{c|}{\cellcolor[HTML]{FFFFC7}584.77}        & \multicolumn{1}{c|}{853.01}        & \multicolumn{1}{c|}{\cellcolor[HTML]{C0FDBF}\textbf{992.56}}        & \multicolumn{1}{c|}{\cellcolor[HTML]{FFCCC9}5.03$^{\circ}$}          & \multicolumn{1}{c|}{\cellcolor[HTML]{FFCCC9}3.93$^{\circ}$}        & \multicolumn{1}{c|}{\cellcolor[HTML]{FFCCC9}4.78$^{\circ}$}        & \multicolumn{1}{c|}{5.90$^{\circ}$}        & \multicolumn{1}{c|}{\cellcolor[HTML]{FFCCC9}11.41}          & \multicolumn{1}{c|}{9.76}        & \multicolumn{1}{c|}{11.39}        & \cellcolor[HTML]{FFCCC9}12.53                          \\ \hline
\multicolumn{1}{|l|}{$p = 128$, WB=\texttt{\{t,f,d,c,s\}}}                                                                      & \multicolumn{1}{c|}{1089.69}          & \multicolumn{1}{c|}{846.21}        & \multicolumn{1}{c|}{1125.59}        & \multicolumn{1}{c|}{1279.39}        & \multicolumn{1}{c|}{5.64$^{\circ}$}          & \multicolumn{1}{c|}{4.15$^{\circ}$}        & \multicolumn{1}{c|}{5.09$^{\circ}$}        & \multicolumn{1}{c|}{6.50$^{\circ}$}        & \multicolumn{1}{c|}{13.75}          & \multicolumn{1}{c|}{11.45}        & \multicolumn{1}{c|}{12.58}        & 15.59                          \\ \hline
\multicolumn{13}{|c|}{\cellcolor[HTML]{B4EBF1}\textbf{Our results}}                                                                                                                                                            \\ \hline
\multicolumn{1}{|l|}{$p = 64$, WB=\texttt{\{t,d,s\}}}         & \multicolumn{1}{c|}{868.01}         & \multicolumn{1}{c|}{649.36}        & \multicolumn{1}{c|}{889.00}       & \multicolumn{1}{c|}{1026.98}       & \multicolumn{1}{c|}{5.73$^{\circ}$}          & \multicolumn{1}{c|}{4.48$^{\circ}$}        & \multicolumn{1}{c|}{5.42$^{\circ}$}        & \multicolumn{1}{c|}{6.34$^{\circ}$}        & \multicolumn{1}{c|}{12.11}          & \multicolumn{1}{c|}{10.42}        & \multicolumn{1}{c|}{12.12}        & 13.36  \\ \hline

\multicolumn{1}{|l|}{$p = 64$, WB=\texttt{\{t,f,d,c,s\}}}   & \multicolumn{1}{c|}{1051.07}        & \multicolumn{1}{c|}{760.86}       & \multicolumn{1}{c|}{1024.00}       & \multicolumn{1}{c|}{1332.50}      & \multicolumn{1}{c|}{6.30$^{\circ}$}          & \multicolumn{1}{c|}{4.43$^{\circ}$}        & \multicolumn{1}{c|}{6.01$^{\circ}$}        & \multicolumn{1}{c|}{7.69$^{\circ}$}        & \multicolumn{1}{c|}{14.43}          & \multicolumn{1}{c|}{11.90}        & \multicolumn{1}{c|}{13.11}        & 16.15                          \\ \hline
\multicolumn{1}{|l|}{$p = 128$, WB=\texttt{\{t,d,s\}}}         & \multicolumn{1}{c|}{\cellcolor[HTML]{FFFFC7}822.77}         & \multicolumn{1}{c|}{\cellcolor[HTML]{C0FDBF}\textbf{576.52}}        & \multicolumn{1}{c|}{\cellcolor[HTML]{C0FDBF}\textbf{840.67}}       & \multicolumn{1}{c|}{1025.26}       & \multicolumn{1}{c|}{5.11$^{\circ}$}          & \multicolumn{1}{c|}{\cellcolor[HTML]{FFCCC9}3.93$^{\circ}$}        & \multicolumn{1}{c|}{4.85$^{\circ}$}        & \multicolumn{1}{c|}{\cellcolor[HTML]{FFCCC9}5.51$^{\circ}$}        & \multicolumn{1}{c|}{11.65}          & \multicolumn{1}{c|}{10.63}        & \multicolumn{1}{c|}{11.86}        & 13.02  \\ \hline

\multicolumn{1}{|l|}{$p = 128$, WB=\texttt{\{t,f,d,c,s\}}}  & \multicolumn{1}{c|}{834.28}        & \multicolumn{1}{c|}{629.95}       & \multicolumn{1}{c|}{\cellcolor[HTML]{FFFFC7}842.71}       & \multicolumn{1}{c|}{\cellcolor[HTML]{FFCCC9}1005.59}       & \multicolumn{1}{c|}{5.71$^{\circ}$}          & \multicolumn{1}{c|}{4.57$^{\circ}$}        & \multicolumn{1}{c|}{5.54$^{\circ}$}        & \multicolumn{1}{c|}{6.19$^{\circ}$}        & \multicolumn{1}{c|}{11.79}          & \multicolumn{1}{c|}{9.84}        & \multicolumn{1}{c|}{12.19}        & 13.00                          \\ \hline
\end{tabular}}
\label{table:results_2}
\end{table*}

\subsection{Post-processing}

We have two post-processing operations that can be applied to the learned weighting maps to further improve the quality of the final sRGB image. Following the prior work, we first apply the multi-scale ensembling for the weighting maps. This strategy is mainly based on generating a set of multi-scale weighting maps, bilinear upsampling to the high-resolution, then averaging them for each WB setting. Secondly, we apply edge-aware smoothing (EAS) --with the help of the fast bilateral solver \cite{10.1007/978-3-319-46487-9_38}-- to the weighting maps with the guidance of high-resolution input image. In our experiments, we pick to apply both operations as the performance noticeably increases, as shown in the prior work.

\section{Experiments}

\subsection{Experimental Details}
In our experiments, we have employed RenderedWB dataset \cite{afifi2019color} as the training set. The dataset contains 65,000 sRGB images captured by different cameras, each of which has a specific pre-defined WB settings. Following the setup in the prior work, we have two sets of pre-defined WB settings, which are \texttt{\{t,f,d,c,s\}} and \texttt{\{t,d,s\}}. \texttt{\{t,f,d,c,s\}} refers to tungsten (2850K), fluorescent (3800K), daylight (5500K), cloudy (6500K), and shade (7500K), respectively. Each image has a corresponding accurately white-balanced sRGB image as ground-truth. We did not apply any data augmentation technique to the images. For all settings, we have trained each building block of our proposed model from scratch by using Adam optimizer \cite{kingma2017adam} ($\beta_1=0.9$, $\beta_2 = 0.999$). The learning rate is set to $1e-4$ and we did not employ any scheduling strategy. We have applied two post-processing operations, which are ensembling of multi-scale weighting maps and edge-aware smoothing. We have used the cropped images with the size of $64$ and $128$ for training, and the batch size is set to 32. We have conducted our experiments on a single NVIDIA RTX 2080Ti for 200 epochs. Our implementation is built on top of prior works \cite{Afifi_2022_WACV,Kinli_2021_CVPR}, and done in PyTorch \cite{NEURIPS2019_9015}. 

\paragraph{Inference:} During inference, we produce low-resolution version (\ie $384 \times 384$) of the input images with the predefined WB settings, and concatenate them in order to feed into the proposed model. The model produces the weighting maps as output, to blend the final sRGB output. Before post-processing, we resize the weighting maps to the input resolution.

\subsection{Evaluation Sets}
To evaluate our method, we have used four different evaluation sets for both scenarios, which are namely Cube+ \cite{banic2017unsupervised} and MIT-Adobe FiveK \cite{bychkovsky2011learning}, and mixed-illuminant evaluation set proposed by \cite{Afifi_2022_WACV} and night photography rendering set \cite{Ershov_2022_CVPR}. The Cube+ contains 1,707 single illumination color-calibrated images taken with Canon EOS 550D camera during various seasons. The MIT-Adobe FiveK dataset contains 5,000 images captured by different DSLR cameras where each image is manually retouched by multiple experts to correct the white-balance of the images. Moreover, we have used mixed-illuminant cases for our evaluation. The mixed-illuminant test set has 150 synthetic images composed of multiple illuminantions, which is rendered from 3D scenes modeled in Autodesk 3Ds Max \cite{autodesk}.

\captionsetup[subfigure]{labelfont=bf, labelformat=parens}
\begin{figure}[!t]
        \centering
        \begin{subfigure}[b]{0.24\linewidth}
                \includegraphics[width=\linewidth]{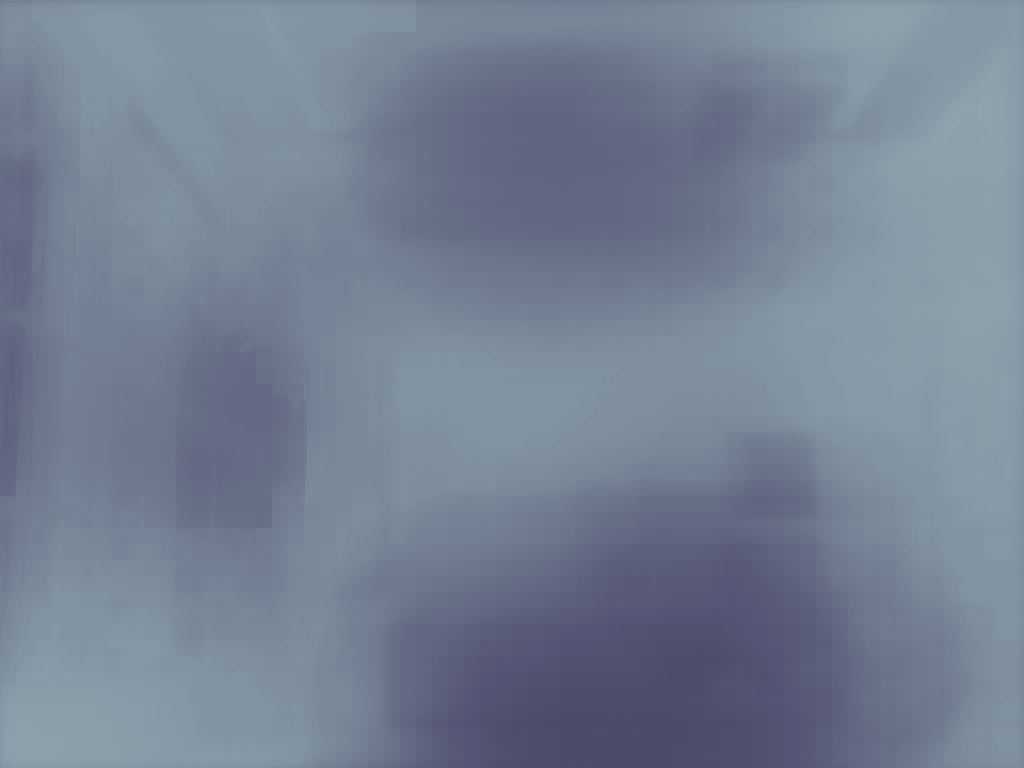}
                \includegraphics[width=\linewidth]{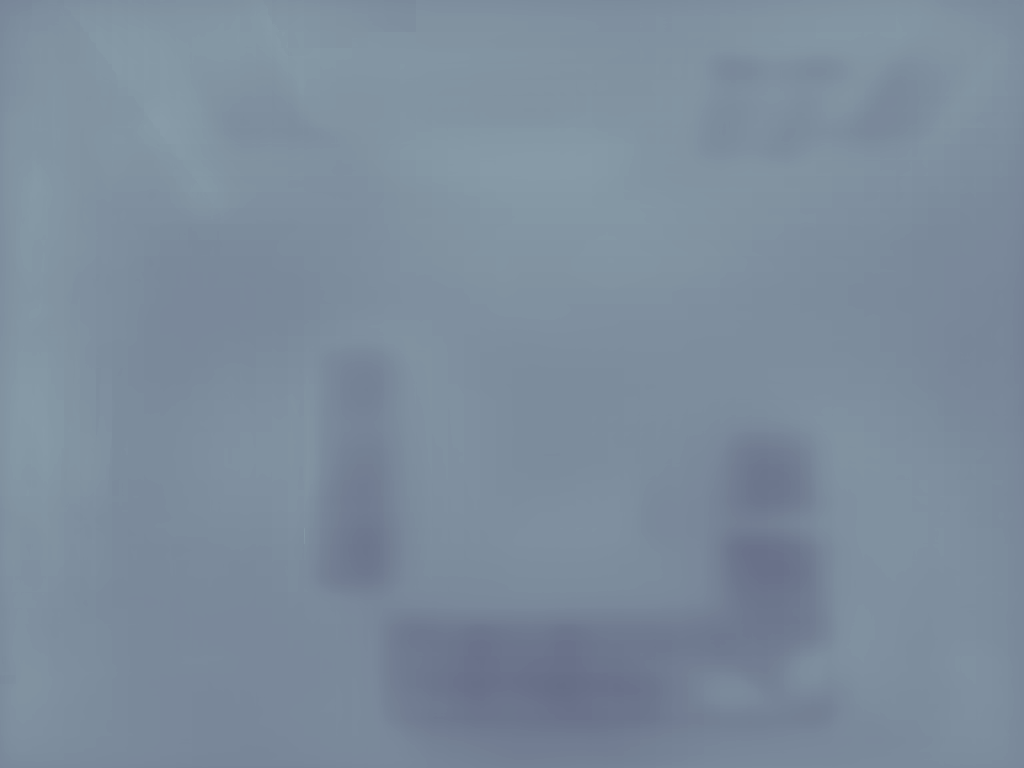}
                \includegraphics[width=\linewidth]{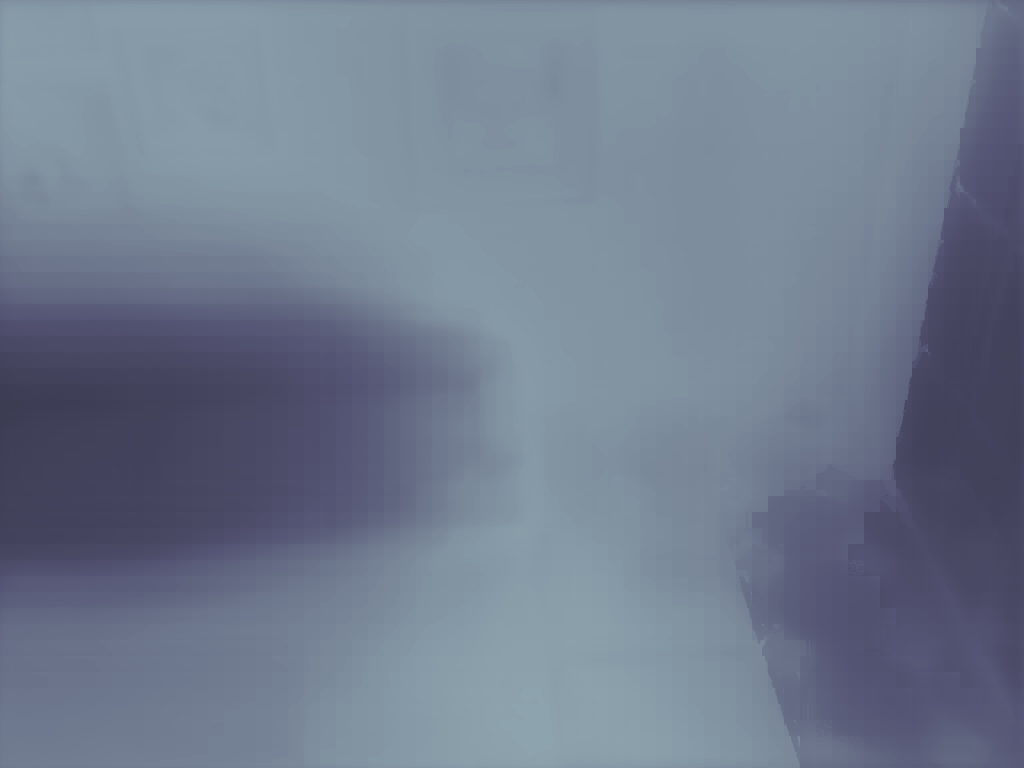}
                \includegraphics[width=\linewidth]{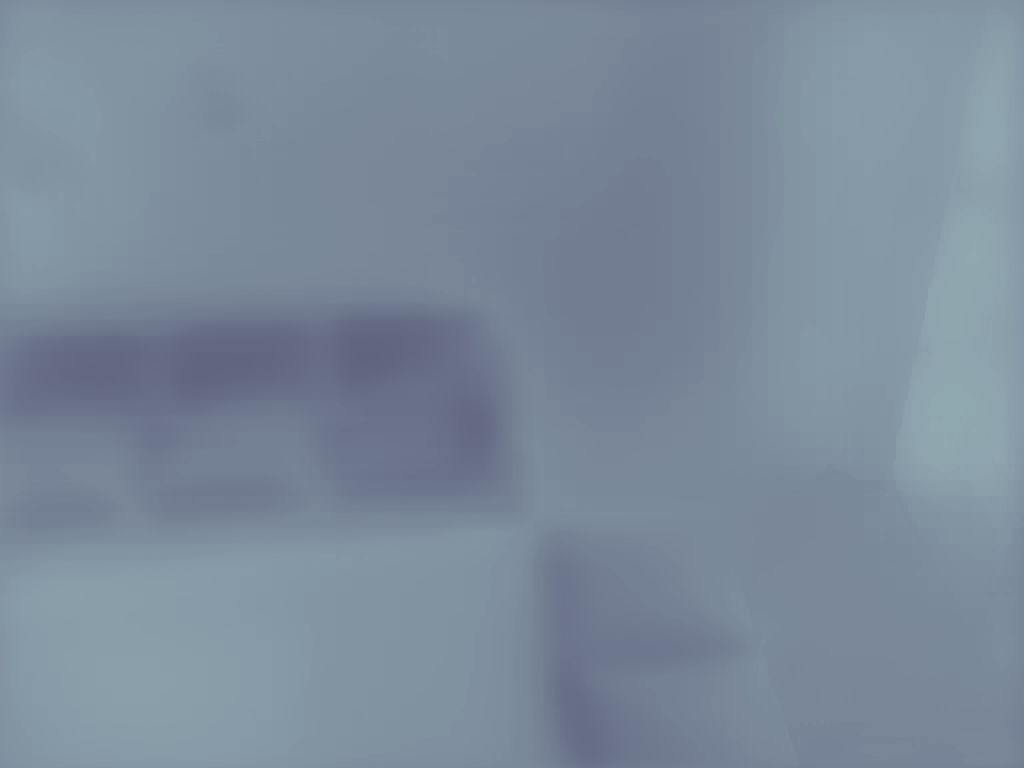}
                \includegraphics[width=\linewidth]{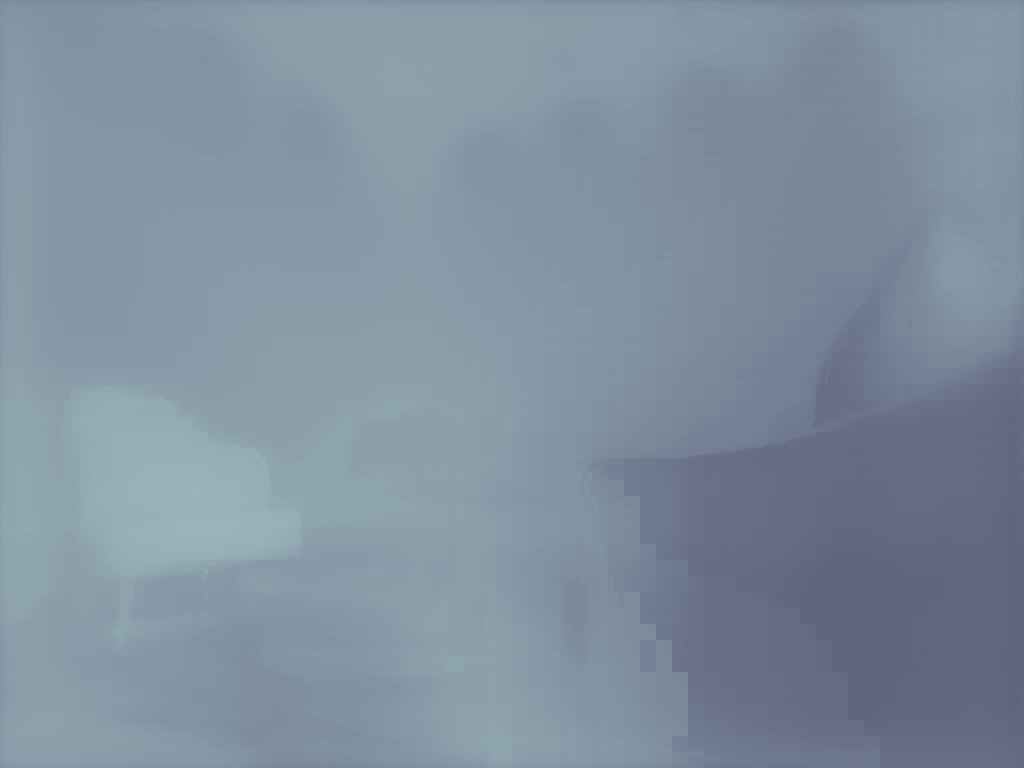}
                \includegraphics[width=\linewidth]{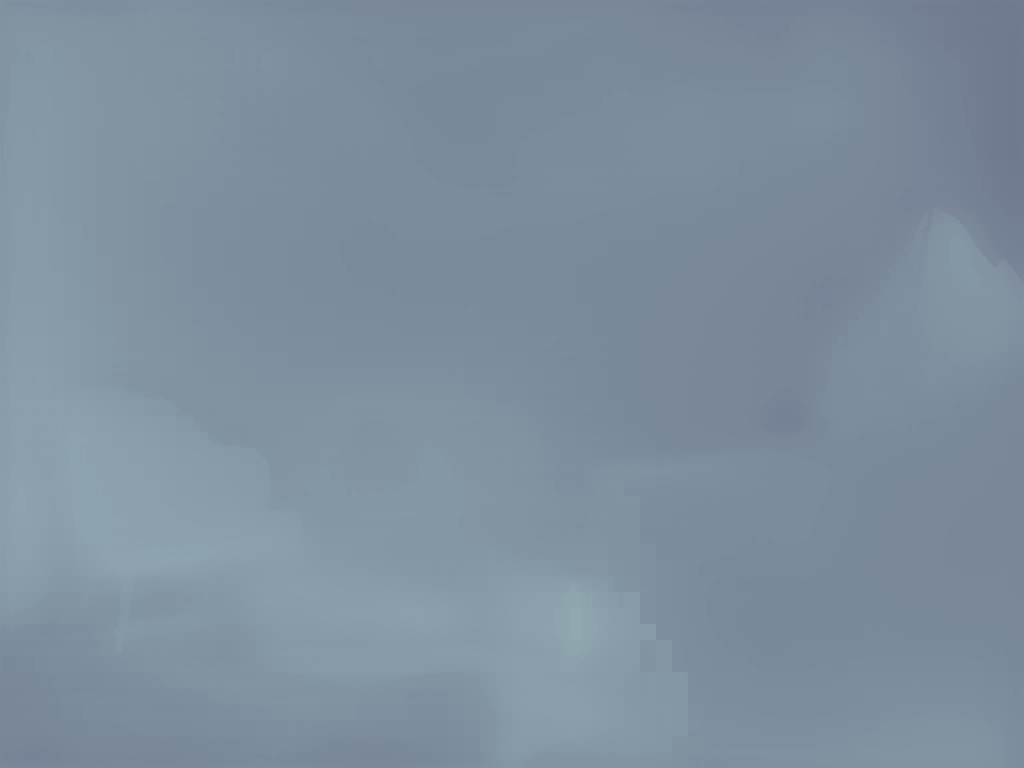}
                \caption{$\textbf{W}_D$}
                \label{fig:camera_awb}
        \end{subfigure}       
        \begin{subfigure}[b]{0.24\linewidth}
                \includegraphics[width=\linewidth]{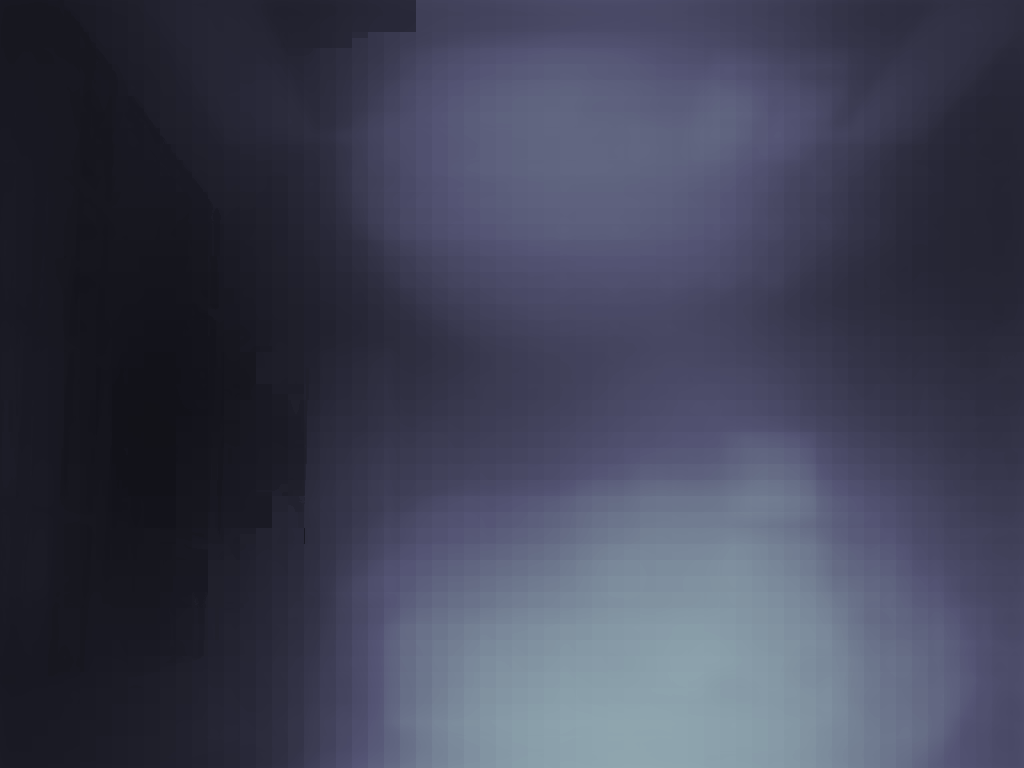}
                \includegraphics[width=\linewidth]{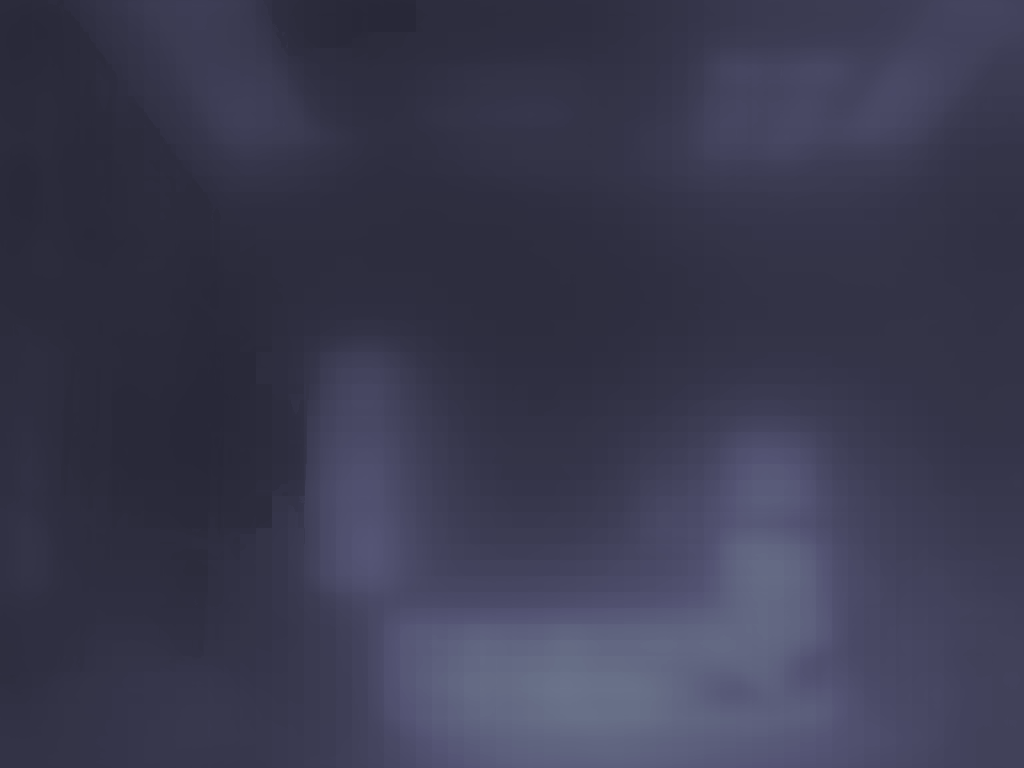}
                \includegraphics[width=\linewidth]{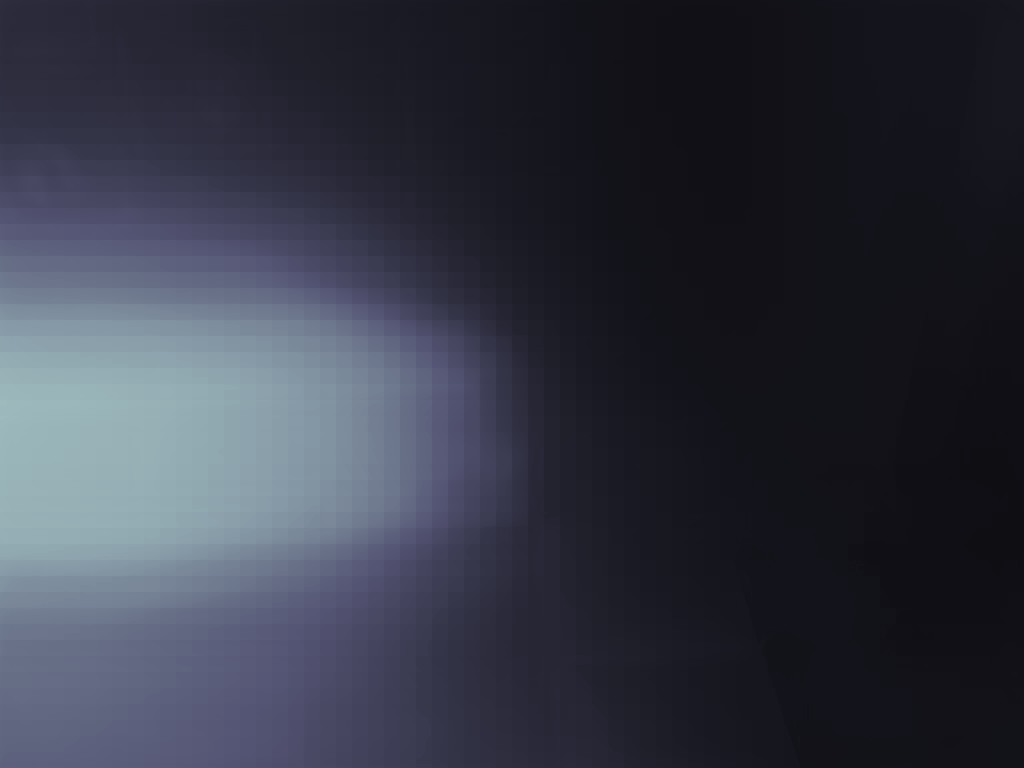}
                \includegraphics[width=\linewidth]{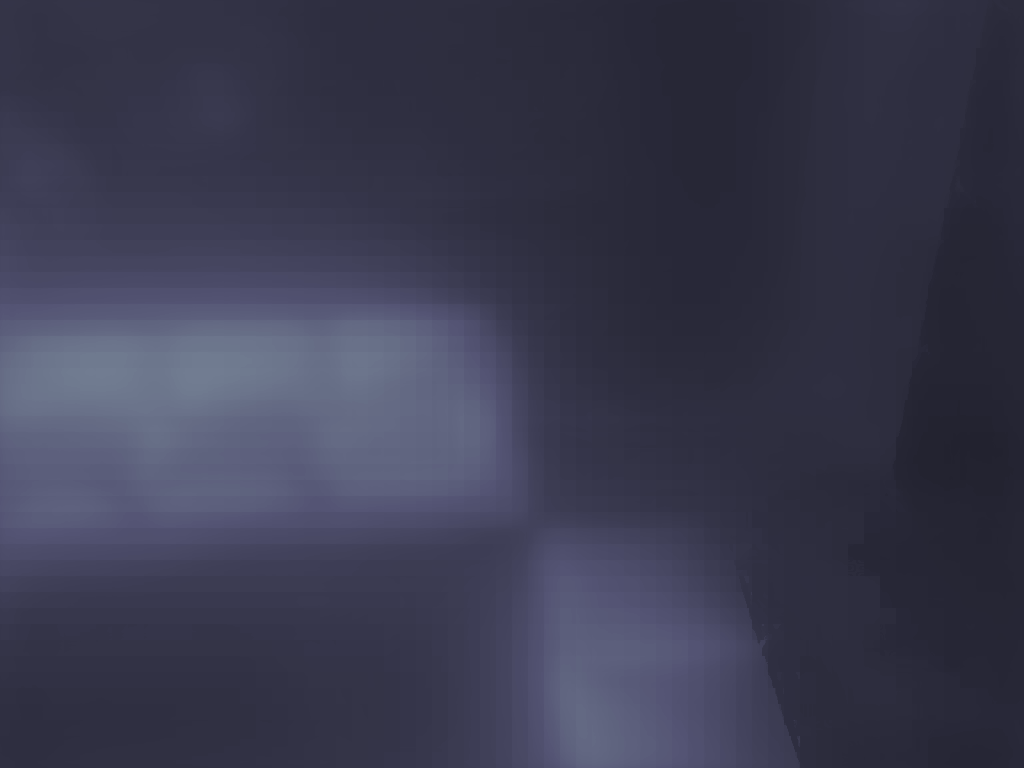}
                \includegraphics[width=\linewidth]{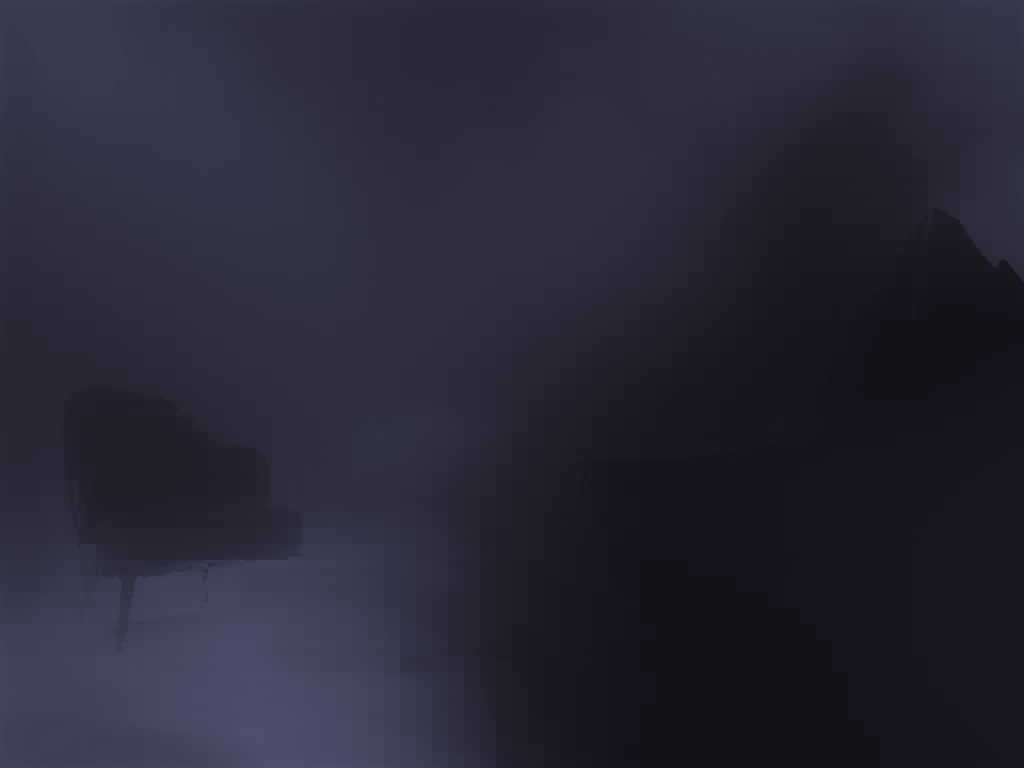}
                \includegraphics[width=\linewidth]{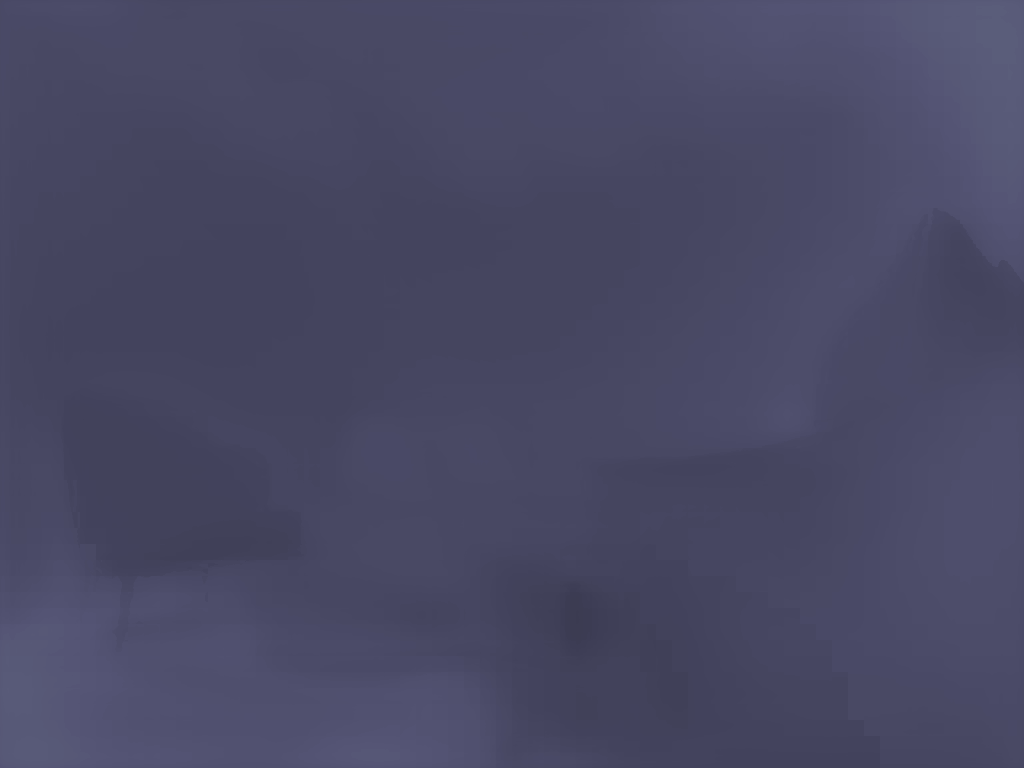}
                \caption{$\textbf{W}_T$}
                \label{fig:deep_wb}
        \end{subfigure}
        \begin{subfigure}[b]{0.24\linewidth}
                \includegraphics[width=\linewidth]{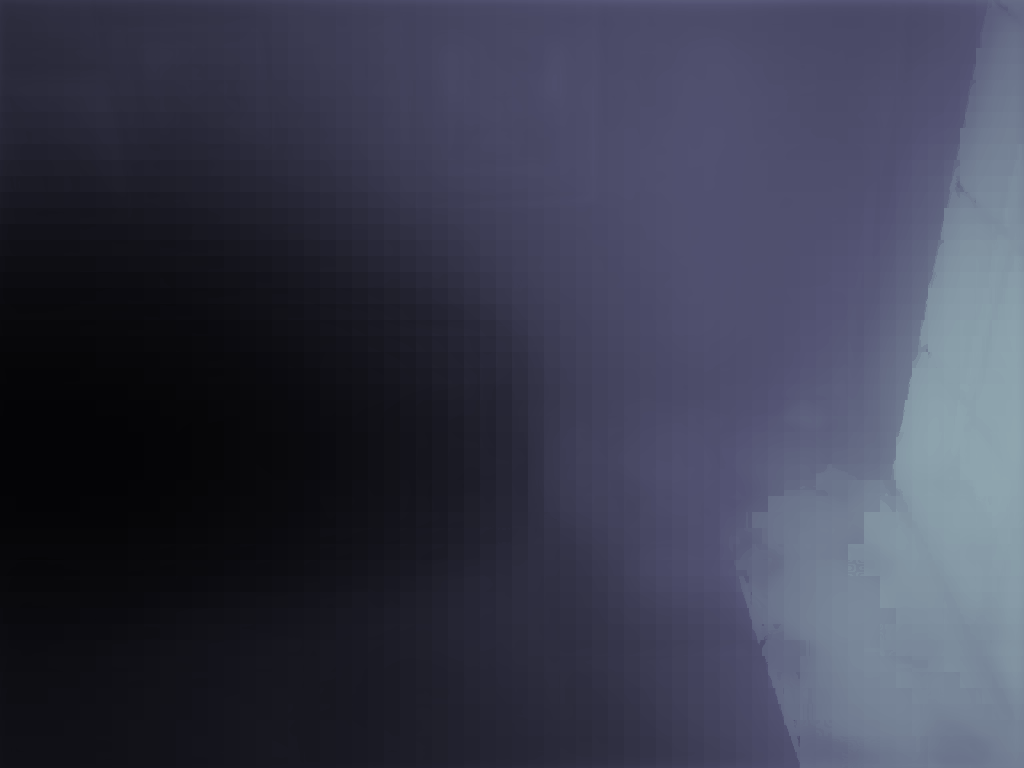}
                \includegraphics[width=\linewidth]{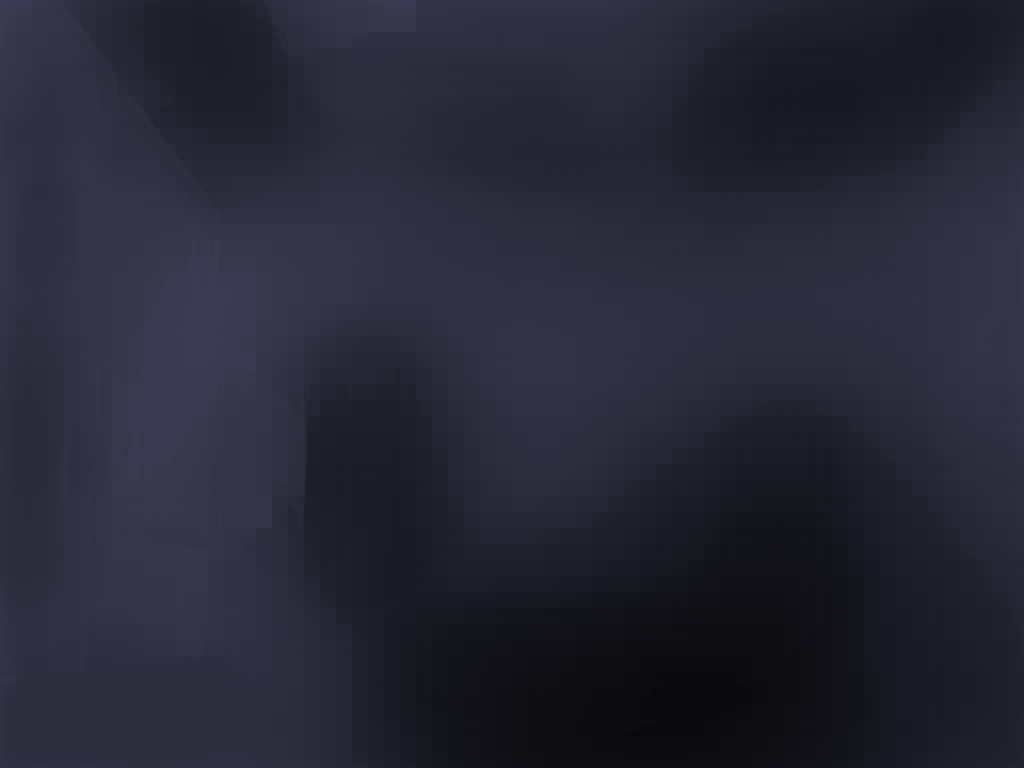}
                \includegraphics[width=\linewidth]{imgs/fig4/mixedIll/scene_16_weight_S_heatmap.png}
                \includegraphics[width=\linewidth]{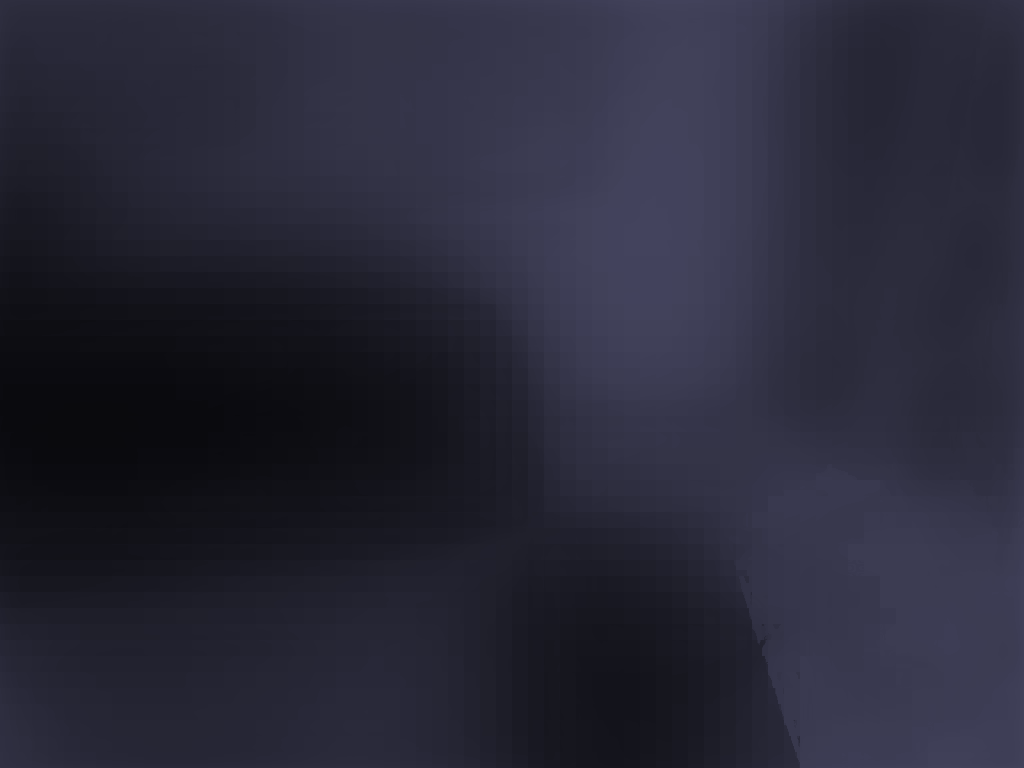}
                \includegraphics[width=\linewidth]{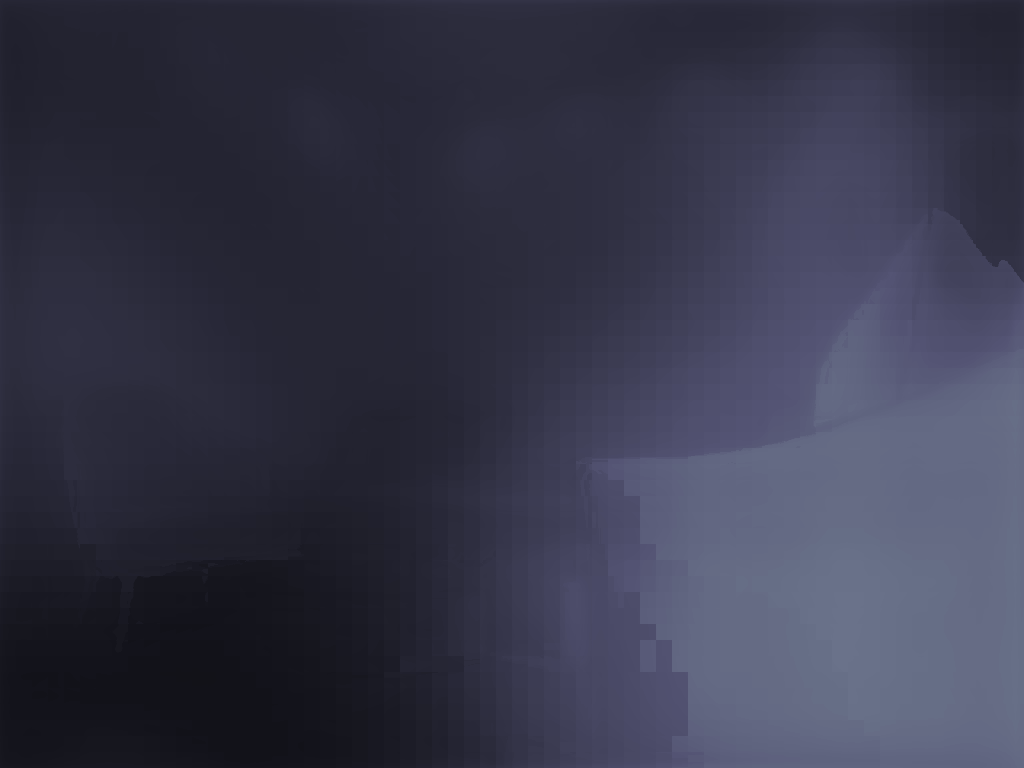}
                \includegraphics[width=\linewidth]{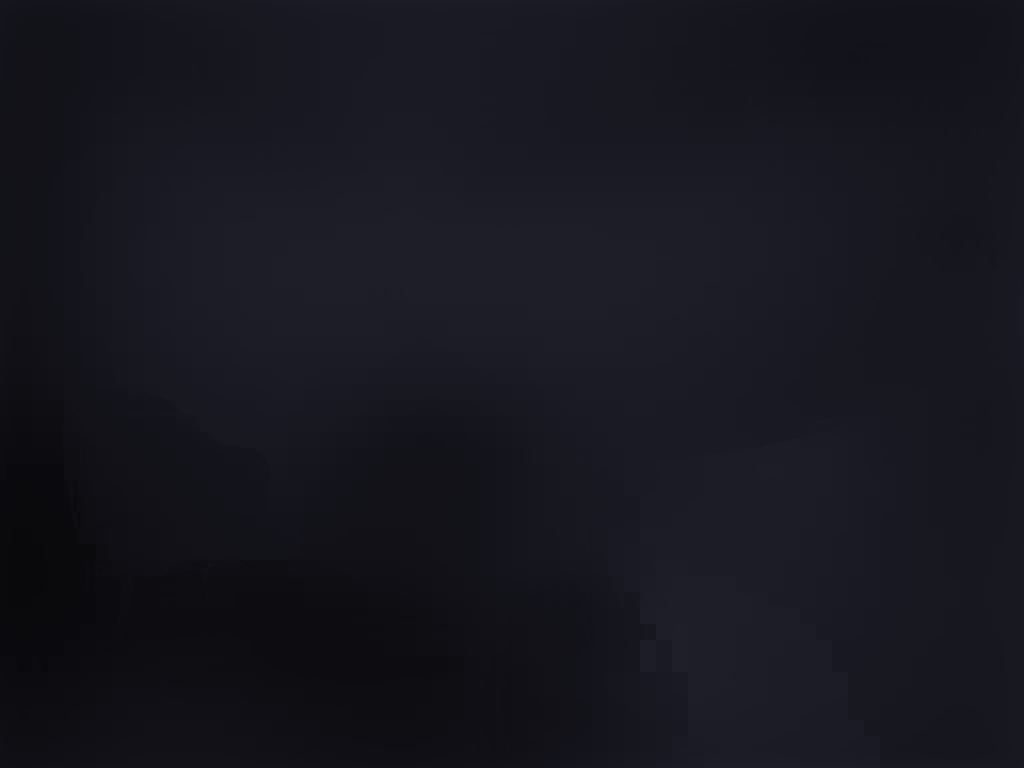}
                \caption{$\textbf{W}_S$}
                \label{fig:mixed_wb}
        \end{subfigure}
        \begin{subfigure}[b]{0.24\linewidth}
                \includegraphics[width=\linewidth]{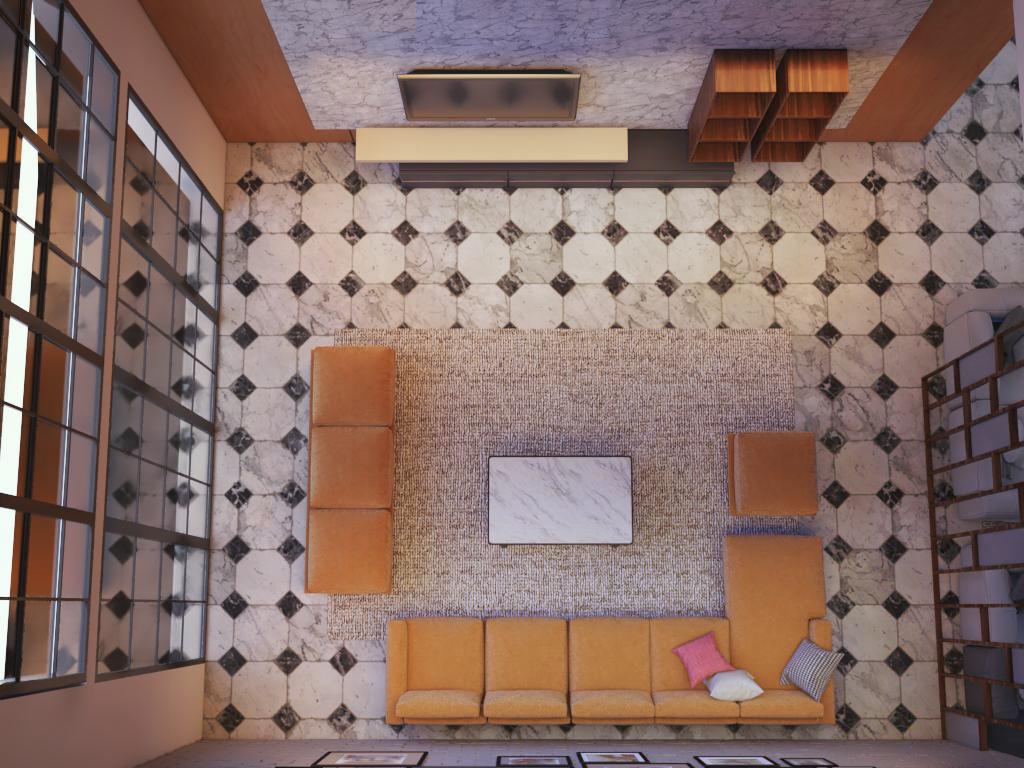}
                \includegraphics[width=\linewidth]{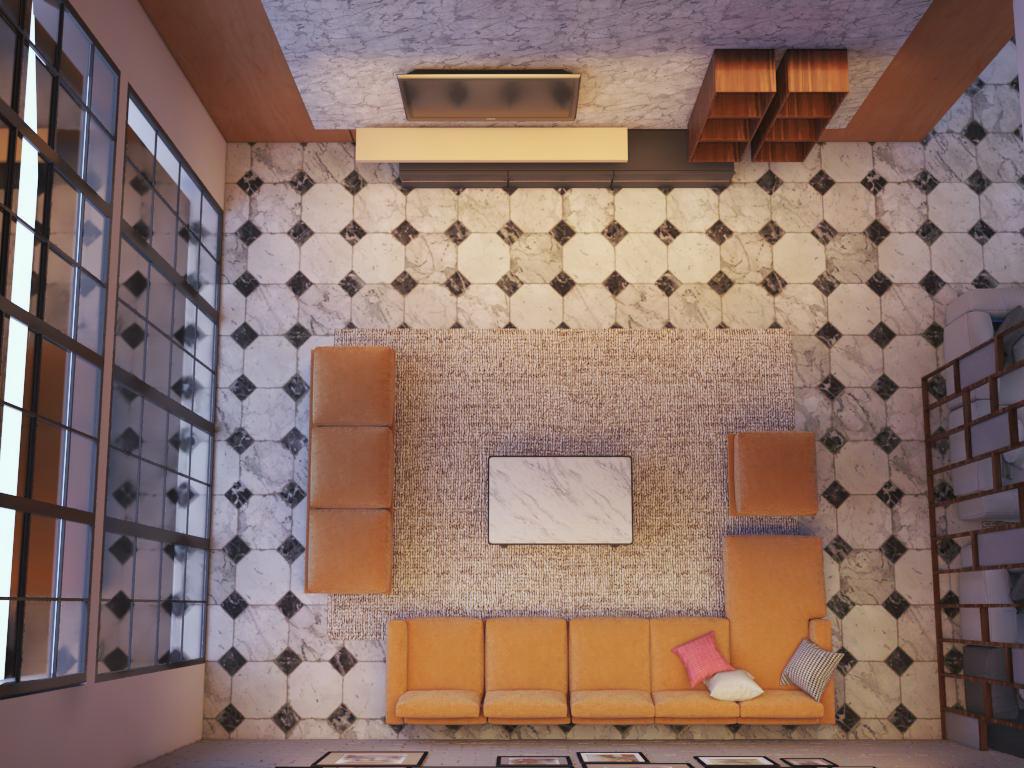}
                \includegraphics[width=\linewidth]{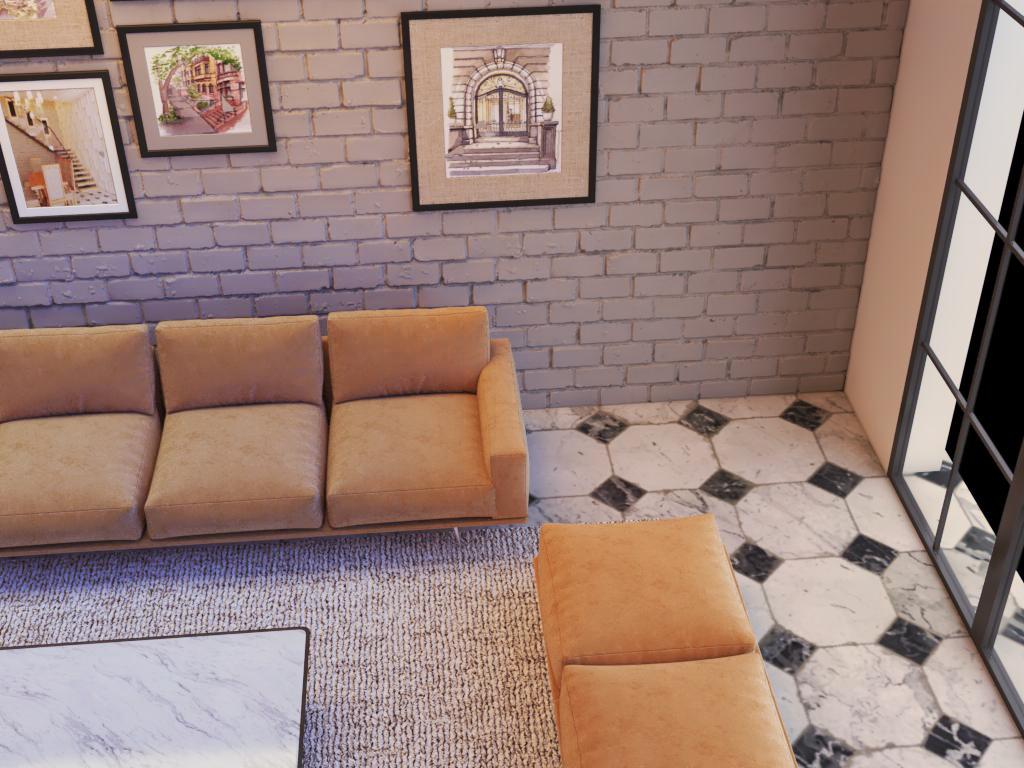}
                \includegraphics[width=\linewidth]{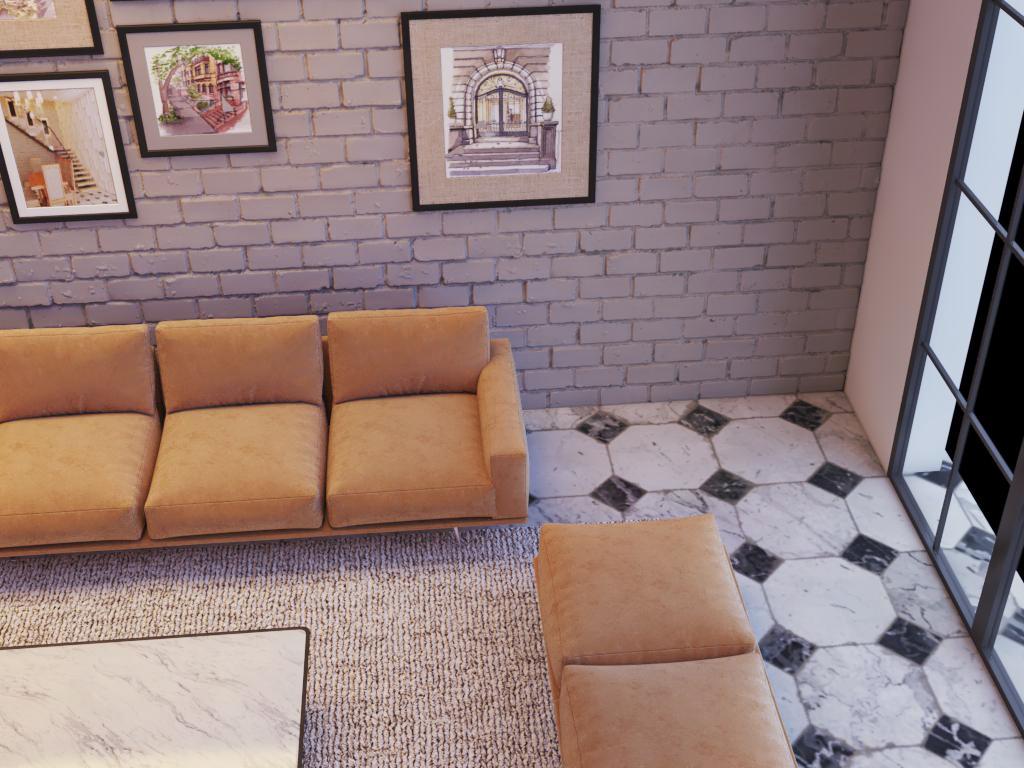}
                 \includegraphics[width=\linewidth]{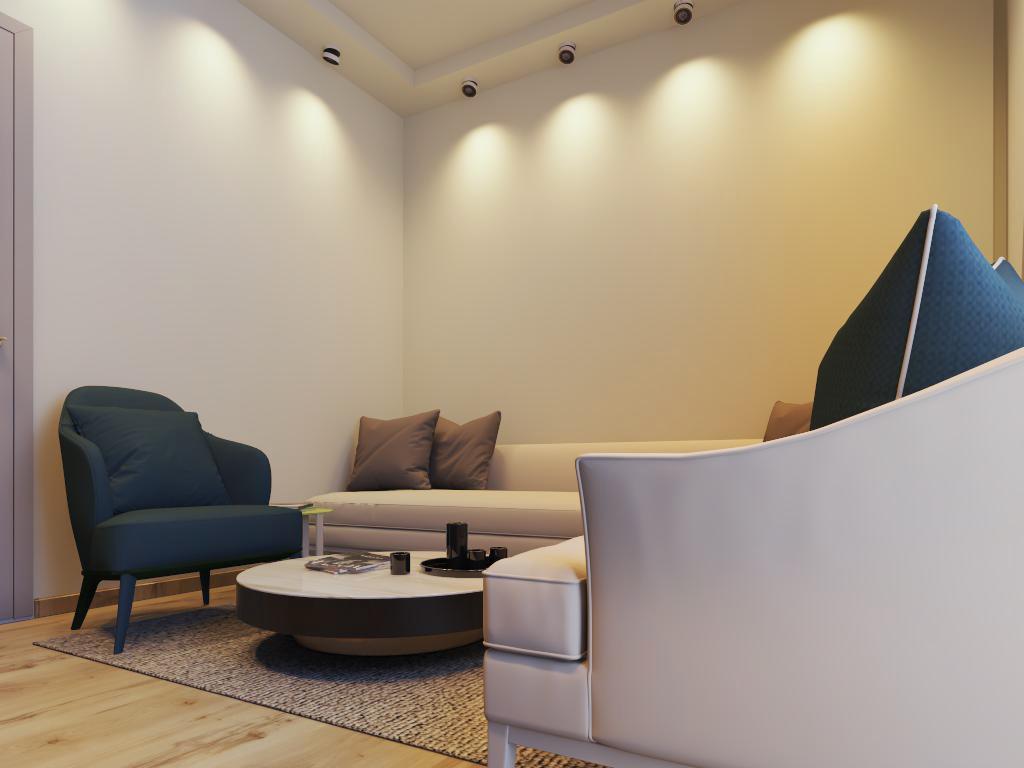}
                \includegraphics[width=\linewidth]{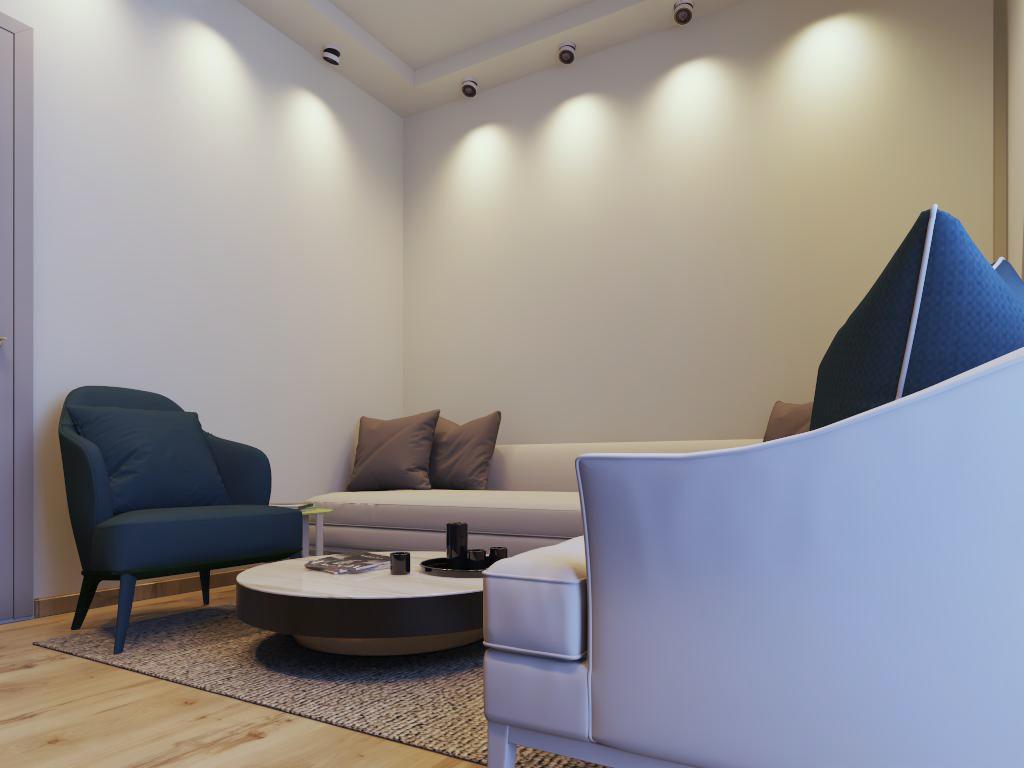}
                \caption{AWB}
                \label{fig:our_wb}
        \end{subfigure}
         \caption{Comparison of the performance of the prior work \cite{Afifi_2022_WACV} and our method on mixed-illuminant dataset. \textbf{Rows:} (odd) Mixed WB results, (even) Our results. Weighting maps for $\textbf{W}_D$: Daylight, $\textbf{W}_T$: Tungsten, $\textbf{W}_S$: Shade.}\label{fig:qual-synt} 
\end{figure}

\subsection{Results}

In this work, we propose to model the lighting in single- and mixed-illuminant scenes as style to improve the AWB strategy proposed in \cite{Afifi_2022_WACV}. To demonstrate the qualitative results of our strategy, we use a set of images containing multiple illuminant scene from MIT-Adobe FiveK dataset \cite{global_tonal_adj}. At this point, we first render the linear raw DNG image files with different WB settings (\textit{e.g.} daylight, tungsten, shade) by using the MATLAB code shared by \cite{color_temp_tuning}. Then, we feed these rendered images to proposed network for extracting their weighting maps to blend the final AWB corrected image. Note that using different re-touched versions of these images may produce different results. Moreover, for single-illuminant scenes, we compare the performance of our proposed strategy with the recent studies \cite{Hu_2017_CVPR,Bianco_2019_CVPR,afifi2019color,afifi2020interactive,Afifi_2020_CVPR,Afifi_2022_WACV} on Cube+ dataset \cite{banic2019unsupervised}. Next, we include our results to the benchmark on mixed-illuminant evaluation set \cite{Afifi_2022_WACV}. Following the prior work, we reported the mean, first (\textbf{Q1}), second (\textbf{Q2}) and third (\textbf{Q3}) quantile of mean-squared error, mean angular error and the color difference error on both datasets.

Figure \ref{fig:qual-wm} demonstrates the examples of predicted weighting maps of different WB settings and the AWB results blended by these maps. Samples are selected from MIT-Adobe FiveK dataset \cite{global_tonal_adj} where their indices are 323 and 2808 in a top-down order. The results indicate that style factor can represent the illuminant in a more detail-oriented way, and thus producing more interpretable weighting maps. Instead of roughly representing the region of the light falling to the objects in the captured scene, our method can differentiate the different illuminants on the same object accurately. This leads to improve the performance of the AWB strategy proposed in the prior work \cite{Afifi_2022_WACV}. Note that our method does not require any illuminant estimation step. Moreover, in Figure \ref{fig:qual-mit}, we introduce the comparison of the qualitative results of our AWB method and the recent methods \cite{Afifi_2020_CVPR,Afifi_2022_WACV} on the selected samples from the same dataset. It shows that our proposed method achieves competitive per-pixel performance on AWB correction in sRGB space when compared to the recent methods.

\begin{table}
  \begin{center}
    \resizebox{\linewidth}{!}{
\begin{tabular}{c|ccc}
\toprule
Models & \textbf{MSE} & \textbf{MAE} & $\Delta$\textbf{E 2000} \\
\midrule
\multicolumn{4}{c}{Single-illuminant dataset, WB = \texttt{\{t,d,s\}}, $p = 64$} \\
\midrule
$ms = 0$, $eas = 0$ & 98.55 & 2.71$^{\circ}$ & 3.32\\
$ms = 1$, $eas = 0$ & 93.78 & 2.59$^{\circ}$ & 3.15\\
$ms = 0$, $eas = 1$ & 97.20 & 2.66$^{\circ}$ & 3.28\\
$ms = 1$, $eas = 1$ & \textbf{92.65} & \textbf{2.47}$^{\circ}$ & \textbf{2.99}\\
\midrule
\multicolumn{4}{c}{Mixed-illuminant dataset, WB = \texttt{\{t,d,s\}}, $p = 128$} \\
\midrule
$ms = 0$, $eas = 0$ & 878.58 & 5.05$^{\circ}$ & 12.12\\
$ms = 1$, $eas = 0$ & 843.50 & \textbf{5.04}$^{\circ}$ & 11.70\\
$ms = 0$, $eas = 1$ & 843.64 & \textbf{5.04}$^{\circ}$ & 11.98\\
$ms = 1$, $eas = 1$ & \textbf{822.77} & 5.11$^{\circ}$ & \textbf{11.65}\\
\bottomrule
\end{tabular}
}
\end{center}
\caption{The ablation study on using multi-scale weighting maps and applying edge-aware smoothing to weighting maps. $p$: patch size, $ms$: multi-scale weighting maps, $eas$: edge-aware smoothing.}
\label{tab:ablation_pp_ms}
\end{table}

Table \ref{table:results_1} presents the quantitative results of our method and the recent methods, which are evaluated on Cube+ dataset. We have conducted our experiments with different patch sizes (\ie $64$ and $128$) and different sets of WB settings (\ie $\{\texttt{t,d,s}\}$ and $\{\texttt{t,f,d,c,s}\}$). All of our results outperform the results of the other compared methods over the most parts of all metrics. Particularly, the model trained with the patch size of $64$ and on WB settings of $\{\texttt{t,d,s}\}$ achieves the best performance among the other models with different settings. As also stated in \cite{Afifi_2022_WACV}, we state that smaller patch sizes lead to better model the illuminant. However, in contrast to \cite{Afifi_2022_WACV}, increasing the number of WB setting in the set of training WB settings does not provide any advantage on modeling the lighting in single-illuminant scenarios. We think that less number of WB settings to blend for the final output makes easier to build the knowledge on the correlation between pixels by weighting maps. We also believe that using the input with more channels for training may increase the required complexity of the architecture to model the illuminant, and it may not be fair to compare them with the exact same architecture. Lastly, due to the style extraction part of the proposed network, our method has larger memory overhead when compared to the recent methods. 

\captionsetup[subfigure]{labelfont=bf, labelformat=parens}
\begin{figure}[!t]
        \centering
        \begin{subfigure}[b]{0.32\columnwidth}
                \includegraphics[width=\textwidth]{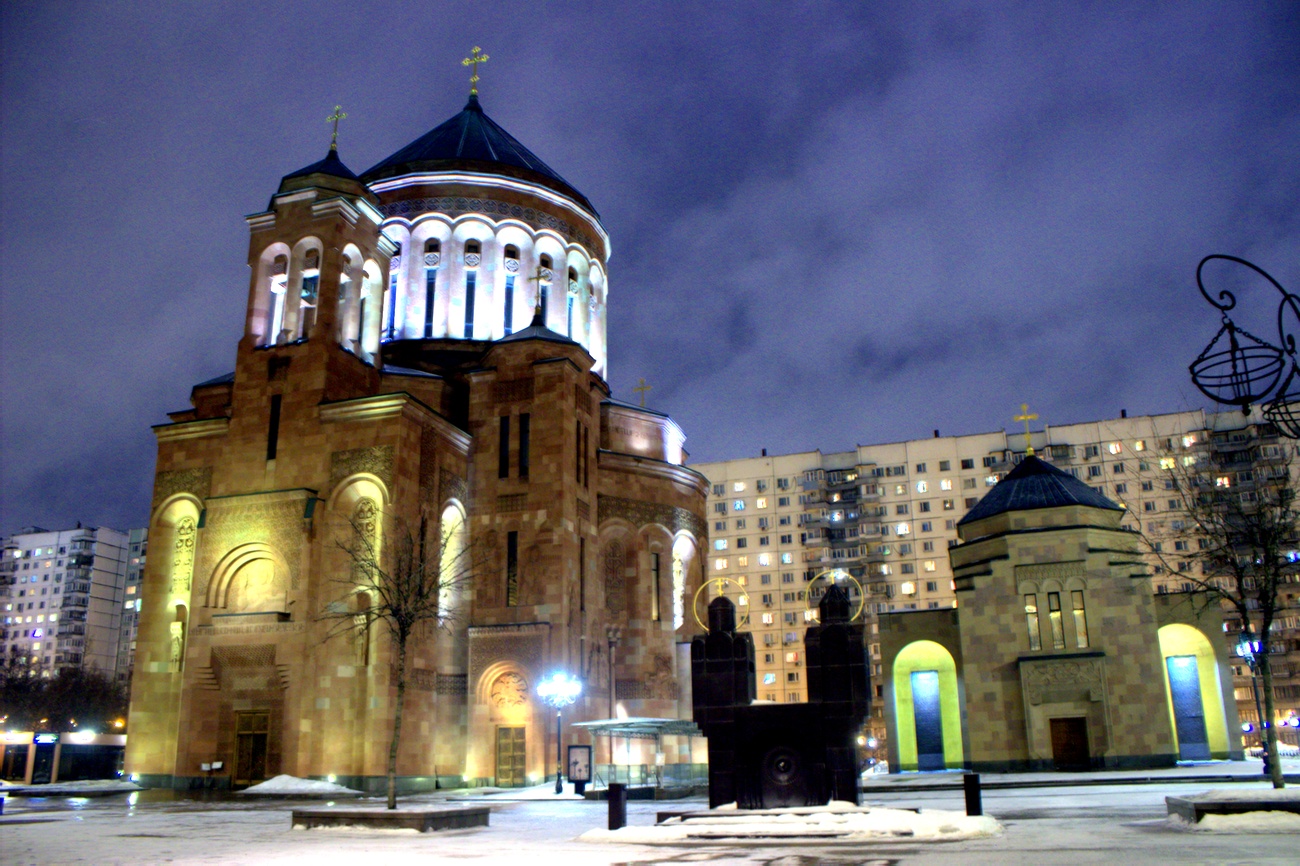}
                \includegraphics[width=\textwidth]{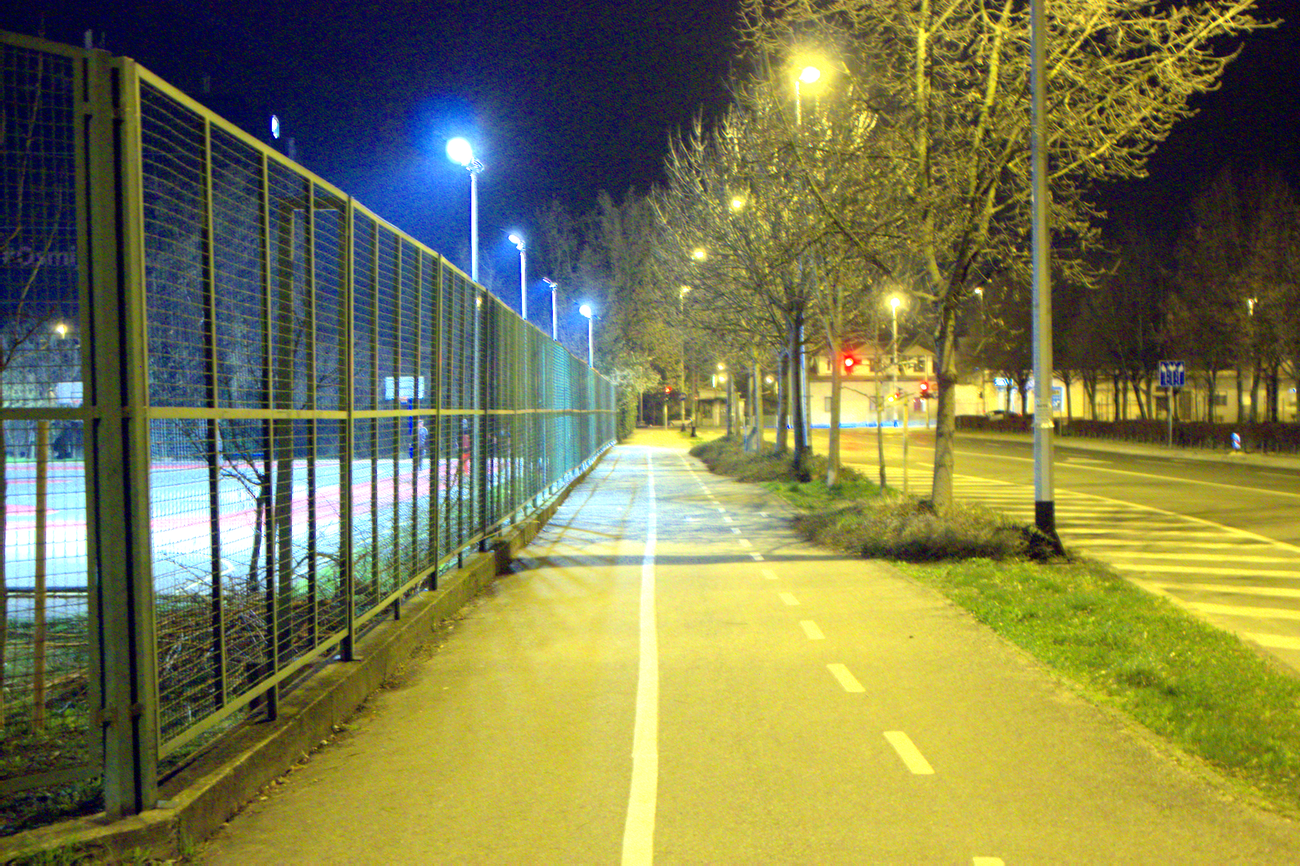}
                \includegraphics[width=\textwidth]{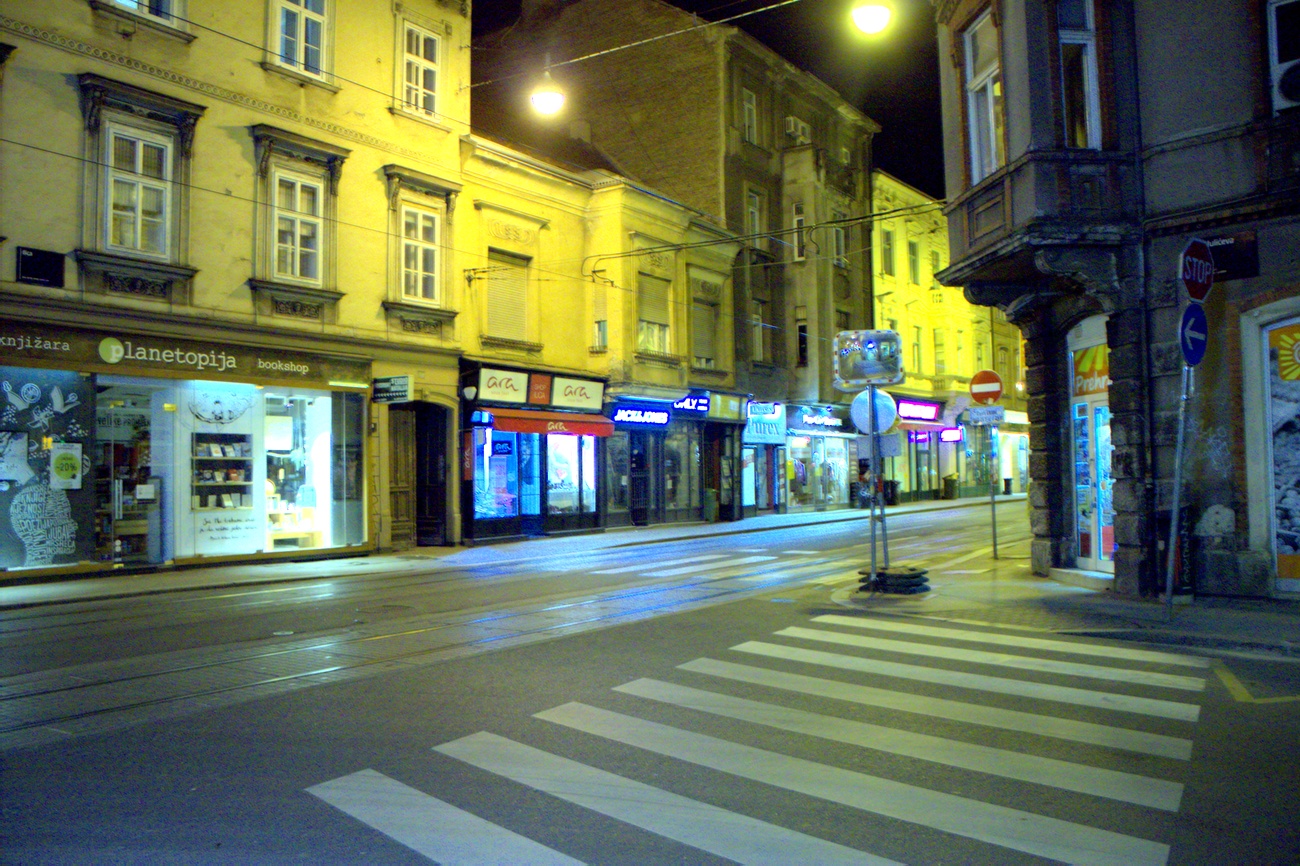}
                \includegraphics[width=\textwidth]{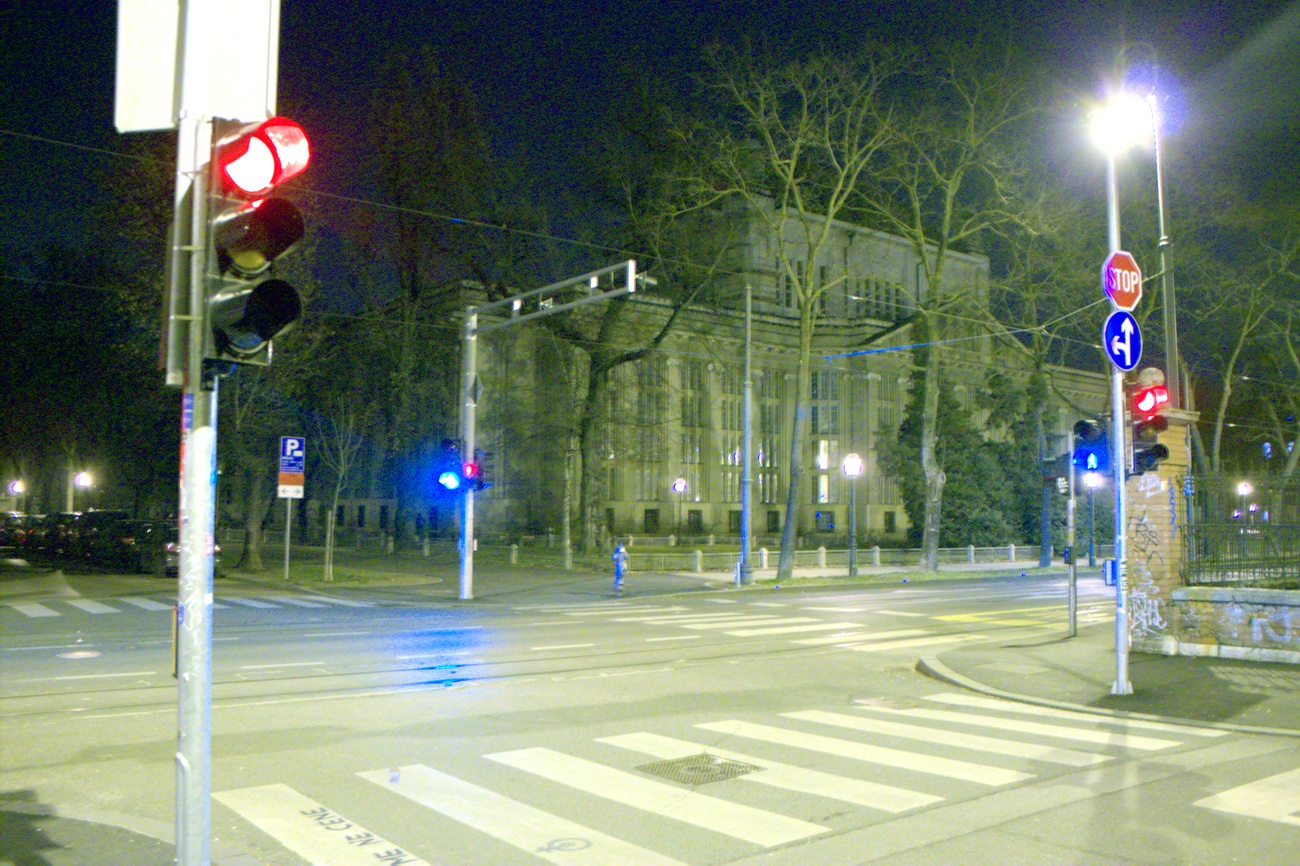}
                \caption{Standard ISP}
                \label{fig:s_isp}
        \end{subfigure}       
        \begin{subfigure}[b]{0.32\columnwidth}
                \includegraphics[width=\textwidth]{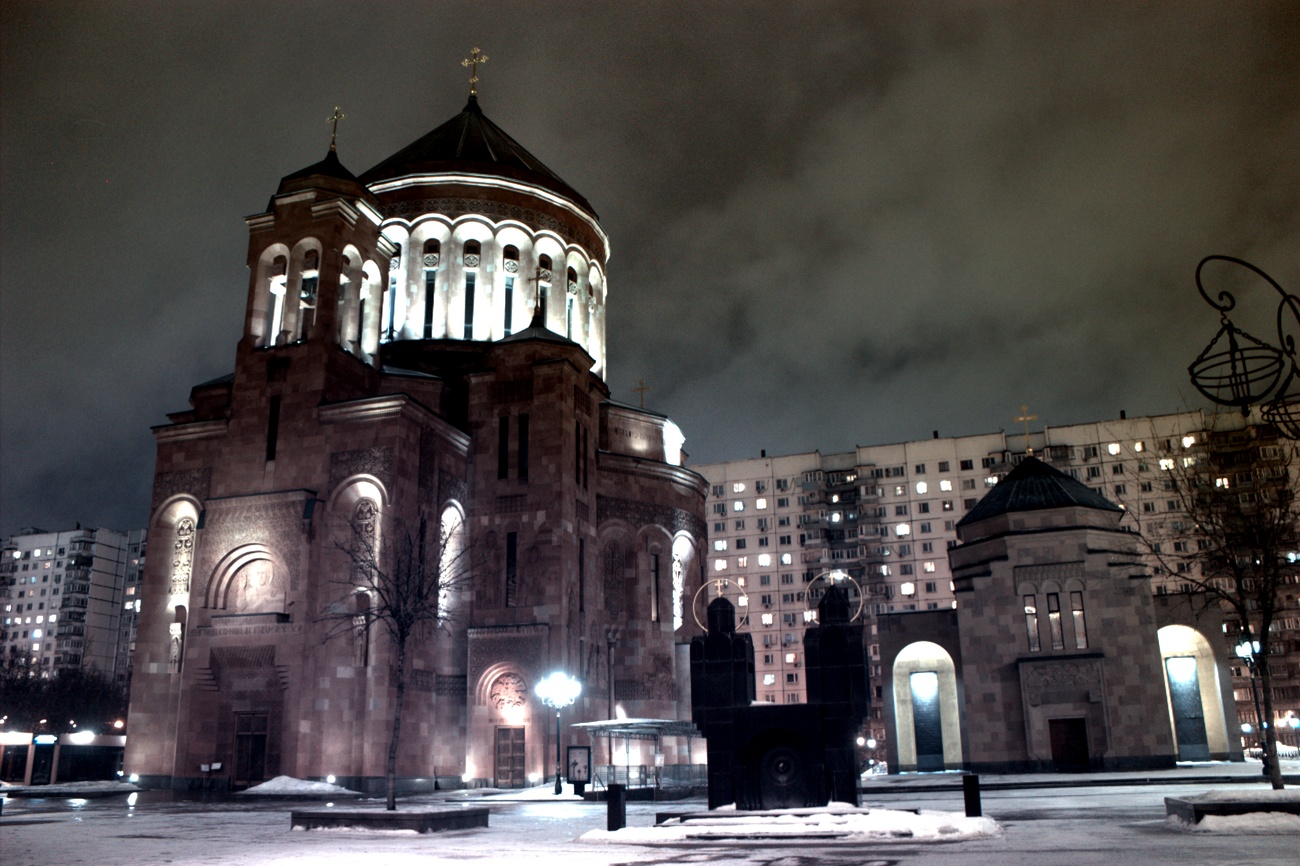}
                \includegraphics[width=\textwidth]{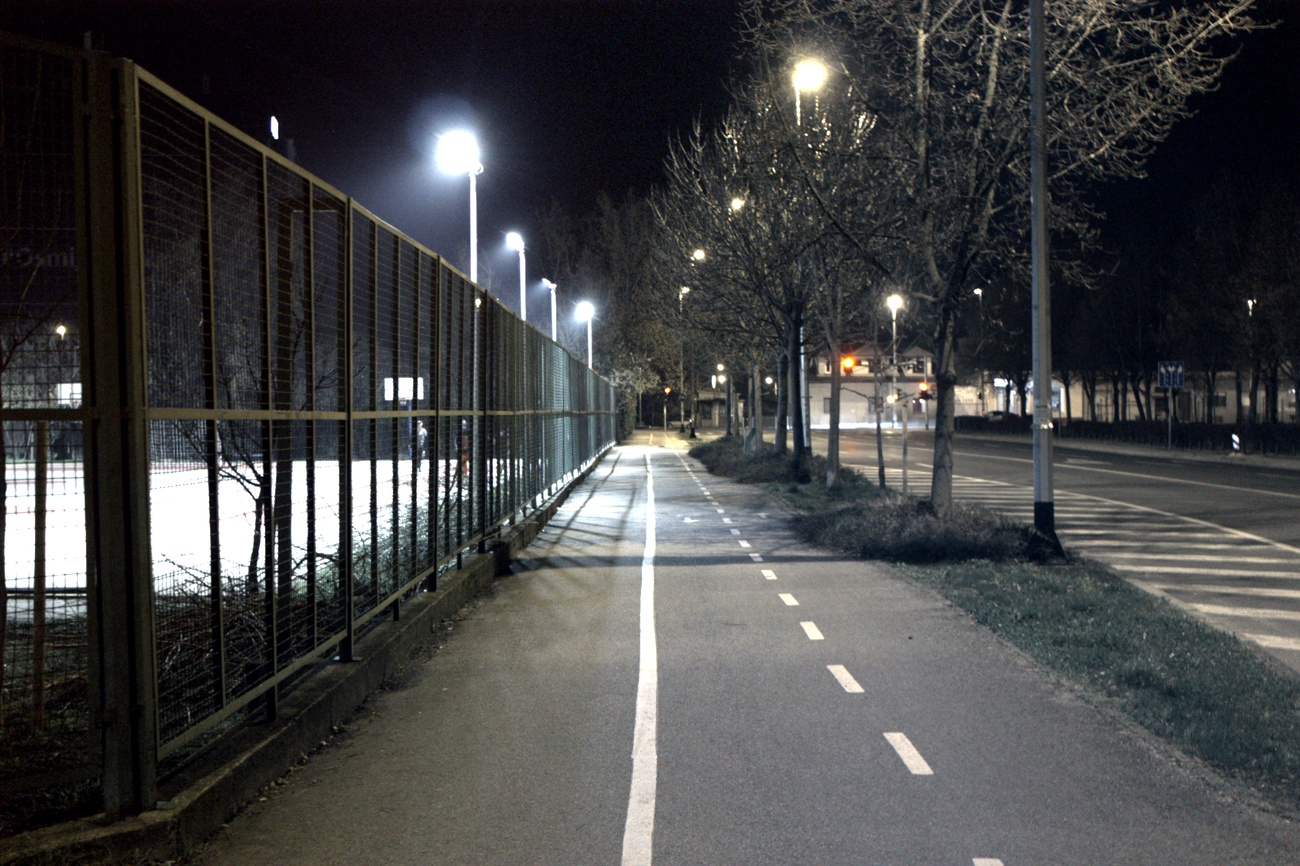}
                \includegraphics[width=\textwidth]{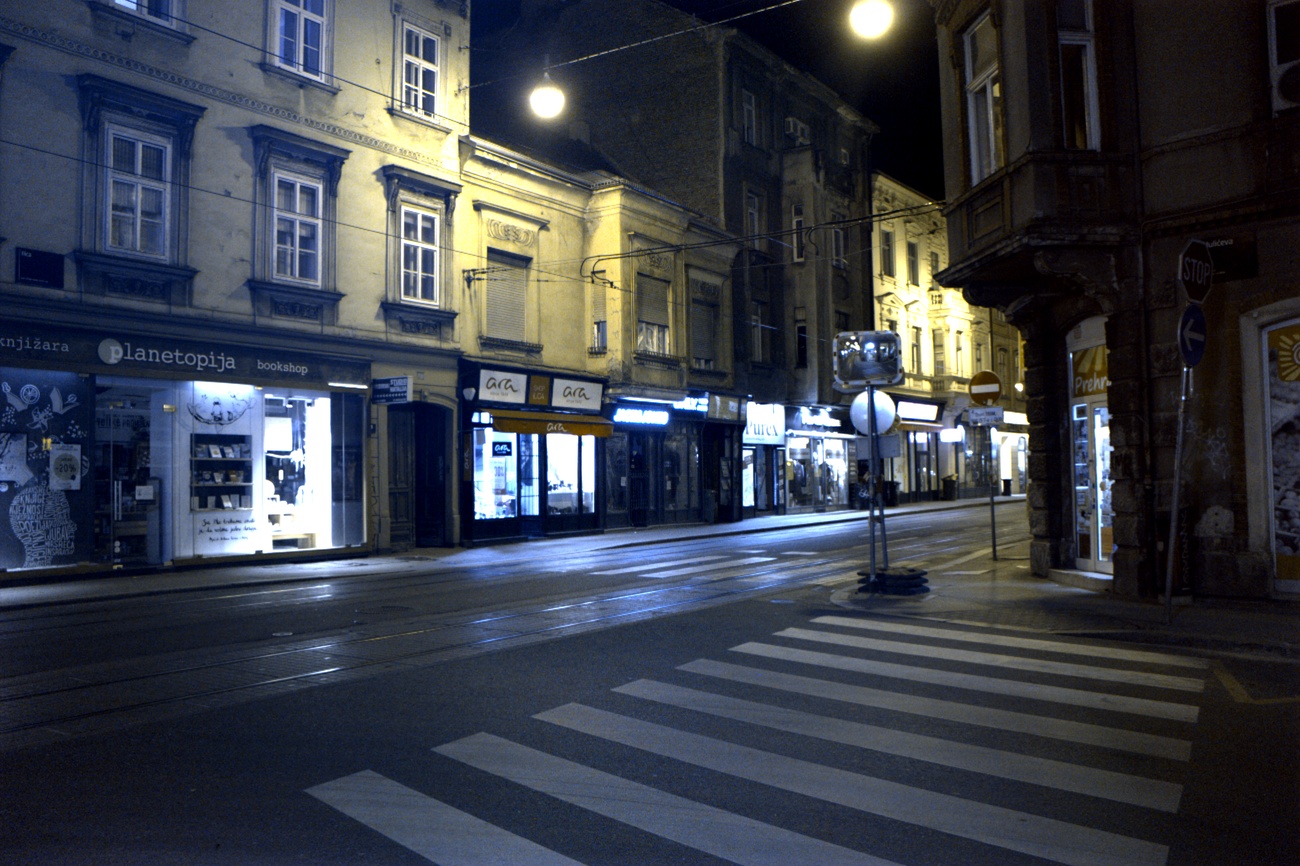}
                \includegraphics[width=\textwidth]{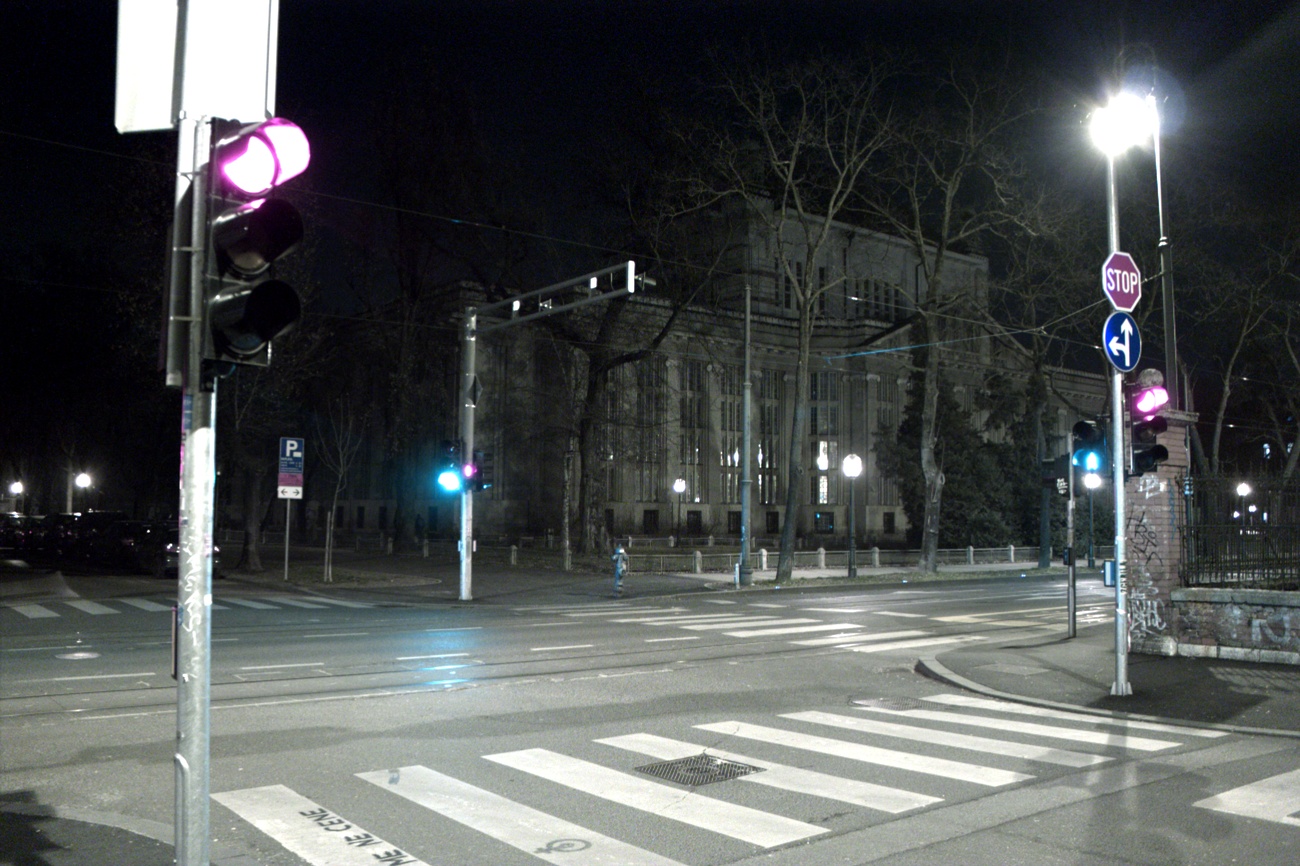}
                \caption{ISP with \cite{Afifi_2022_WACV}}
                \label{fig:mixed_wb_isp}
        \end{subfigure}
        \begin{subfigure}[b]{0.32\columnwidth}
                \includegraphics[width=\textwidth]{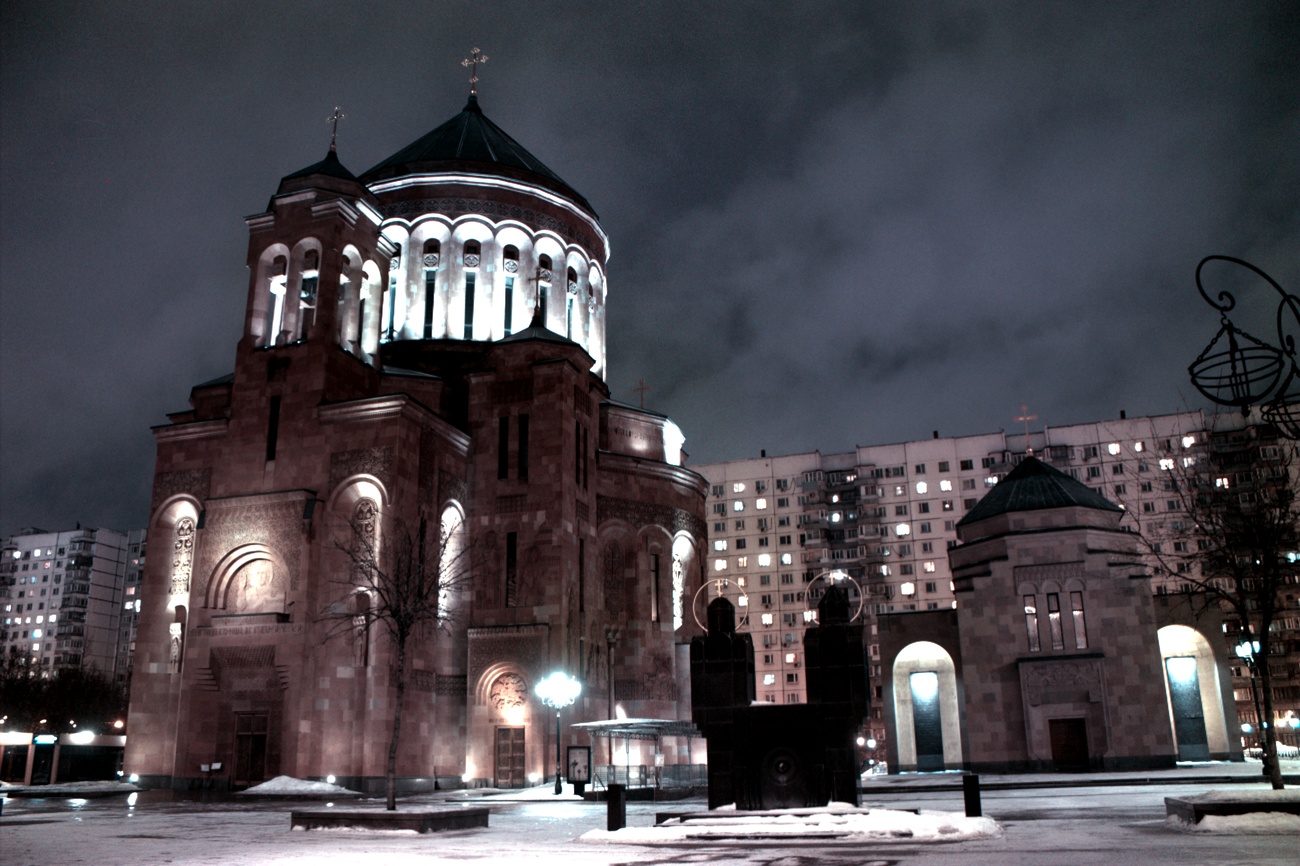}
                \includegraphics[width=\textwidth]{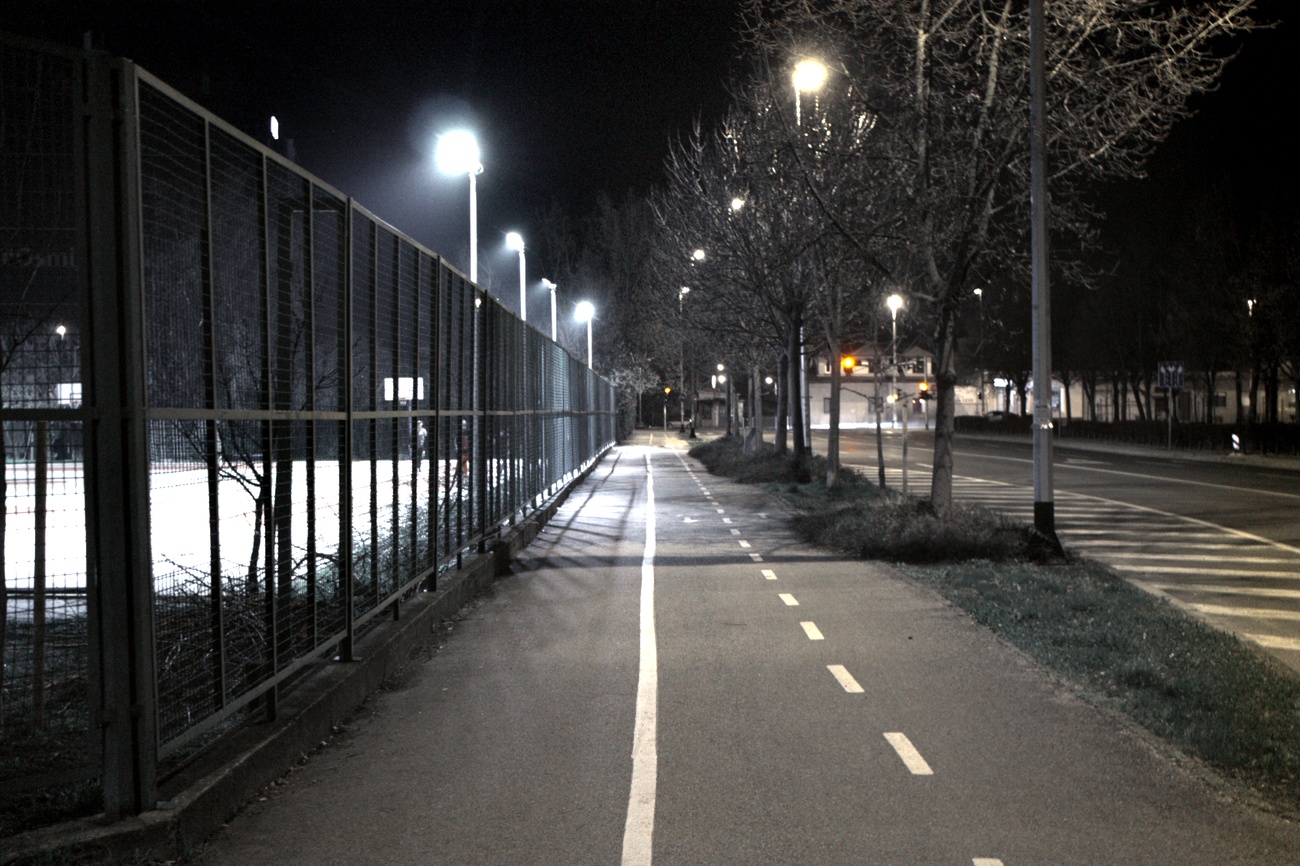}
                \includegraphics[width=\textwidth]{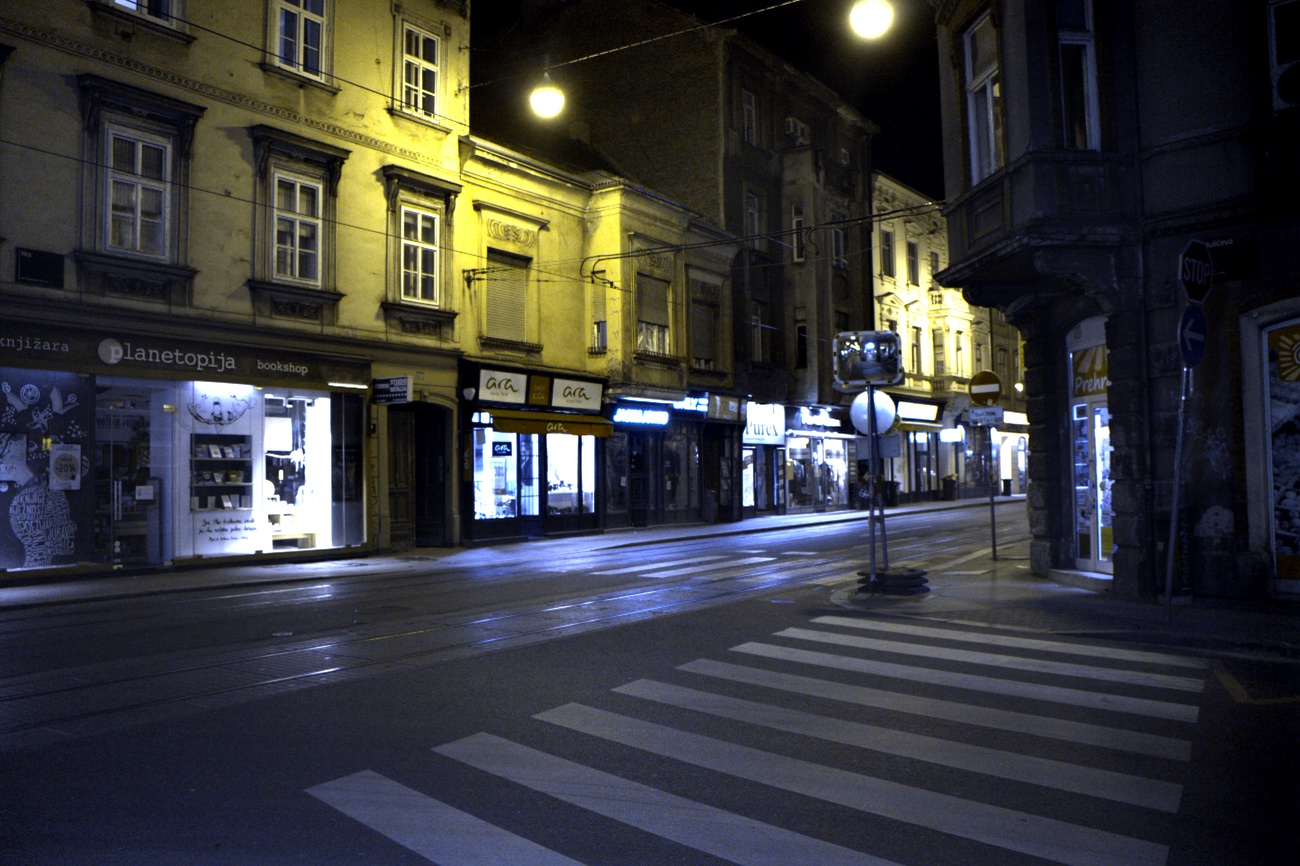}
                \includegraphics[width=\textwidth]{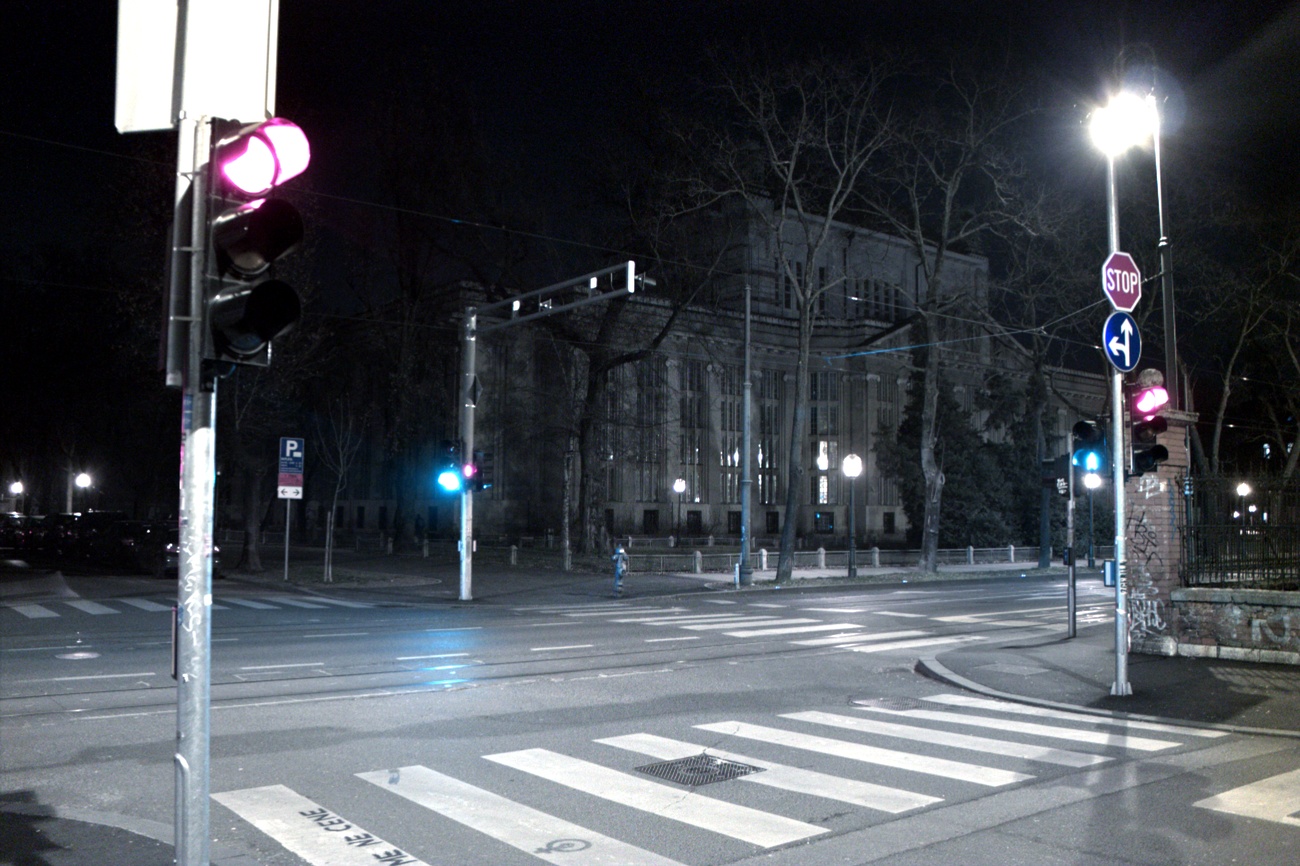}
                \caption{ISP with ours}
                \label{fig:our_isp}
        \end{subfigure}
         \caption{Comparison on the night photography rendering results of the standard camera pipeline, the prior work \cite{Afifi_2022_WACV} and our method. }\label{fig:qual-night} 
\end{figure}

We have conducted the same experiments on synthetic mixed-illuminant evaluation set \cite{Afifi_2022_WACV}, and the overall results are reported in Table \ref{table:results_2}. According to these results, it is hard to pick a superior method over all other methods, as depending on the metric. Our method performs better when evaluated on mean-squared error, while achieving competitive results compared to the recent methods for mean angular error and color difference metrics. Moreover, Figure \ref{fig:qual-synt} demonstrates the visual comparison of the weighting maps and the blended AWB results of the prior work and our method on this synthetic dataset. As several illumination sources can affect the different parts of a single object at the same time, modeling the lighting as style helps to produce more detailed weighting maps, especially for the parts containing objects. The weighting maps produced by our proposed network give more detail-oriented results when compared to the prior work, even if it falls behind the prior work on some quantitative metrics. Synthetic data could have more sharp edges than the real-world images. When we use this kind of blending strategy with detail-oriented weighting maps and not including such samples to training, it may have caused the color discrepancy on the edges of the final output. This may reduce the quantitative performance of our method on this dataset.

As ablating the effects of two post-processing methods proposed in \cite{Afifi_2022_WACV} on the performance of our strategy, we have conducted additional experiments on our best-performed settings for both datasets where the post-processing methods are alternately excluded during inference time. Table \ref{tab:ablation_pp_ms} presents the results of different combinations of the post-processing methods applied. It verifies that these methods help to improve the quality of the weighting maps, so do the qualitative results. At this point, one may argue that adding total variation regularization term to the final objective function can also achieve a similar improvement, and resolve the need of post-processing. This addition would require re-considering the smoothing loss and the pipeline proposed by \cite{Afifi_2022_WACV}, which is beyond the scope of this study.

Moreover, night photography rendering \cite{Ershov_2022_CVPR} is another challenging task containing different scenarios affected by multiple illuminants. Correcting AWB for the images captured at night may not be easily handled by assuming the illuminant in the scene is global. To show the validity of our strategy, we integrate our AWB method into the camera ISP for processing night images. Figure \ref{fig:qual-night} presents the rendered night images by the standard camera pipeline, and its variants that include Mixed WB \cite{Afifi_2022_WACV} and our AWB method. Results show that the pipeline including our AWB method produces more natural and realistic images over a wide range of night images. At this point, we assess the results according to how similar the produced colors of the objects in the scene to the human visual perception, not visual plausibility.  Note that we include the same operations (\textit{i.e.} denoising \cite{10.1007/978-3-030-58577-8_11,mentes2021re}, gamma correction, tone mapping and auto-contrast) to the pipeline in the same order, except white-balancing strategies.

\section{Conclusion}
In this work, we have proposed a novel idea of modeling the lighting as style factor for improving the recent AWB correction methods for single- and mixed-illuminant scenes. Our proposed network extracts the additional style information injected by the lighting sources to the scene, and learns to weight the maps of different WB settings to blend them for AWB correction. We have conducted several experiments on the datasets containing mostly single-illuminant scenes, the synthetic mixed-illuminant evaluation set and night photography rendering set. The results indicate the illuminant can be modeled by the style factor, and our method produces promising correction results in both real-world scenarios and the synthetic scenes without requiring illuminant estimation. The next steps for this task could be to design a style extraction module with lower memory overhead without sacrificing the performance. 



{\small
\bibliographystyle{ieee_fullname}
\bibliography{egbib}
}

\end{document}